\documentclass[10pt,journal,compsoc]{IEEEtran}
\usepackage{url}
\usepackage{graphicx}
\usepackage{multirow}
\usepackage{multicol}
\usepackage{amsmath,amssymb}
\usepackage{bbding}
\usepackage{subfigure}
\usepackage{color}
\usepackage{ulem}
\usepackage{xcolor}
\usepackage{color}
\usepackage{hyperref} 
\usepackage{epsfig}
\usepackage{float} 
\usepackage[T1]{fontenc}
\usepackage{ulem}
\usepackage{algorithm}
\usepackage{algorithmic}
\usepackage{float}
\usepackage{threeparttable} 
\useunder{\uline}{\ul}{}
\newenvironment{myitemize}[1][]{
	\begin{list}{$\bullet$}
		{
			\setlength{\leftmargin}{0mm}
			\setlength{\parsep}{1mm}
			\setlength{\topsep}{0mm}
			\setlength{\itemsep}{0mm}
			\setlength{\labelsep}{1.5mm}
			\setlength{\itemindent}{5mm}
			\setlength{\listparindent}{5mm}
	}}
	{\end{list}}
\usepackage[justification=centering]{caption}
\ifCLASSOPTIONcompsoc
\usepackage[nocompress]{cite}
\else
\usepackage{cite}
\fi
\ifCLASSINFOpdf
\else
\fi
\begin{document}
	%
	\title{LuoJiaHOG: A Hierarchy Oriented Geo-aware Image Caption Dataset for Remote Sensing Image-Text Retrival}
	
	\author{
		Yuanxin Zhao,  Mi Zhang $^{{\dag}}$ ~\IEEEmembership{Member,~IEEE,} Bingnan Yang, Zhan Zhang, Jiaju Kang, Jianya Gong.
		\IEEEcompsocitemizethanks{
			\IEEEcompsocthanksitem Yuanxin Zhao, Mi Zhang, Bingnan Yang, Jianya Gong are with the School of Remore Sensing and Information Engineering, Wuhan University, No.129, Luoyu Road, Wuhan 430079, China.
			\IEEEcompsocthanksitem $^{{\dag}}$ Corresponding Author: Mi Zhang is also with Hubei Luojia Laboratory, No.129, Luoyu Road, Wuhan 430079, China. \protect\\
			E-mail: mizhang@whu.edu.cn.
		}
	}

	
	\IEEEtitleabstractindextext{%
		\begin{abstract}
			Image-text retrieval (ITR) plays a significant role in making informed decisions for various remote sensing (RS) applications, such as urban development and disaster prevention. Nonetheless, creating ITR datasets containing vision and language modalities not only requires significant geo-spatial sampling area but also varing categories and detailed descriptions. To this end, we introduce an image caption dataset LuojiaHOG, which is geospatial-aware, label-extension-friendly and comprehensive-captioned. LuojiaHOG involves the hierarchical spatial sampling, extensible classification system to Open Geospatial Consortium (OGC) standards, and detailed caption generation. In addition, we propose a CLIP-based Image Semantic Enhancement Network (CISEN) to promote sophisticated ITR. CISEN consists of two components, namely dual-path knowledge transfer and progressive cross-modal feature fusion. The former transfers the multi-modal knowledge from the large pretrained CLIP-like model, whereas the latter leverages a visual-to-text alignment and fine-grained cross-modal feature enhancement. Comprehensive statistics on LuojiaHOG reveal the richness in sampling diversity, labels quantity and descriptions granularity. The evaluation on LuojiaHOG is conducted across various state-of-the-art ITR models, including ALBEF, ALIGN, CLIP, FILIP, Wukong, GeoRSCLIP and CISEN. We use second- and third-level labels to evaluate these vision-language models through adapter-tuning and CISEN demonstrates superior performance. For instance, it achieves the highest scores with WMAP@5 of 88.47\% and 87.28\% on third-level ITR tasks, respectively. In particular, CISEN exhibits an improvement of approximately 1.3\% and 0.9\% in terms of WMAP@5 compared to its baseline. These findings highlight CISEN advancements accurately retrieving pertinent information across image and text. LuojiaHOG and CISEN can serve as a foundational resource for future RS image-text alignment research, facilitating a wide range of vision-language applications.
		\end{abstract}
		
		}

	\maketitle
	\IEEEdisplaynontitleabstractindextext
	\IEEEpeerreviewmaketitle
	
	\IEEEraisesectionheading{\section{Introduction}\label{sec:introduction}}

 Image-text retrieval (ITR) is a critical area of interest that supports various remote sensing challenges such as geo-localization \cite{yu2014full, wang2022visual,jovanovic2007multi}, disaster rescue\cite{le2022multiscale,panteras2018enhancing,ge2020review}, economic assessment\cite{rivest2005solap, de2014potential,milesi2003assessing}, and ecology prediction\cite{reichman2011challenges}. It is essential for automated decision-making and intelligent recommendations, enhancing the capability to access geo-spatial information swiftly and accurately. 
 
 Current works in ITR primarily relies on datasets like UCM-captions\cite{qu2016deep}, RSICD\cite{lu2017exploring}, and NWPU-Captions~\cite{cheng2022nwpu}.  which lack geographic diversity, offer only brief descriptions, and are confined to fixed or mixed classes (Tab.~\ref{datasets_summary}). This limitation hinders the development of more sophisticated and advanced ITR models due to insufficient data variety and a lack of intra-modal and inter-modal semantic similarity. Recognizing the critical role of high-quality datasets in ITR, there is an urgent need for a dataset that incorporates geographic awareness, provides detailed captions, and is adaptable for extensions. Such a dataset would not only advance ITR algorithm development but also enhance related image-text tasks, including image text generation and visual question answering.
 
 \begin{figure}[htbp]
 	\centering
 	\includegraphics[width=0.48\textwidth]{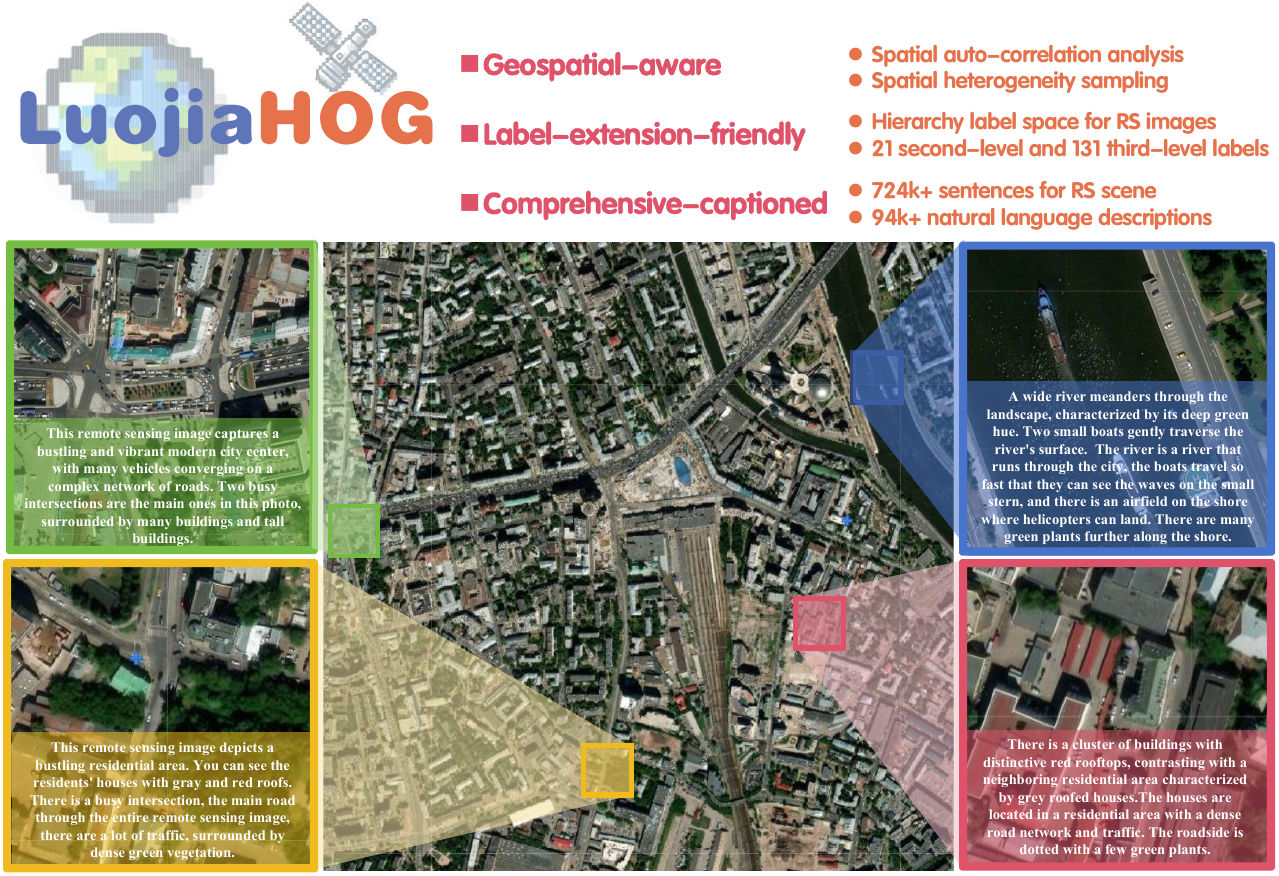}
 	\caption{Overview of ITR dataset LuojiaHOG.}
 	\label{example}
 	\vspace{-2em}
 \end{figure}
 In this study, we introduce a novel image caption dataset, named LuojiaHOG (Fig.~\ref{example}), which is geospatial-aware, label-extension-friendly and comprehensive-captioned, to address the aforementioned issues. Unlike the majority of existing datasets, such as UCM-Captions and RSICD, all images are collected from regions around the world with varing levels of development and topography through geo-spatial analysis. Besides, LuoJiaHOG classification system adopts the OGC standards\cite{kamel2011crowdsourcing} and thus compatible with various new data under different task requirements. It comprises 94,856 images, categorized into 131 third-level categories that fall into 21 second-level classes, including residential, farmland, cemetery, and playground, etc. In addition to rich categories, we have diligently conducted extensive data cleaning and professional annotations, leveraging Vision-Language Models (VLMs) to generate and augment the textual captions automatically. Moreover, prompt engineering is adopted to improve the quality of generated text. 
 LuojiaHOG supports two basic retrieval tasks: text-to-image (T2I) and image-to-text (I2T). By evaluating performance across different granularities using tailored metrics for multi-label retrieval, we establish baseline for state-of-the-art models on ITR. We anticipate it as a fine-grained ITR benchmark, thus facilitating the development of RS vision-language learning. 
 
 The primary contributions of this study can be summarized as follows:
 \begin{myitemize} 
 	\item A hierarchical sampling method and automatic are employed to collect RS images. Both manual and automatic annotation methods are utilized to generate detailed descriptions.
 	\item We establish an extensible classification system, which is aligned with the Open Geospatial Consortium (OGC) standards. It supports dynamic expansion of database for new samples and enables the mapping and conversion of different classification systems.
 	\item Extensive ITR baselines on LuojiaHOG are provided across two levels of granularity.
 \end{myitemize}

 The rest of this paper is organized as follows: In Section~\ref{Related Work}, we review the related work of image caption datasets, image caption and image-text retrieval. The construction procedure of our dataset are described in Section~\ref{Dataset}. Then, we provide the details of our dataset in Section \ref{Method}. In Section~\ref{Experiment}, the evaluation of baseline image retrieval methods under different experimental settings are given. Finally, we draw some conclusions with several ways for further improving LuojiaHOG in Section~\ref{Conclusion}.
	\section{Related Work}
\label{Related Work}
\begin{table}[!t]
	\caption{Comparison of current datasets.}
	\renewcommand\arraystretch{1.5}
	\vspace{-1em}
	\begin{center}
		\resizebox{1\columnwidth}{!}{
			\begin{threeparttable} 
			\begin{tabular}{cccc}
				\hline
				Dataset&Classes/Images&Geographic area&Classification system\\
				\hline
				Sydney-Captions~\cite{qu2016deep} &7/613&Sydney&fixed\\
				UCM-Captions~\cite{qu2016deep}&21/2,100&UC Merced&fixed\\
				RSICD~\cite{lu2017exploring}&30/10,921& - & fixed \\
				RSITMD~\cite{yuan2022exploring}& 32/4,743& - &mixed\\
				NWPU-Captions~\cite{cheng2022nwpu}& 45/31,500&global&mixed\\
				RS5M~\cite{zhang2023rs5m} & -/5 million&global& - \\
				RSGPT~\cite{hu2023rsgpt}&-/2,585&multi-cities& -\\						 
				\hline
				\textbf{LuojiaHOG(Ours)}&\textbf{131/94856}&global sample&extensible\\
				\hline
			\end{tabular}
			\begin{tablenotes}    
				\footnotesize               
				\item[1] 
					\textbf{Fixed} classification system (CS) is usually constructed according to expert experience.\\ \textbf{Mixed} CS adds some new labels based on fixed CS.\\ \textbf{Extensible} represents a complete CS standard which can be expanded according to diffenrent task requirements.
			\end{tablenotes}            
		\end{threeparttable}      
		}
	\end{center}
	\vspace{-2em}
	\label{datasets_summary}
\end{table}
\textbf{Image Caption Datasets.} 
Considerable efforts have been directed towards advancing benchmark datasets and novel caption techniques in the remote sensing domain. For instance, Qu et al. ~\cite{qu2016deep}introduced a pioneering deep multimodal neural network model alongside two benchmark datasets, Sydney-Captions~\cite{qu2016deep} and UCM-Captions~\cite{qu2016deep}. Their model ingeniously combined different CNNs with RNN/LSTMs to enhance performance. UCM-Captions includes 2,100 images of 21 categories, each of which is 256\texttimes256 pixels. The data, based on UC Merced Land Use Dataset~\cite{yang2010bag}, were extracted from urban area images of the National Map of the United States Geological Survey. Whereas Sydney-Captions, contains 613 images of 7 categories, which were collected from Sydney, Australia. Both datasets offer 5 descriptions for each image. Building upon this work, Lu et al.~\cite{lu2017exploring} and Cheng et al.~\cite{cheng2022nwpu} conducted a comprehensive analysis of the challenges associated with RS image captioning. They further contributed to the field by creating a larger benchmark dataset separately known as RSICD and NWPU-Captions, aimed at generating more precise and adaptable descriptions. NWPU-Captions, based on NWPU-RESISC Dataset~\cite{cheng2017remote}, encompasses 31,500 images along with 157,500 captions of 45 categories. RSICD comprises 10,921 images and 54,605 captions, with 24,333 of these being unique captions. Subsequently, numerous enhanced approaches have emerged, each carefully tailored to the unique characteristics of RS images. Yuan et al. ~\cite{yuan2022exploring} used manual annotation to construct a fine-grained and more challenging Remote Sensing Image-Text Match dataset (RSITMD) to address the problem of excessive repetition of text descriptions in traditional RS image-text dataset. RSITMD selects 4,743 images from RSICD and provide 23,715 captions. One particularly effective strategy involves the incorporation of diverse attention mechanisms into the standard encoder-decoder architecture. Notably, some of these methods\cite{li2023blip}\cite{zhu2023minigpt}\cite{liu2023visual} have demonstrated promising performance improvements in image caption. The RS5M dataset, a recent creation by Zhang et al.~\cite{zhang2023rs5m}, stands out as the most extensive RS image-text pairing dataset available to date. It was meticulously curated by filtering existing publicly available image-text paired datasets and leveraging a pre-trained VLM specifically fine-tuned for RS datasets, utilizing only subtitle labels. RS5M collects 5 million data from 11 publicly available image-text paired datasets~\cite{schuhmann2022laion}~\cite{schuhmann2021laion}~\cite{sharma2018conceptual}~\cite{srinivasan2021wit}~\cite{kakaobrain2022coyo-700m}~\cite{changpinyo2021conceptual}~\cite{desai2021redcaps}~\cite{thomee2016yfcc100m}~\cite{ordonez2011im2text} and 3 large-scale RS image classification dataset~\cite{sumbul2019bigearthnet}~\cite{christie2018functional}~\cite{long2021creating}. Motivated by the impressive image and text comprehension capabilities of VLMs, Hu et al. ~\cite{hu2023rsgpt} embarked on the creation of the Remote Sensing Image Captioning dataset (RSICap). This dataset collected 2585 image-text pairs that have been carefully annotated by professionals. Each image corresponds to a sentence that describes in detail the attributes of the features in the image. They also provided an evaluation dataset (RSIEval) dataset that can be used for the evaluation of domain-specific or general VLMs. RSIEval consists of 100 human-annotated captions and 936 visual question-answer pairs with rich information and open-ended questions and answers. There work serves as a valuable resource, designed to support the development of robust vision language models within the remote sensing domain. In Tab.\ref{datasets_summary}, we give statistics of existing image caption datasets together with LuojiaHOG.

\textbf{Image Caption. } Although the access to remote sensing images is getting easier, how to quickly obtain detailed and accurate text descriptions of remote sensing images is still a problem. For this reason, a large research effort has been devoted to image captioning, i.e. the task of describing images with syntactically and semantically meaningful sentences. For sentence generation, the studies has developed from traditional template-based and retrieved-based methods to Recurrent Neural Network (RNN) and LLM. Template-based methods generate descriptive sentences for a given image through fixing templates with a number of blank slots. In these approaches, different objects, attributes, actions are detected first and then the blank spaces in the templates are filled. Farhadi et al.~\cite{farhadi2010every}use a triplet of scene elements to fill the template slots for generating image captions. A Conditional Random Field (CRF) is adopted by Kulkarni et al. ~\cite{bybaby} to infer the objects, attributes, and prepositions before filling in the gaps. Retrieval-based approaches first extracted a candidate caption set from a set of caption pool with a basic retrieval (pre-retrieval) model. The final best-matching captions for the input image are then chosen from the captions pool by the re-ranking method. For example, Hodosh et al. ~\cite{hodosh2013framing} treated the image captioning as a ranking or retrieval task, and introduced a ranking-based method to extract image description. Gong et al. ~\cite{gong2014improving}associated the query image with a textual description by projecting them into a shared latent space. Although retrieval-based methods can produce syntactically correct captions, the retrieved captions are not tailored for the query images and limited by the size of the pre-constructed image-caption repository. Motivated by the remarkable success of deep neural networks in CV and NLP, the seq2seq paradigm has become the mainstream in image captioning. Attention mechanisms play an essential role in enhancing the performance of the seq2seq models. For example, an attentive seq2seq model was introduced in ~\cite{xu2015show}, which learned to dynamically attend to different locations of the query image at different decoding step. Mun et al.~\cite{mun2017text} used associated captions that were retrieved from training data to learn visual attention for image captioning. Besides, Yang et al.~\cite{yang2020ensemble}focused on the improvement of both retrieval- or generation-based model by using a dual generator generative adversarial network with two generators and one discriminator. With the rapid development of LLMs in recent years, VLM that combines vision and language, has been recently introduced and demonstrated several impressive capabilities of vision-language understanding and generation. Flamingo~\cite{alayrac2022flamingo}, for instance, integrates visual adaptation layers into an LLM and is trained on a large-scale interleaved image text dataset. ML-MFSL~\cite{najdenkoska2023meta} is similar to Flamingo, where a visual prefix is introduced as a learnable feature to extract information related to text from the image. After enhancing the visual prefix with the meta mapper network and concatenating it with textual features, LLM is employed to predict the responses. BLIP-2 \cite{li2023blip}utilizes multiple vision-language losses to align visual features with text via the Q-Former model, and tunes a simple fully connected layer to feed the queried embedding to a frozen language model. Based on BLIP-2 , MiniGPT4 \cite{zhu2023minigpt}and InstructBLIP \cite{instructblip} retain the Q-Former model, replace the language model with a larger one, and fine-tune on meticulously collected instruction data. In addition, simpler and more direct methods, such as LLaVA \cite{liu2023llava}, directly feed visual features to the LLM using only a learnable fully connected layer. RSGPT utilizes high-quality RS image and text pairs and fine-tunes them on the basis of minigpt4 to obtain a RS image caption model. These image caption models can obtain corresponding text descriptions for images, but the quality of text generation will be limited by the LLM model.

\textbf{Image-Text Retrieval} Image-Text retrieval from RS big data refers to finding RS images/descriptions that satisfies a text description/ remote sensing image from large RS image collections. Thanks to the prosperity of deep models for language and vision, we have witnessed the great success of image-text retrieval over the past few years. Frome et al. ~\cite{frome2013devise}firstly encoded image and text features independently for image-text retrieval. Afterwards,
a stream of works~\cite{wu2019learning,chun2021probabilistic} tries to excavate the high-order data information for learning powerful features. Wang et al.~\cite{wang2018learning}proposed a maximum-margin ranking loss with the neighborhood constraints for better extracting features. Lee et al. ~\cite{lee2018stacked}made the first attempt to consider the dense pairwise cross-modal interaction and yielded tremendous accuracy improvements at the time. Jia et al. \cite{jia2021scaling} tended to learn image-text representation by scaling up the dataset with some noise. As a milestone, OpenAi~\cite{radford2021learning} proposed a large vision language model CLIP, which achieved amazing results in retrieval tasks. Yao et al. ~\cite{yao2021filip} conducted more fine-grained image-text matching research based on CLIP. On the basis of fine-grained image-text matching, Gu et al.\cite{gu2022wukong} introduced a token reduction layer to further improve the retrieval capabilities of this type of method. Li et al.\cite{li2022blip} and Li et al.\cite{li2021align} explored the fusion of visual and textual features and add a classification head to determine whether the image-text pairs match. In remote sensing, Yuan et al. ~\cite{yuan2022exploring} introduced an asymmetric multimodal feature match network to extract multi-scale features. Yuan et al. \cite{yuan2022remote} fused multi-level image features and added a multivariate rerank algorithm to improve the retrieval performance. In view of the great success of CLIP, Zhang et al. \cite{zhang2023rs5m} integrated large-scale remote sensing (RS) and computer vision (CV) datasets, specifically screening remote sensing images for pre-training, and developed a RS CLIP model named GeoRSCLIP. These models predominantly adopted dual-encoders to enhance retrieval capabilities, emphasizing dataset scale, fine-grained image-text matching, and fusion of image-text features.

	\section{LuojiaHOG Dataset}
\label{Dataset}
\subsection{Dataset Construction}

\begin{figure}[htbp]
	\centering
	\includegraphics[width=0.48\textwidth]{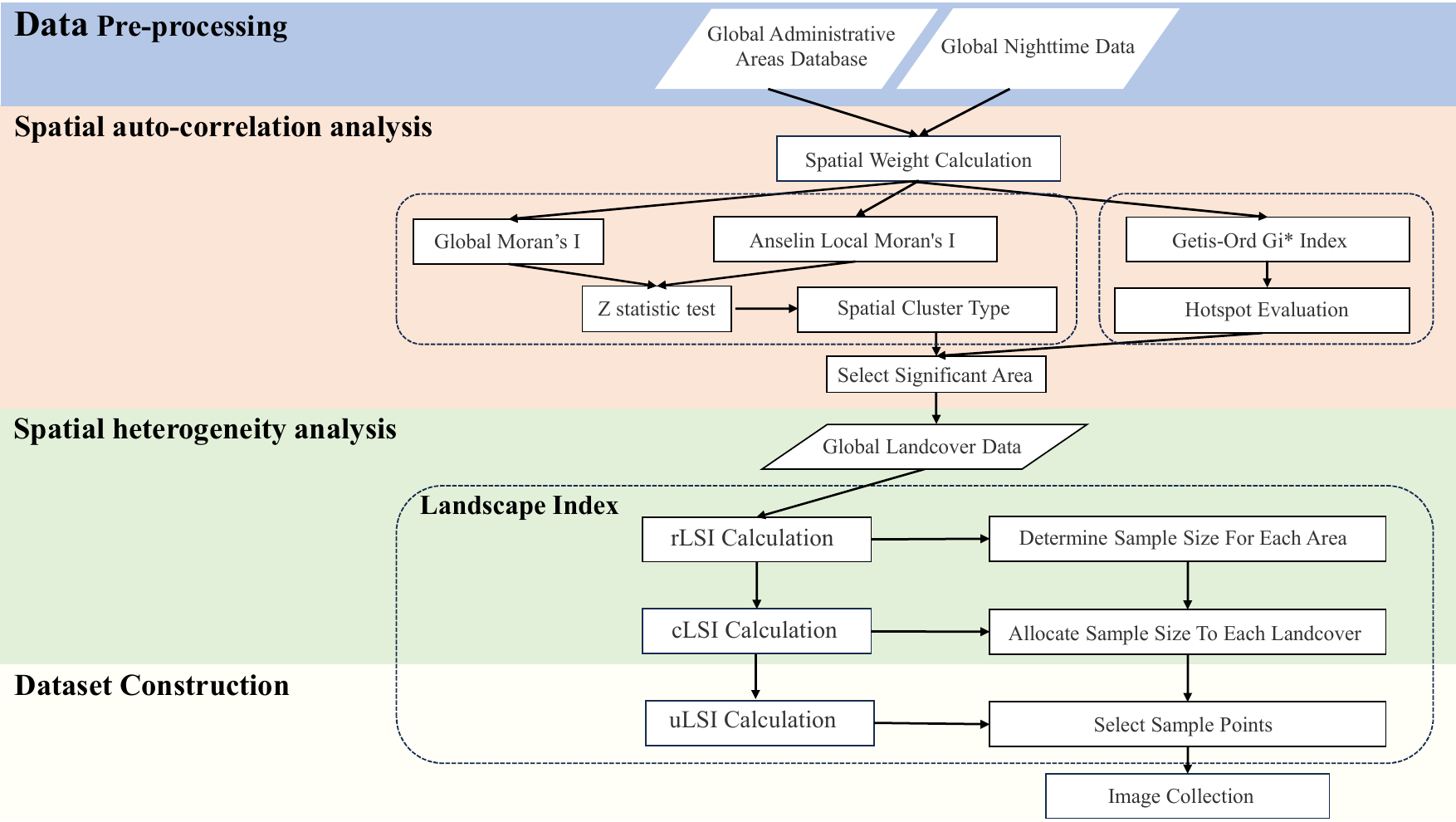}
	\caption{Flowchart of sampling method.}
	\label{flowchart}
	\vspace{-1em}
\end{figure}

Four example descriptions of a sample scene are depicted in Fig.~\ref{example}. This dataset is sourced from Google Maps and OpenStreetMap (OSM). Google Maps contributes an extensive collection of remote sensing images, while OSM offers a wealth of comprehensive geographical information. As shown in Fig~\ref{flowchart}, we firstly acquired global sampling points through spatial analysis and the evaluation of landscape indices in subsection~\ref{Sampling methods}. It allows us to obtain remote sensing images of countries and regions with various topography and different economic levels. Next, we built an extensible classification system and integrated the obtained OSM labels into this classification system in subsection~\ref{classification system}. Finally, we adopted a variety of annotation strategies and dataset enhancement methods to generate text descriptions and construct final image caption dataset from the collected images and labels in subsection~\ref{annotation}.

\subsubsection{Hierarchical sampling method.}\label{Sampling methods}
\textit{Spatial auto-correlation for sampling area.}
The Moran's I and Getis-Ord Gi* Index are widely-used methodologies for spatial auto-correlation analysis, which can help select globally representative regions to optimize the subsequent sampling procedure. In our approach, Moran's I is used to distinguish global regional development patterns, while G* further focuses precisely on regions where hotspots and coldspots exist. The calculation of the Moran's I and Getis-Ord Gi* Index contains two parts, analysis data and spatial weights. For analysis data, we employed global nighttime data due to its capacity to depict urbanization levels. According to Tobler's First Law of Geography, everything is related, but similar things are more closely related. It explains that spatial locations are involved in the spread of objects or actions. Thus for spatial weight $W$, we adapt the weights originally proposed by Moran (1950) and specify the neighborhood as follows. Regions $i$ and $j$ are viewed as‘ neighbours’ if they share the boundary or node, which is represented by $\odot$. Thus, when $i \neq j$, weight $w_{i, j}$ indicates whether i and j are neighbors in space. The spatial weight $w_{i, j}$ is formulated as follows:
\begin{equation}
	w_{i j}= \begin{cases}\|i-j\|^{-\gamma}, & \text { if } i \odot j \\ 0, & \text { otherwise }\end{cases},
\end{equation}
which reflects the degree of connection between region $i$ and $j$.

\textit{Moran's I} 
contains global Moran's I and local Moran's I (Anselin Local Moran's I). The positive Moran's I denotes a positive spatial correlation, with a larger value signifying a more pronounced spatial correlation. Conversely, the negative Moran's I signifies a negative spatial correlation, with a smaller value indicating greater spatial dissimilarity. 
The Global Moran's I assesses the pattern of a dataset spatially and determines if it is dispersed, clustered, or random based on the locations and values of the analysis data. It is calculated using the below formula,
\begin{equation}
	I_{global}=\frac{n}{S_0} \frac{\sum_{i=1}^n \sum_{j=1}^n w_{i, j} z_i z_j}{\sum_{i=1}^n z_i^2}\label{1},
\end{equation}
where $z_{i}$ is the deviation of the nighttime light value $x_{i}$ of region $i$ from its average value. $S_{0}$ is the aggregation of all spatial weights,
\begin{equation}
	S_0=\sum_{i=1}^n \sum_{j=1}^n w_{i, j}\label{2}.
\end{equation}\par
The range of the Global Moran's I is between 1 and -1. When $I_{global}$  approaches 1, it suggests that the pattern observed is clustered spatially, while the opposite indicates dispersion. When $I_{global}$ is close or equal to zero,  it signifies the absence of auto-correlation. The final conclusions about the observed pattern are drawn only after looking at the $z$-score and the $p$-value of the Index. Only when there is clustering or dispersion in the study region, the Local Moran's I $I_{local}^i$ in region $i$ is calculated to further determine regional spatial clustering patterns of all regions around the world,
\begin{equation}
	I_{local}^i=\frac{z_{i}}{S_i^2} \sum_{j=1, j \neq i}^n w_{i, j}z_{j}\label{3},
\end{equation}
where $n$ is the total number of regions and the function of $S_{i}$ is as follow,
\begin{equation}
	S_i^2=\frac{\sum_{j=1, j \neq i}^n\left(x_j-\bar{X}\right)^2}{n-1}\label{4}.
\end{equation}

In Eq.~\ref{3}, $z_{i}$ reflects the level of economic development of the region $i$ and the average level of the entire region. $\sum_{j=1, j \neq i}^n w_{i, j}z_{j}$ is referred to Local indicators of spatial association (LISA), reflecting the level between the surrounding regions of the region $i$ and the level of the entire region.\\
\textit{Getis-Ord Gi* Index}
is used to identify clusters of high or low-value elements in space and determine whether they possess significant statistical significance. By examining each region within its neighborhood, it helps establish whether high-value features have statistical significance. The comparison involves evaluating the local against the overall value, and if a substantial disparity exists, it signifies the presence of a hotspot. The model is formulated as follows,
\begin{equation}
	G_i^*=\frac{\sum_{j=1}^n w_{i, j} x_j-\bar{X} \sum_{j=1}^n w_{i, j}}{S \sqrt{\frac{\left[n \sum_{j=1}^n w_{i, j}^2-\left(\sum_{j=1}^n w_{i, j}\right)^2\right]}{n-1}}},
\end{equation} 
where $x_{j}$ is the nighttime light value of region $j$, $\bar{X}$ is the average nighttime light value of the whole region, the function of S is as follows,
\begin{equation}
	S=\sqrt{\frac{\sum_{j=1}^n x_j^2}{n}-(\bar{X})^2}.
\end{equation} \par
If $G_i^*$ is greater than 0 and the higher $G_i^*$ is, the high values of the target object are clustered more tightly (hot-spots). Oppositely, the low values of the target object are clustered more tightly (cold-spots).\par
Sampling area procedure is as follows,
\begin{equation}
	r=\left(M(D)^{+} \cap G(D)\right) \cup M(D)^{-},
\end{equation}
where $M(\cdot)^{+}$ represents high-high and low-low region calculated based on Local Moran'I and $M(\cdot)^{-}$ is the high-low and low-high region. $G(\cdot)$ represents the hot-spots and cold-spots calculated based on Getis-Ord Gi* Index, $D$ is the global nighttime light data, $r$ is the selected significant regions.\par
\textit{Spatial heterogeneity sampling points. }
Spatial sampling design is the key steps in building a dataset, and many traditional sampling methods may not achieve credible sampling due to the high spatial heterogeneity of land cover. Landscape index (LSI) is the ratio of landscape perimeter to region within a certain range, which quantitatively represents the landscape heterogeneity of the region. For raster data,
\begin{equation}
	L S I=\frac{1}{4} \sum \frac{b_p}{\sqrt{q}}(p=1,2, \ldots, q),
\end{equation}
where $q$ is the number of pixels, $b_{p}$ represents the number of four neighborhood pixels belonging to different classes than pixel $p$. Guided by  Chen et al. \cite{chen2016landscape}, for a given region $i$, we use LSI as three levels to characterize the spatial heterogeneity: rLSI for regional sampling points, cLSI for land cover classes under such region and uLSI for each geographic sampling unit. Following their method, corresponding sample sizes and their spatial distributions according to landscapes classes can be determined.

In order to enable more heterogeneous regions with higher sample density and larger sample size, the number of regional sampling points $N_{i}$ in region $i$ is determined by $rLSI$. 
\begin{equation}
	N_i=\frac{r L S I_i \times A_i}{\sum_{j=1}^{j=n} r L S I_j \times A_j} \times N,(i=1,2, \ldots, n),
\end{equation}
where $A_{i}$ and $A_{j}$ represent the areas of region $i$ and $j$ respectively, $N$ is the total sample size, and $n$ is the total number of regions.

Subsequently, the $cLSI$ represents the spatial variability in land cover classes. A class with a lager $cLSI$ has a more complex spatial distribution and higher spatial heterogeneity; thus, more samples are allocated. For the number of samples of class $k$ in region $i$, the sample number $cN_{i,k}$ is as follows,
\begin{equation}
	c N_{i, k}=\frac{c L S I_{i, k} \times W_{i, k}}{\sum_{k=1}^{k=m} c L S I_{i, k} \times W_{i, k}} \times N_i,(k=1,2, \ldots, m),
\end{equation}
where $N_{i}$ is the number of regional sampling points, $W_{i,k}$ is the proportion of category $k$ in region $i$, and $m$ is the total number of categories.

Lastly, $uLSI$ adaptively selects the sample point location. Suppose that region $i$ can be divided into $R\times L$ geographical units. In each geographical unit, the $uLSI$ of class $k$ in row $r$ and column $l$, ${ }_k^i u L S I_{a, b}$, is calculated.
\begin{equation}
	u L S I={ }_k^i u L S I_{a, b}.
\end{equation}

A distribution curve $C_k^i$ depicts the heterogeneity of each unit ranked from large to small. The x-axis is the geographical unit coordinate, and the y-axis is ${ }_k^i u L S I_{a, b}$. Firstly, remove the part of $C_k^i$ where the values are equal to zero. Then divide $cN_{i,j}$ equal parts on the x-axis. Finally, sampling points are randomly selected in each interval. 

\subsubsection{Extensible classification system construction. }\label{classification system}

\begin{figure}[htbp]
	\centering
	\captionsetup{justification=justified}
	\includegraphics[width=0.4\textwidth]{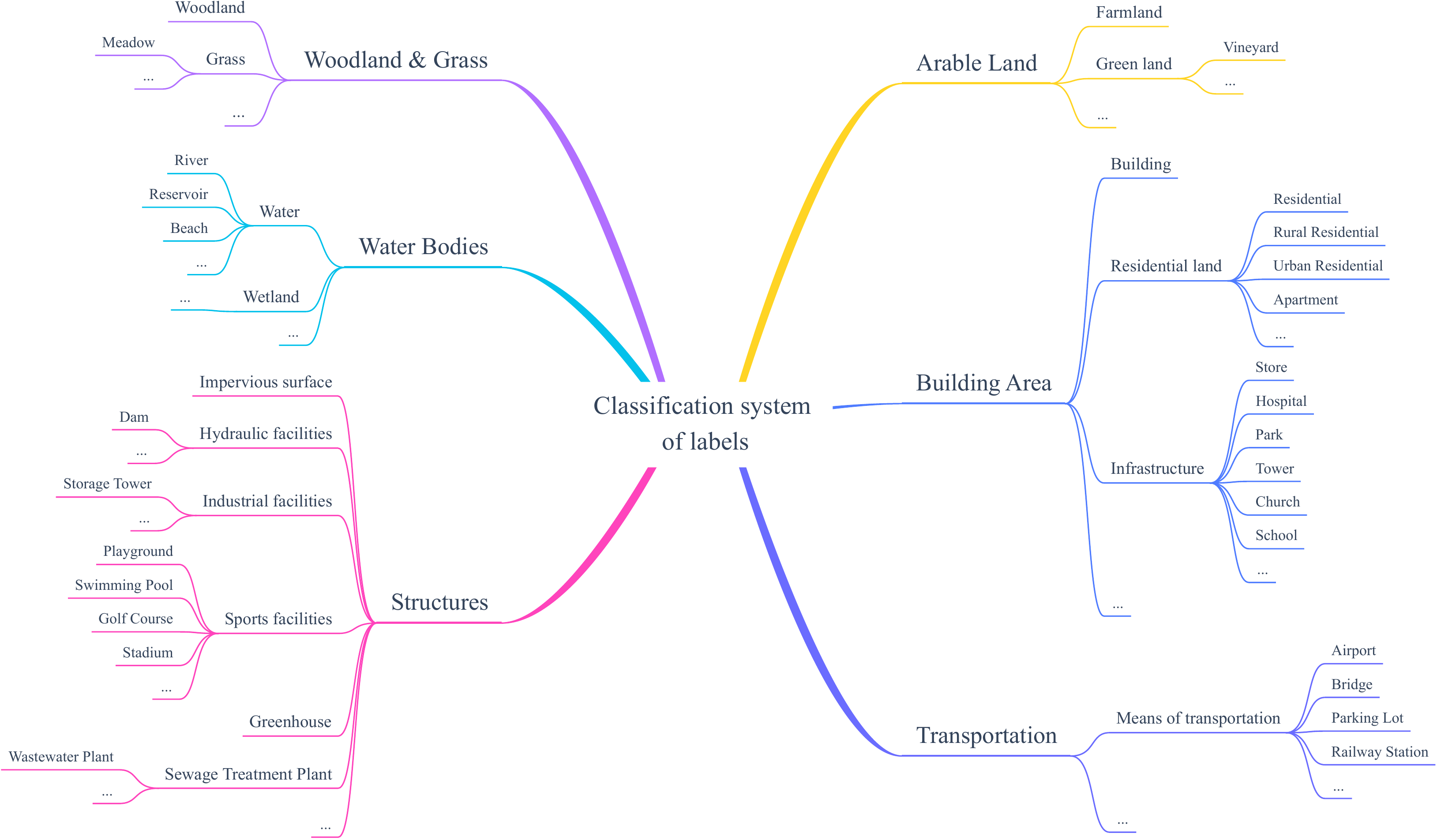}
	\caption{Classification system of dataset: there are 7 first-level labels, 21 second-level labels for each first-level labels, and 131 third-level labels to describe more detailed type. The figure shows part of the classification system.}
	\label{classification system pic}
\end{figure}

Firstly, we adopt the OGC-based classification system to solve the issues of different existing classification systems in terms of category naming, category hierarchy, category semantics and compatibility. In OGC-based classification system $T$ \cite{cao2023large}, all categories are hierarchically organized in a three-level tree: third-level labels $T_{3}$ fall into second-level labels $T_{2}$ and then grouped into first-level labels $T_{1}$, which is the highest level. We utilize $\prec$ to represent the low-level label belongs to high-level label. For OSM labels set $L$, we select labels that cannot directly correspond to the classification system $L^{-}$ for processing.
\begin{equation}
	L^{-} = \{l | l \in L  \quad and \quad l \notin T\},  
\end{equation}
\begin{equation}
	T = T_{1} \cup T_{2} \cup T_{3}, 
\end{equation}
\begin{equation}
	\mathrm{T}_2=\bigcup_{\mathrm{i}}\left\{\mathrm{t}_2 \prec \mathrm{t}_1^{\mathrm{i}} \mid \mathrm{t}_1^{\mathrm{i}} \in \mathrm{T}_1, \mathrm{t}_2 \in \mathrm{T}\right\}, 
\end{equation}
\begin{equation}
	\mathrm{T}_3=\bigcup_{\mathrm{i}}\left\{\mathrm{t}_3 \prec \mathrm{t}_2^{\mathrm{i}} \mid \mathrm{t}_2^{\mathrm{i}} \in \mathrm{T}_2, \mathrm{t}_3 \in \mathrm{T}\right\}.
\end{equation}
Through the establishment of  principles for the inclusion of novel labels, the consolidation of duplicated labels, and the execution of label mapping, we construct the final classification system denoted as $T^*$ in Fig.~\ref{classification system pic}. Unlike fixed and mixed classification systems, the extensible classification system can be updated through novel labels inclusion, duplicated labels consolidation and label mapping.

\textit{Novel labels inclusion. }
For $l \in L^{-}$, a top-down strategy is adopted to add it into the $T$. We analyze the category $l$ belongs to from the OSM classification system, and perform a semantic search starting from the first-level label $T_{1}$ by using LLMs. As we confirm that $t \in T_{n}$ is the best matching label, we compare the function (OSM descriptions) of $l$ with the the candidate label set $N\{n \prec t | n \in T\}$ to judge whether there is a relationship with the label in $N$. If not exists, $l$ is added to $T_{n+1}$; otherwise, continue searching until no relationship exists. For instance, the description of 'farmyard' in OSM is: '\textbf{\textit{buildings}} for keeping animals, or crop supplies would typically be part of a farmyard tagged landuse=farmyard.' Therefore second-level label 'building' is the best matching label. According to the judgment of experts and LLMs, 'farmyard' has no relationships with the existing candidate labels of 'building', so it is determined that 'farmyard' can be added as the third-level label belongs to 'building'. Another example is 'restaurant', which belongs to 'amenity' according to OSM. Through semantic comparison, it is found that 'amenity' and 'infrastucture' have the same meaning, and 'restaurant' can be added as the third-level label belongs to 'infrastucture' following above rules. If possible, forth-level labels can be added in our classification system.

\textit{Duplicated labels consolidation. }
For some common synonyms found in OSM labels, such as "cemetery" and "graveyard" or words with different spellings like "reservior" and "reservoir", we perform the first merging step by asking LLMs about the synonyms for each geographical objects. To avoid potential omissions in the first step, we further merge similar words based on the function of geographical objects. By consulting the descriptions of labels on OSM and then querying LLMs to compare the function with the labels in the existing classification system, we select possible duplicate labels. Finally, human inspection of the label merging results is conducted to prevent errors in merging and potential omissions.

\textit{Label mapping. }
Ultimately, we perform statistics of the labels associated with all the images within the dataset. Subsequently, we curate a sub-classification system labeled as $T^*$ and determine the categories for the final dataset by selecting those labels with a frequency exceeding zero.
\subsubsection{Detailed caption generation. }\label{annotation}

\indent With the collected images for a specific interpretation task, annotation is performed to assign specific semantic labels to the content of interest in the images. In this step, we adopt both professionally manual and automatic annotation to generate corresponding text descriptions for each image.\par
\begin{figure}[htbp]
	\centering
		\captionsetup{justification=justified}
	\includegraphics[width=0.48\textwidth]{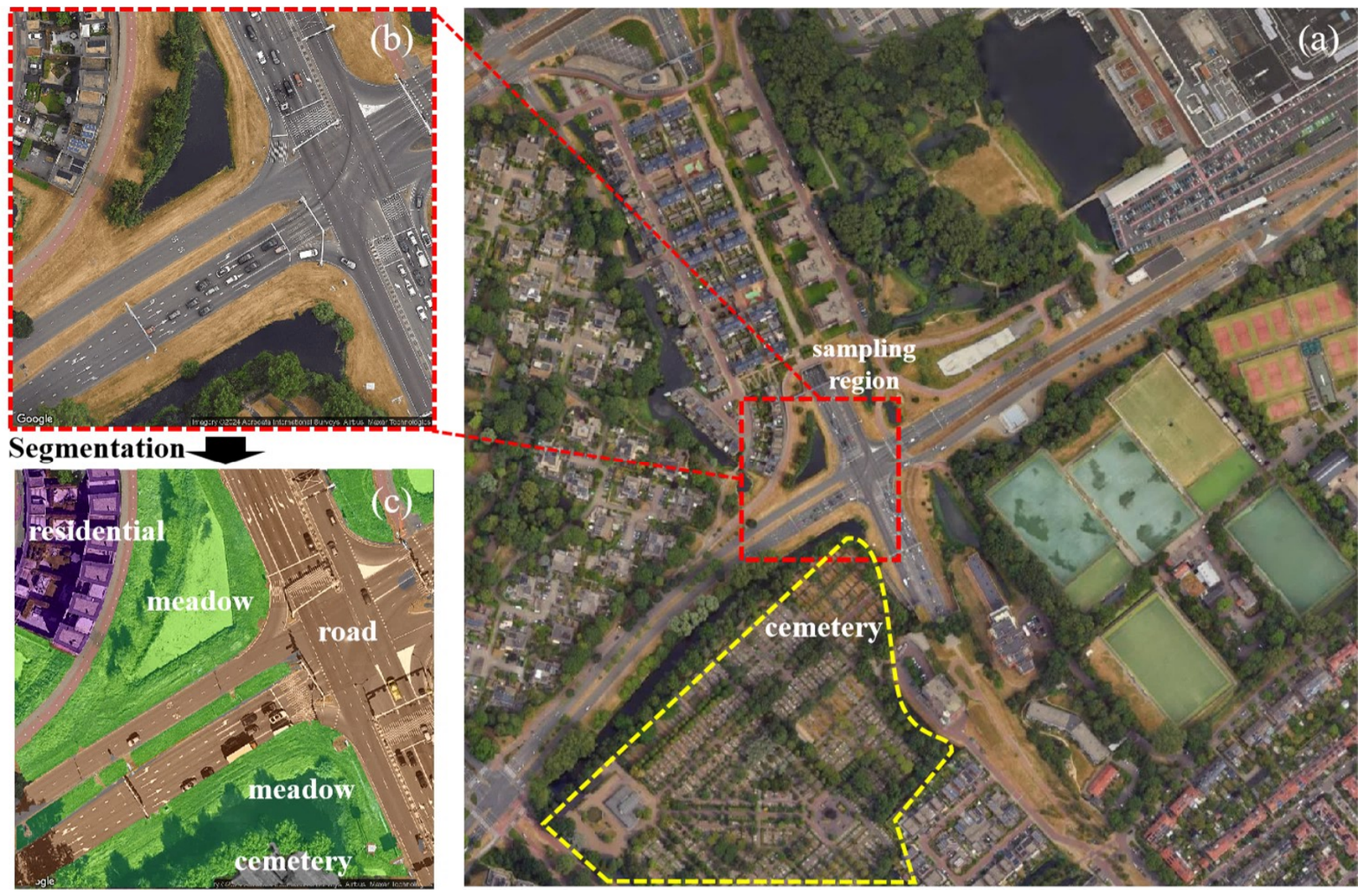}
	\caption{An example illustrates the scope problems in RS images. As only a small corner of cemetery is captured in the sampled image, it should be excluded from the labels.}
	\label{label problem}
\end{figure}
\textit{Manual Annotation. }In practice, constructing a large-scale image dataset by manual scheme is laborious and time-consuming. To relieve this problem, crowd-sourcing annotation becomes an alternative solution that can be employed to create a large-scale image dataset \cite{lin2014microsoft,christie2018functional} while paying efforts to its challenge with quality control. 
Therefore, annotators with rich experience in remote sensing annotation manually correct the OSM labels corresponding to all images in dataset. In addition, inspired by previous effort\cite{hu2023rsgpt,lu2017exploring}, we acquire accurate descriptions of dataset through the fully supervised annotation process.\par
The rectification of image labels primarily deals with two prevalent issues. The first arises due to the scope problems in RS images, while the second emanates from inaccuracies present within the crowdsourced data. RS images only contain a small part of objects, causing the labels to be discarded. As illustrated in the Fig.~\ref{label problem}, there is a cemetery in this area according to OSM labels. However, the collected image (in red box) contains a small corner of the cemetery, so the 'cemetery' needs to be removed from the image labels. To solve this type of problem, professional annotators are required to refer to the original map image to determine whether there are features that have been mistakenly added to the label because a small part of it is included in the image. Furthermore, as OSM is crowd-sourced data, the lack of professional annotations may lead to potential errors or outdated labels. Annotators need to visually inspect the images, remove clearly erroneous labels, and fill in any obviously missing labels.\par
\begin{figure*}[t]
	\centering
	\includegraphics[width=\textwidth]{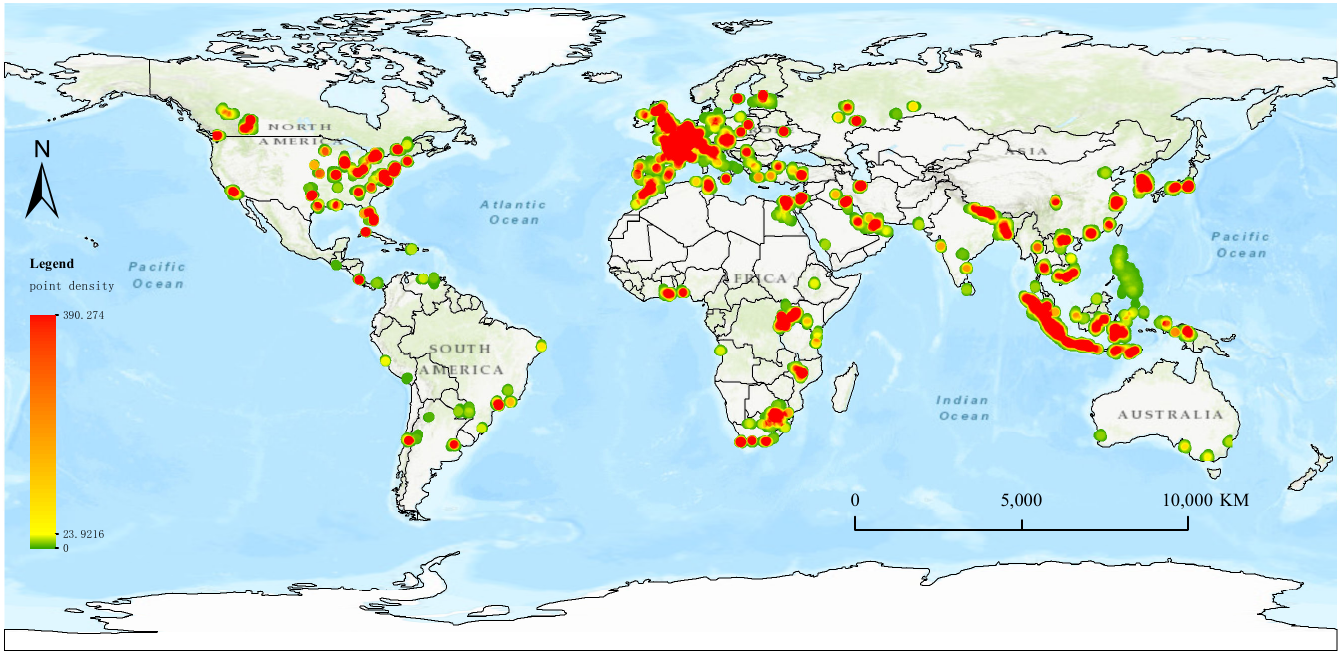}
	\caption{Distribution of sampling points around the world. }
	\label{sampling}
\end{figure*}

Remote sensing images contain numerous geographical objects, each with distinct attributes and interrelationships with other objects. Therefore, it is imperative to establish specific guidelines for standardizing text descriptions. In the course of formulating guidelines, we take the previous work as a reference\cite{chen2015microsoft, hu2023rsgpt, lu2017exploring}. The final annotation procedure follows the principles of: (1) describing object attributes, including color, shape, size, relative position between objects and special symbols (such as character 'H' for Helipad). (2) reducing vague words, such as using specific numbers to replace words like many, some, etc. for countable objects. (3) using words like ‘near’ and ‘next to’ to replace direction, such as up, down, left, right, since the remote sensing images are aerial view. (4) generally, the annotation process involves first describing the main objects (occupying most of the image), followed by describing detailed objects (5) adding some synonym substitutions to reduce duplication\par 
Additionally, unlike previous work that described each image in five sentences, we do not impose any restrictions on the number of sentences and only require that the image can be fully described. The manual annotation can be formulated as follows,
\begin{equation}
	Desc=annot(I,\ rect(I,L)),
\end{equation}
where $I$ is the image, $L$ is the corresponding labels and $Desc$ is the image descriptions. $rect(\cdot)$ represents the annotator's correction of OSM labels based on image. $annot(\cdot)$ represents the procedure annotator following to describe the image.

\textit{Automatic Annotation}
Although relatively high-quality datasets can be obtained using manual annotation, its time-consuming and labor-intensive characteristics are not suitable for large-scale image text generation in today's era of remote sensing big data. In this step, we use image-based text generation methods to automatically get the description of the image separately with carefully designed prompts to boost performance.\par
With the emergence of various powerful VLMs, it has become feasible to automatically generate a large number of accurate image descriptions. 
Referenced to \cite{hu2023rsgpt}, Minigpt4\cite{zhu2023minigpt} has very powerful capabilities in remote sensing image caption tasks. In view of the problems existing in the VLMs in terms of details, position and hallucination, we designed different prompts to improve the generated results. Finally, we randomly sampled 10 percent of the generated texts to evaluate the generation quality to ensure the quality of the generated text.\par
\begin{figure}[t]
	\centering
	\captionsetup{justification=justified}
	\subfigure[The PDF of caption length.]{\includegraphics[width=0.24\textwidth]{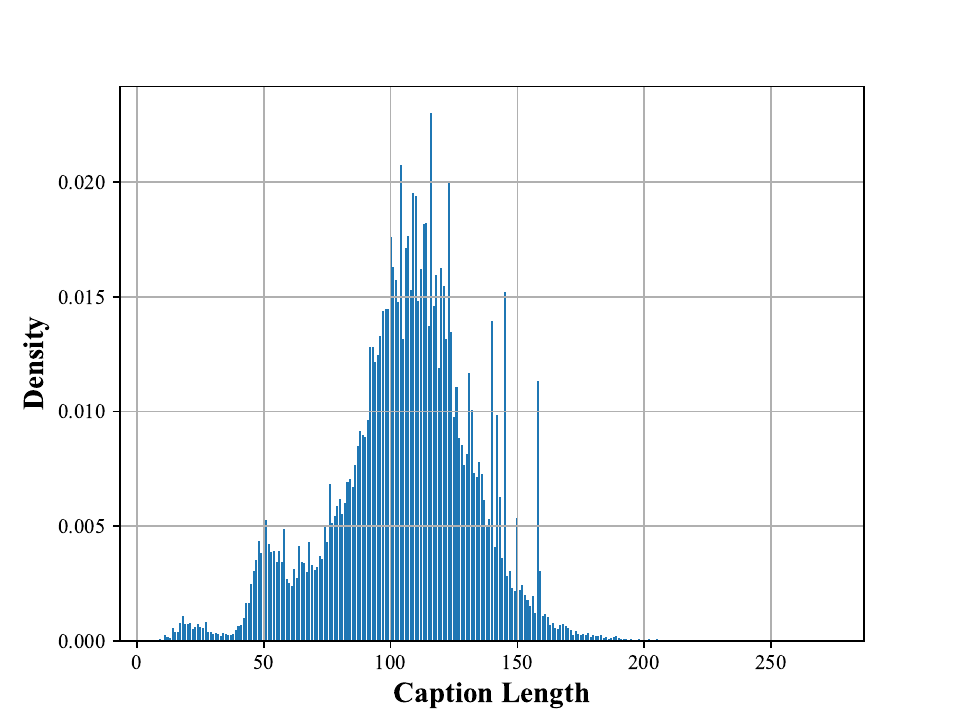}\label{word_distribution}}
	\subfigure[The PDF of sentence length.]{\includegraphics[width=0.2345\textwidth]{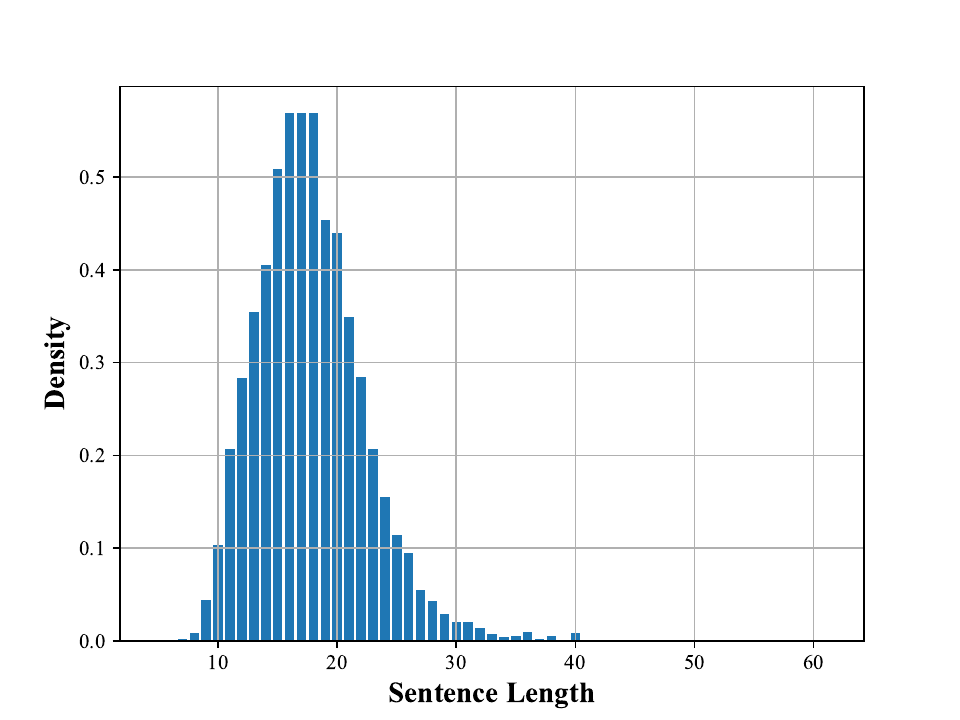}\label{sentence_distribution}}
	\caption{The probability density function (PDF) visualization on LuojiaHOG}
	\label{pdf}
	\vspace{-0.8em}
\end{figure}
Furthermore we adopt prompt engireering to improve caption quality. We design prompts from the following aspects: direct task specification, task demonstration, memetic proxy, constraining behavior. A direct specification consists in constructing a signifier for the task, which is a pattern for the intended behavior. We designed some templates to constructing the signifier. For example, we set the task to provide a text description of the features contained in a RS image. In task demonstration, formulating guidelines mentioned in manual annotation is adopted. Since Few-shot examples are effective for task specification, some description examples are added to help LLMs better understand the task. Specification by memetic proxy is mechanistically similar to direct specification, which specifies intended tasks from memespace/cultural consciousness. LLMs' ability to create simulations of well-known figures and to draw on cultural information far exceeds the ability of most humans. Therefore, we allow LLMs to play the role of professional annotators, experienced remote sensing scientists, etc., so that LLMs can more accurately understand the task targets. Lastly, in order to make the generated text more suitable for remote sensing images and reduce unreasonable descriptions, we impose constraints in terms of word count, content elimination, etc.

\subsection{Dataset Statistics and Analysis}\label{DatasetStatistics}
In this section, we perform a thorough analysis of data statistics and visual examination, focusing on sampling diversity, labels quantity, and descriptions granularity.

\textit{Sampling diversity. } We collected images from Google Earth with different resolutions from all over the world. The size of images is 1280 $\times$ 1280 and the total number is 94856. Images in dataset are actually multisource, as Google Earth images are from different remote imaging sensors. Fig.~\ref{sampling} shows the distribution of sampling points in a global level. \par

\begin{table}[!t]
	\centering
	\caption{Statistical indicators of the LuojiaHOG dataset.}
	\captionsetup{justification=justified}
	\renewcommand\arraystretch{1.5}
	\resizebox{1\columnwidth}{!}{
		\begin{tabular}{cc}
			\hline
			\textbf{Indicators} & \textbf{Count} \\ \hline
			\textbf{Number of vocabularies} & 10044775 \\
			\textbf{Number of distinctive vocabularies} & 14128 \\
			\textbf{Number of sentences} & 565231 \\
			\textbf{Average length of captions} & 123.56 \\
			\textbf{Average number of sentences per caption} & 6.95 \\ \hline
			\textbf{Number of images} & 94856 \\ \hline
		\end{tabular}
	}
	\label{statistics table}
	\vspace{-1.5em}
\end{table}

\textit{Labels quantity. }Fig.~\ref{classification system pic} illustrates the classification system of dataset: there are 7 first-level labels (like "Building area" and "Arable land"), 21 second-level labels for each first-level labels (like "Building" and "Infrastucture" in "Building area"), and 131 third-level labels to describe more detailed type (like "Church" and "Cemetery" in "Infrastucture"). As presented in Tab.~\ref{datasets_summary}, the number of labels in our dataset surpasses that of existing image caption datasets.\par

\begin{figure}[ht]
	\centering
	\captionsetup{justification=justified}
	\includegraphics[width=0.48\textwidth]{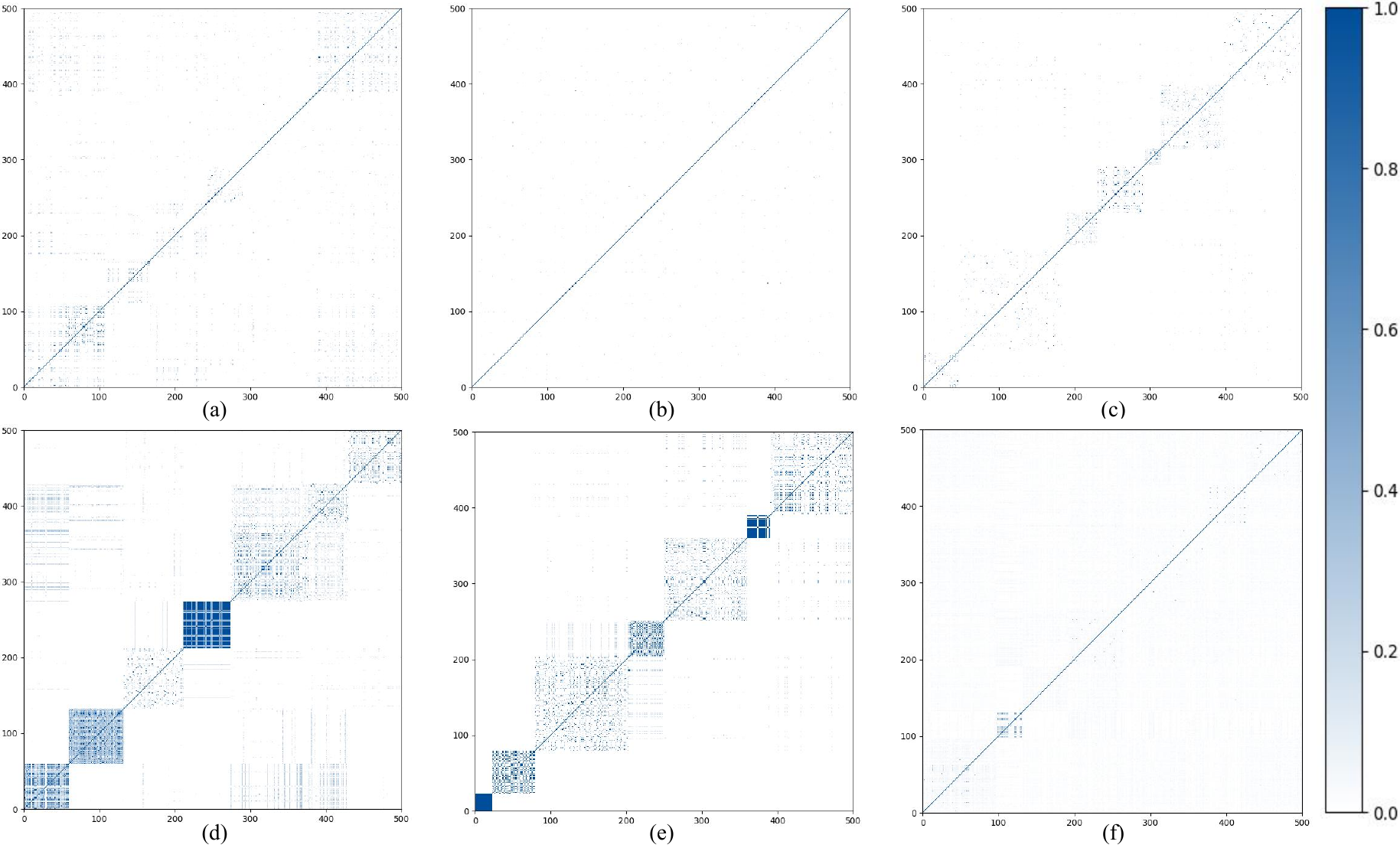}
	\caption{The similarity visualization results of the six datasets, where the similarity scores are weighted by commonly used metrics in the field of natural language processing, BLEU and METEOR. (a): RSICD  (b): RSITMD  (c): NWPU  (d): Sydney  (e): UCM  (f): LuojiaHOG (ours).}
	\label{dataset heatmap}
	\vspace{-1em}
\end{figure}
\textit{Descriptions granularity. } Fig.~\ref{word_distribution} displays the probability density function (PDF) of caption length, which (takes on a shape similar to a normal distribution). The longest caption length contains 188 vocabularies, with an average length of 123.562 vocabularies per caption. Fig.~\ref{sentence_distribution} illustrates the PDF of the sentence length, with the longest containing 35 sentences and an average of 6.953 sentences per caption. Tab.~\ref{statistics table} shows several statistical indicators of dataset, such as the total number of vocabularies in captions being 10,044,775, the number of distinct vocabularies being 14,128, and the total number of sentences being 565,231. The datasets with high inter-text similarity inadequately support retrieval models within the domain of remote sensing. We adopted BLEU and METEOR weighted scores as evaluation metrics to assess the quality of existed datasets as well as our LuojiaHOG. For better comparison, the captions in each dataset are clustered according to the text feature, and then randomly selected from each cluster for evaluation. The visualizaiotn is shown in Fig.~\ref{dataset heatmap}. Compared with other datasets, most of our captions have the similar templates at the beginning or end, such as 'This is an image of...' or 'In conclusion, the image. ..'etc., so there are some light blue (representing very weak correlation) in our results. RSTIMD showed the best results in this evaluation due to its carefully processed and relatively short captions. Overall, our result has the second-least severe chunking effect, only lay behind RSTIMD with carefully processed short captions, which reflects the uniqueness of our captions. Datasets like Sydney and UCM, there is a considerable amount of 'noise' indicative of the high language similarity present within these datasets. 
	 \begin{figure*}[htb]
 	\centering
 	\captionsetup{justification=justified}
 	\includegraphics[width=0.78\textwidth]{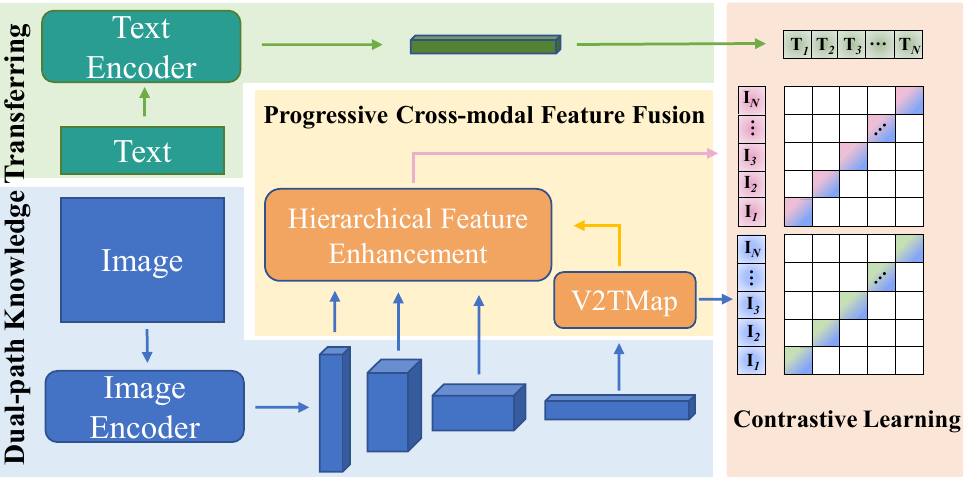}
 	\caption{The framework of CISEN. CISEN employs a \textit{\textbf{dual-path knowledge transfer}} approach for extracting multi-modal features. \textit{\textbf{Progressive cross-modal feature fusion}} contains V2TMap and Hierarchical feature enhancement (HFE). Through \textit{\textbf{V2TMap}}, global visual features are transformed into text-like representations. \textit{\textbf{HFE}} module facilitates the fusion of transformed visual features with multi-level visual features, resulting in the production of enhanced visual features. }
 	\label{CISEN}
 	\vspace{-1.5em}
 \end{figure*}

\section{Our Method}
\label{Method}
Motivated by~\cite{wang2022cris}, we present the proposed CLIP-based Image Semantic Enhancement Network (CISEN) in Fig.~\ref{CISEN}. CISEN  mainly  consists of dual-path knowledge transferring (in subsection~\ref{dual-path}) and progressive feature fusion (in subsection~\ref{feature fusion}). The former mainly transfers the multi-modal knowledge from the large pre-trained vision-language model. The progressive feature fusion consists of two stages, visual to text feature mapping (V2TMap) and hierarchical feature enhancement (HFE), to fuse semantic information from textual features to visual features. The V2TMap utilizes an image adapter to transfer global visual features to textual-like features. The HFE adopts feature pyramids network to incorporate textual-like features into local visual features. It enhances the local visual feature representation. Note that global text features are used to obtain text-like features and guide the learning of fused features. 


\subsection{Dual-path transfer learning}\label{dual-path}
Dual encoder models can align two modalities representations in the same embedding space. We adopt pretrained CLIP and GeoRSCLIP as our backbone to extract features since they are effective model to learn strong feature representations. 

\textbf{Multi-level Vision Transformer} For image encoder, the Modified ResNet used in CLIP provides multi-level visual features, while ViT used in CLIP and GeoRSCLIP only produces single-scale feature maps. To reconcile this discrepancy, we follow the technique introduced in~\cite{ali2021xcit} to generate multi-scale feature from ViT. We integrate four resolution-modifying modules at evenly distributed intervals of last four transformer blocks.
The initial module upsamples the feature map by a factor of 4 using two stride-two 2$\times$ transposed convolutions, group normalization and GeLU activation. The output of the second block is upsampled by 2$\times$ using a single stride-two 2$\times$2 transposed convolution. The next block's output is taken as is, and the final ViT block's output is downscaled by a factor of 2 using stride-two 2$\times$2 max pooling. Each of these modules preserves the ViT's embedding/channel dimension.

\textbf{Image Encoder.} For an input image $I \in \mathbb{R}^{H \times W \times 3}$, the global visual feature $\mathbf{o}_{v c} \in R^D$ and multi-level features are extracted. We select 2th-5th level visual features for further fusion, which are defined as  $\mathbf{o}_{v2} \in \mathbb{R}^{\frac{H}{8} \times \frac{W}{8} \times D_2}$, $\mathbf{o}_{v3} \in \mathbb{R}^{\frac{H}{16} \times \frac{W}{16} \times D_3}$, $\mathbf{o}_{v4} \in \mathbb{R}^{\frac{H}{32} \times \frac{W}{32} \times D_4}$, and $\mathbf{o}_{v5} \in \mathbb{R}^{1 \times D_5}$ respectively. The 2th-4th level features are the local representations of the image, with the 5th visual feature serving as the global representation. Note that $D$ and $D_{i}$ are the $i$th level feature dimension, $H$ and $W$ are the height and width of the original image.\par
\textbf{Text Encoder.} For an input caption $T \in \mathbb{R}^{L}$, textual features $\mathbf{o}_{t}^{\prime} \in R^{L \times D}$ is extracted by Transformer with the architecture modifications described in \cite{radford2019language,radford2021learning}. $T$ is bracketed with \texttt{[SOS]} and \texttt{[EOS]} tokens. The activations of the highest layer of the transformer at the \texttt{[EOS]} token are treated as the global textual feature $\mathbf{o}_{t} \in R^{D}$, which is transformed into the multi-modal embedding space. Note that $D$ is the feature dimension, $L$ is the length of the caption.\par

\subsection{Progressive Cross-modal Feature Fusion}\label{feature fusion}
Given the abundance of geographical objects within RS images, relying solely on the aligned global visual feature and textual feature acquired through CLIP may not yield the most optimal results for RS ITR. Consequently, we design two training stage to progressively fuse fine-grained semantic features to enrich the visual representation.\par
\textbf{Visual-to-text feature mapping.} Different from just fusing visual features, the first stage of training aims to learn visual-to-text feature mapping (V2TMap).
The global visual feature is firstly transformed through image adapter\cite{gao2023clip}. It exclusively integrates a limited number of supplementary learnable bottleneck linear layers into the image encoder, maintaining the original backbone in a frozen state throughout the training process. For the extracted global image feature $\mathbf{o}_{v5}$, a learnable feature adapter $\mathcal{F}_{adp}$ transforms $\mathbf{o}_{v5}$ into $\mathbf{o}_{v5}'$ , which contains two layers of linear layers. 
\begin{equation}
	\mathbf{o}_{v5} ' = \mathcal{F}_{adp}(\mathbf{o}_{v5})
\end{equation}
A residual connection is adopted for the feature adapter to avoid forgetting the original knowledge encoded by the pretrained CLIP. The residual ratio $\alpha$  helps adjust the degree of maintaining the original knowledge for better performance.
The new transformed feature $\mathbf{o}_{v5}^*$ is calculated as follows.
\begin{equation}
	\mathbf{o}_{v} = \alpha \mathbf{o}_{v5}' + (1 - \alpha)\mathbf{o}_{v5}
	\label{residual ratio}
\end{equation}\par 
We project the newly transformed visual feature $\mathbf{o}_{v}$ and paired text feature $\mathbf{o}_{t}$ into a shared embedding space, allowing $\mathbf{o}_{v}$ to acquire semantic information, akin to textual features.

\textbf{Hierarchical Feature Enhancement.} Inspired by \cite{lin2017feature}, $\mathbf{o}_{v2}, \mathbf{o}_{v3}, \mathbf{o}_{v4}$ is fused with $\mathbf{o}_{v}$  in a top-down pathway in the second training stage, named hierarchical feature enhancement (HFE). We firstly enhance $\mathbf{o}_{v4}$ with $\mathbf{o}_{v}$ by element-wise multiplication and then upsample the spatial resolution by a factor of 2 to obtain the multi-modal feature $\mathbf{o}_{m4} \in R^{\frac{H}{16} \times \frac{W}{16} \times D}$:
\begin{equation}
	\mathbf{o}_{m 4}=\mathcal{C}_{3 \times 3} (\mathcal{F}_{up}\left(\sigma\left(\mathcal{F}_{proj}(\mathbf{o}_{v 4})\right) \cdot \sigma\left(\mathcal{F}_{proj}(\mathbf{o}_{tc})\right)\right)),
\end{equation}
where $\mathcal{F}_{up}(\cdot)$ denotes 2 $\times$ upsampling, $\cdot$ denotes the elementwise multiplication, $\sigma$ demotes RELU, and $\mathcal{F}_{proj}(\cdot)$ denotes a projector with 1 $\times$ 1 convolution to transform the visual and textual feature into the same feature dimension. $\mathcal{C}_{3 \times 3}(\cdot)$ is a 3 $\times$ 3 convolution to reduce the aliasing effect of upsampling. Then, $\mathbf{o}_{m4}$ is merged with $\mathbf{o}_{v3}$ to generate $\mathbf{o}_{m3}$:
\begin{equation}
	\mathbf{o}_{m 3}=\mathcal{C}_{3 \times 3}\left(\mathcal{F}_{\text {concat }}\left(\sigma\left(\mathcal{F}_{\text {proj }}\left(\mathbf{o}_{v 3}\right)\right), \sigma\left(\mathcal{F}_{\text {proj }}\left(\mathbf{o}_{m 4}\right)\right)\right)\right),
\end{equation}
where $\mathcal{F}_{concat}(\cdot)$ denotes the concatenation operation. Afterwards, $\mathbf{o}_{v2}$ undergoes a 2 $\times$ 2 average pooling with 2 strides\cite{wang2022cris} and then is fused with $\mathbf{o}_{m 3}$:

\begin{equation}
	\begin{aligned}
		& \mathbf{o}_{m 2}=\mathcal{C}_{3 \times 3}\left(\mathcal{F}_{\text {concat }}\left(\sigma\left(\mathcal{F}_{\text {proj }}\left(\mathbf{o}_{v_2}^{\prime}\right)\right), \sigma\left(\mathcal{F}_{\text {proj }}\left(\mathbf{o}_{m 3}\right)\right)\right)\right) \\
		& \mathbf{o}_{v 2}^{\prime}=\mathcal{M}\left(\mathbf{o}_{v 2}\right),
	\end{aligned}
\end{equation}
where $\mathcal{M}$ denotes a kernel size of 2 $\times$ 2 average pooling
with 2 strides.
Subsequently, we aggregate three multi-modal features with a 2D spatial-aware feature $\mathbf{o}_{spatial}$ into enhanced visual feature $\mathbf{o}_{e} \in R^{N \times D}$:
\begin{equation}
	\begin{aligned}
		&\mathbf{o}_{m}=\mathcal{F}_{fuse} ^ {1 \times 1}(\mathcal{F}_{concat}(\mathbf{o}_{m2}, \mathbf{o}_{m3},  \mathbf{o}_{m4})) \\
		&\mathbf{o}_{e}^{prime} = \mathcal{F}_{flatten}(\mathcal{F}_{fuse} ^ {3 \times 3}({F}_{concat}(\mathbf{o}_{m}, \mathbf{o}_{spatial})),
	\end{aligned}
\end{equation}
where $\mathcal{F}_{fuse} ^ {1 \times 1}$ is a 1 $\times$ 1 convolution layer and $\mathcal{F}_{fuse} ^ {3 \times 3}$ is a 3 $\times$ 3 convolution layer, $\mathcal{F}_{flatten}$ flattens the spatial domain of $\mathbf{o}_{v}$ into a sequence and $N = \frac{H}{16} \times \frac{W}{16}$. Finally, attention pooling(AP) extract the global visual feature $o_{e}$ from $o_{e}^{\prime}$ as follows.
\begin{equation}
	\begin{aligned}
		&o_{z} = [o_{cls}; o_{e}^{\prime}] + E_{pos},
		\\
		&o_{e} = \mathcal{F}_{MHSA}(o_{z})[0,:],
	\end{aligned}
\end{equation}
where $o_{cls}$ serves as image representation capturing global visual feature, $E_{pos}$ is the positional embedding and  $\mathcal{F}_{MHSA}(\cdot)$ denotes multi-head self-attention.

\subsection{Model Training }\label{training}
Assume a batch of $B$ image-text pairs $\{(I_i,T_i)\}_{i=1} ^{B}$, where $I_i$ and $T_i$ are the image and text inputs of the $i$-th pair. We first train V2TMap by utilizing contrastive loss following original CLIP. The image and text inputs are encoded into $\{ \mathbf{o}_v^i \}_{i=1}^{B}$ and  $\{ \mathbf{o}_t^i \}_{i=1}^{B}$, respectively. The contrastive loss $\mathcal{L}_{\theta_1}$ is adopted to maximize the similarity between the paired $\mathbf{o}_t^i$ and $\mathbf{o}_v^i$ and minimize the similarity with other irrelevant $\mathbf{o}_v^j$ or $\mathbf{o}_t^j$: 

\begin{equation}
	\begin{aligned}
		&\mathcal{L}_{\theta_1} = \mathcal{L}_{I \rightarrow T} + \mathcal{L}_{I \leftarrow T} \\
		&\mathcal{L}_{I \rightarrow T}=-\frac{1}{B} \sum_{i=1}^B \log \frac{\exp \left(\mathbf{o}_v^i \cdot \mathbf{o}_t^i / \tau\right)}{\sum_{j=1}^B \exp \left(\mathbf{o}_v^i \cdot \mathbf{o}_t^j / \tau\right)}\\
		&\mathcal{L}_{I \leftarrow T}=-\frac{1}{B} \sum_{i=1}^B \log \frac{\exp \left(\mathbf{o}_t^i \cdot \mathbf{o}_v^i / \tau\right)}{\sum_{j=1}^B \exp \left(\mathbf{o}_t^i \cdot \mathbf{o}_v^j / \tau\right)},
	\end{aligned}
\end{equation}
where $\tau$ is the temperature parameter to scale the logits.\par


	\section{Experiment}
\label{Experiment}

\subsection{Significance Testing on Spatial Sampling }
\label{Exper-BR}
\begin{figure}[htbp]
	\vspace{-0.3cm}
	\centering
	\captionsetup{justification=justified}
	\includegraphics[width=0.4\textwidth]{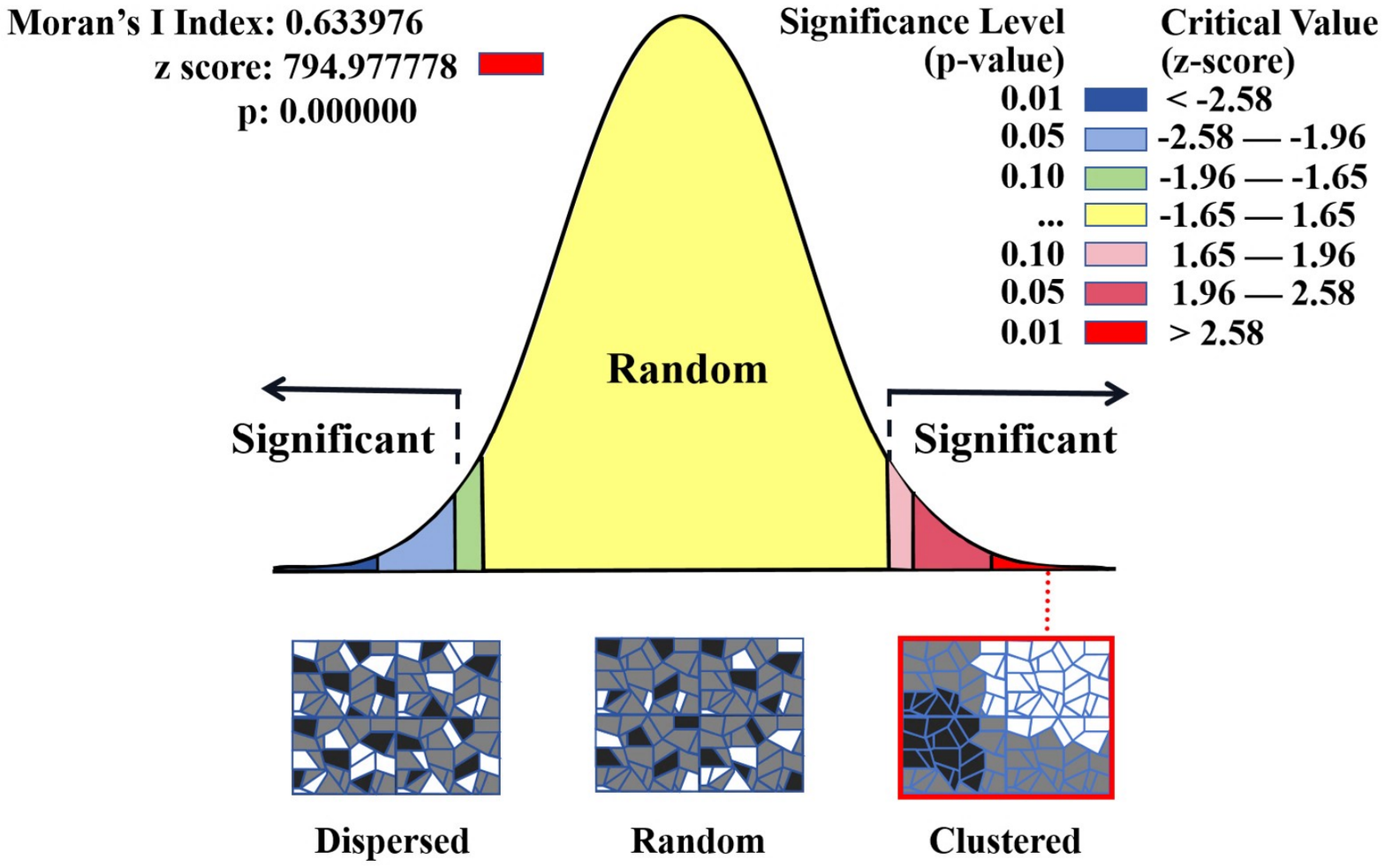} 
	\vspace{-0.3cm}
	\caption{Significant test of Global Moran's Index on global nighttime data.}
	\label{moran}
	
\end{figure}

Global nighttime data can reflect the intensity of human activities, social and economic development degree~\cite{zheng2023nighttime}, etc., which is related to the richness of OSM labels and Google images. Consequently, we opted to utilize the VIIRS Stray Light Corrected Nighttime Day/Night Band~\cite{mills2013viirs} data in 2022 for spatial analysis (Moran'I), aiming to delineate the sampling area.
The basic assumption for the Moran's I statistic is that the data values are independent and randomly distributed in the geographical space. When the p-value obtained is greater than 0.05, the basic assumption is accepted implying that the data values are randomly spread out spatially. Oppositely, the p-value is less than 0.05 and the z-score is negative, the basic assumption of randomness is rejected, inferring that the high and the low values in the dataset are dispersed spatially. Similarly, when the P-value is less than 0.05 with a positive Z-score, the assumption of randomness is again ruled out and the inference drawn is that the high and/or low data values are spatially clustered in the geographical space. As shown in  Fig.~\ref{moran}, nighttime data around the world are clustered. The cluster conditions in neighborhood areas were classified into four cluster types: High-High, High-Low, Low-High and Low-Low, according to the positive and negative values of the $Z$ score and the $LISA$ value. The cluster type is shown in Tab.~\ref{moran's I}, where $\uparrow$ represents positive value; otherwise, negative value. The High-High cluster type suggests spatial agglomerations of neighboring areas marked by high levels of economics. Conversely, the Low-Low cluster type indicates spatial agglomerations of neighboring areas with limited urbanization. The High-Low and Low-High cluster types imply significant development disparities among neighboring areas.

\begin{table}[!t]
	\captionsetup{justification=justified}
	\caption{Cluster type according to Local Moran's I. $\uparrow$ denotes value is greater than zero, while $\downarrow$ denotes less than zero.}
	\begin{center}
		\resizebox{.9\columnwidth}{!}{
			\begin{tabular}{c|c|c|c}
				\hline
				$Z_{i}$&$LISA$&$I_{local}$&Cluster Type\\
				\hline
				$\uparrow$ & $\uparrow$ & $\uparrow$ &High-High\\
				$\downarrow$ & $\downarrow$ & $\uparrow$ &Low-Low\\
				$\downarrow$ & $\uparrow$ & $\downarrow$ &Low-High\\
				$\uparrow$ & $\downarrow$ & $\downarrow$ &High-Low\\
				\hline
			\end{tabular}
		
		}
	\end{center}
	\label{moran's I}
	\vspace{-0.4cm}
\end{table}

\subsection{Experiments}
\subsubsection{Experimental preparation}
\textbf{Implementation details. }
LuojiaHOG is divided into 70\%, 10\% and 20\% images for training, validation and test set, respectively. The proposed CISEN is implemented on the PyTorch platform, and all deep models are trained using CPU with i7-6850K, GPU with 32GB Tesla V100. We select CLIP with ResNet-50, ViT-B and GeoRSCLIP with ViT-B/32 as our backbone. Input images are resized to 224 $\times$ 224 pixels, and the input tokens length are set with a maximum sentence length of 328 instead of default 77. Initially, the backbone, except position embedding, is frozen, we train the network for 60 epochs using the AdamW optimizer with the learning rate $\lambda$ = 0.0001. The learning rate is decreased by a factor of 0.1 at the 40th epoch. Following the same settings, V2TMap is trained with backbone frozen in the first stage. In the second stage, V2TMap is frozen as well and only the HFE module is trainable.\par
\textbf{Evaluation metrics. }
We evaluate the image retrieval quality using four widelyused metrics: Average Cumulative Gains (ACG) ~\cite{jarvelin2017ir}, Normalized Discounted Cumulative Gains (NDCG)  ~\cite{jarvelin2002cumulated}, Mean Average Precision (MAP) \cite{baeza1999modern} and Weighted Mean Average Precision (WMAP) \cite{zhao2015deep}.
ACG represents the average number of shared labels between the query image and the top $n$ retrieved images. Given a query image $I_q$, the ACG score of the top $n$ retrieved images is calculated by
\begin{equation}
	\text{ACG} @ n=\frac{1}{n} \sum_i^n C(q, i),
\end{equation}
where $n$ denotes the number of top retrieval images and $C(q, i)$ is the number of shared labels between $I_q$ and $I_i$.
NDCG is a popular evaluation metric in information retrieval. Given a query image Iq, the DCG score of top $n$ retrieved images is defined as
\begin{equation}
	\text{DCG} @ n=\sum_i^n \frac{2^{C(q, i)}-1}{\log (1+i)} .
\end{equation}

Then, the normalized DCG (NDCG) score at the position $n$ can be calculated by 
$N D C G @ n=\frac{D C G @ n}{Z_n}$, where $Z_n$ is the maximum value of $DCG@n$, which constrains the value of $NDCG$ in the range $[0,1]$.
MAP is the mean of average precision for each query, which can be calculated by
\begin{equation}
	\text{MAP}=\frac{1}{Q} \sum_q^Q A P(q),
\end{equation}
where
\begin{equation}
	\text{AP}(q)=\frac{1}{N_{T r}(q) @ n} \sum_i^n\left(\operatorname{Tr}(q, i) \frac{N_{T r}(q) @ i}{i}\right),
\end{equation}
and $Tr(q,i) \in {0, 1}$ is an indicator function that if $I_q$ and $I_i$ share some labels, $Tr(q,i)=1$; otherwise $Tr(q,i)=0$. $Q$ is the number of query sets and $N_{T r}(q)@i$ indicates the number of the relevant imgaes w.r.t the query image $I_q$ within the top $i$ images. \par
The definition of WMAP is similar with MAP. The only difference is that WMAP computes the average ACG scores at each top $n$ retrieved image rather than average precision. WMAP can be calculated by
\begin{equation}
	\text{WMAP}=\frac{1}{Q} \sum_q^Q\left(\frac{1}{N_{T r}(q) @ n} \sum_i^n(T r(q, i) \times ACG @ i)\right)
\end{equation}

\textbf{Comparison with state-of-the-art models. }To verify the effectiveness of CISEN, we conducted some experiments on LuojiaHOG. We select current state-of-the-art (SOTA) in image-text retrieval tasks. They are ALBEF ~\cite{li2021align}, ALIGN~\cite{jia2021scaling}, CLIP~\cite{radford2021learning}, FILIP~\cite{yao2021filip}, Wukong~\cite{gu2022wukong}, BLIP~\cite{li2022blip}, GeoRSCLIP~\cite{zhang2023rs5m}. For fair comparison, we froze the backbone of all models and utilize image adapter \cite{gao2024clip} for finetuning on ITR tasks.
We train all the networks with pre-trained weights with a learning rate of 0.0001, and divided by 10 after 40 epoches. All the networks are optimized using the AdamW with a momentum of 0.9, and weight decay of 0.0001. Further, the relevant parameters can be slightly adjusted, making it applicable to the ITR. \par 

\begin{figure}[htbp]
	\centering
	\captionsetup{justification=justified}
	\subfigure[CLIP (ViT)]{\includegraphics[width=0.46\textwidth]{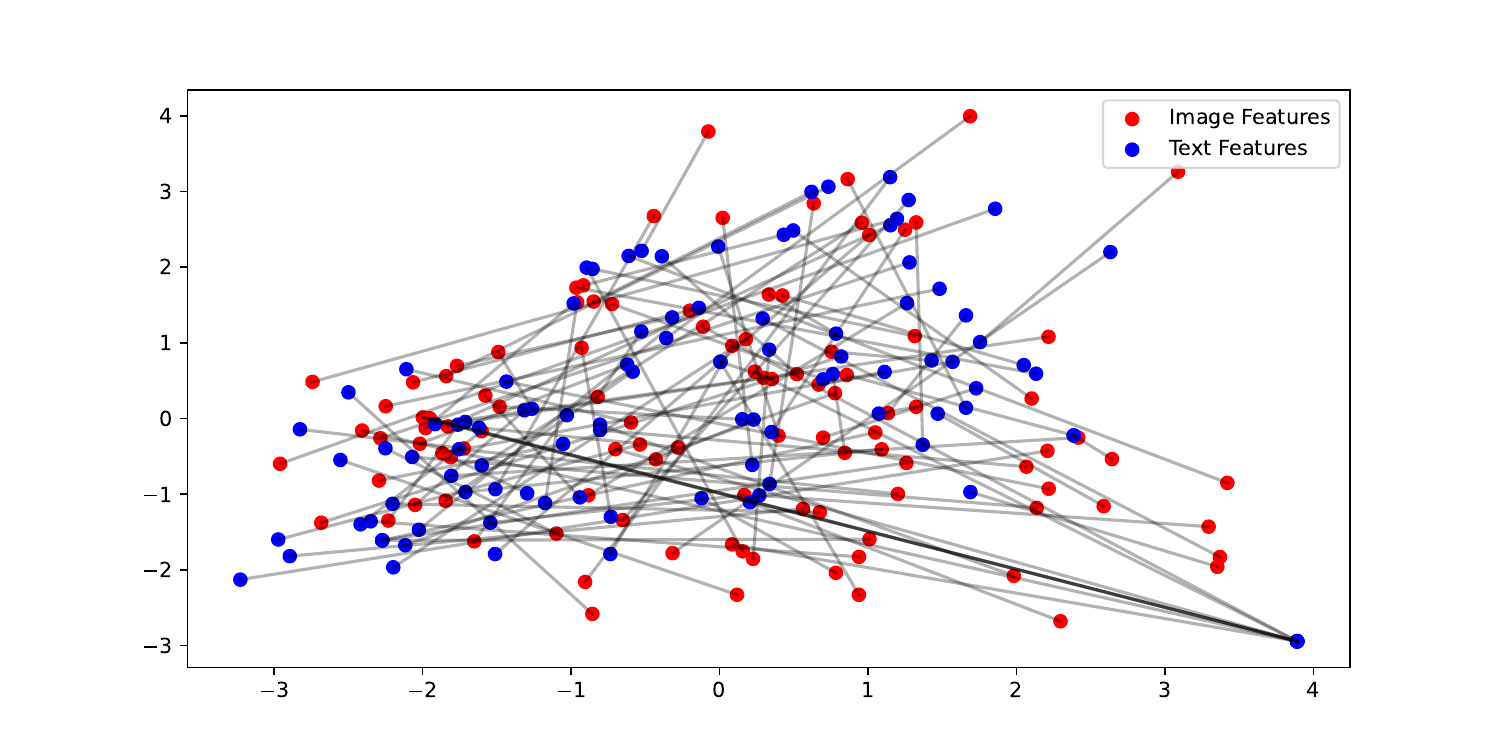}\label{zeroshot}}
	\subfigure[CLIP (ViT) + V2TMap]{\includegraphics[width=0.46\textwidth]{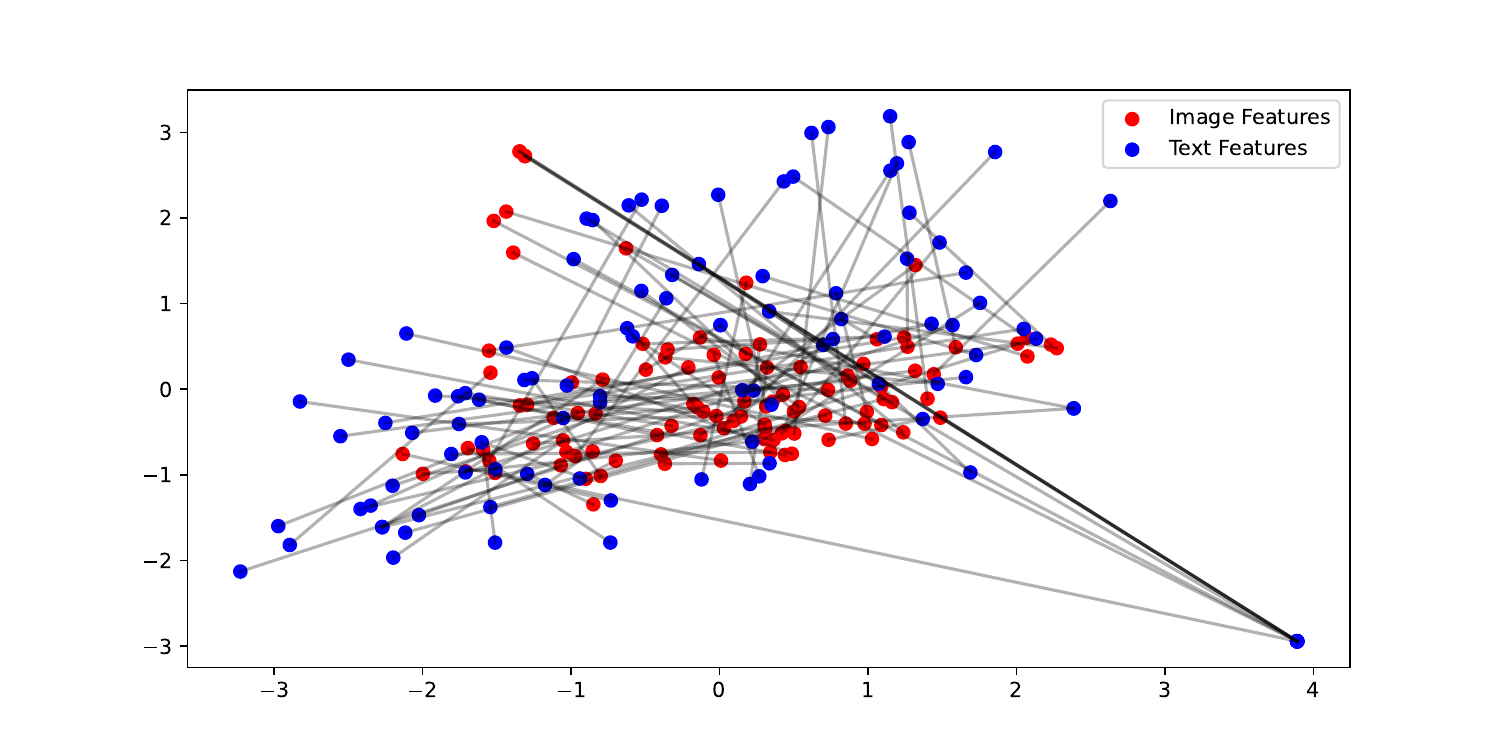}\label{2}}
	\subfigure[CISEN (ViT)]{\includegraphics[width=0.46\textwidth]{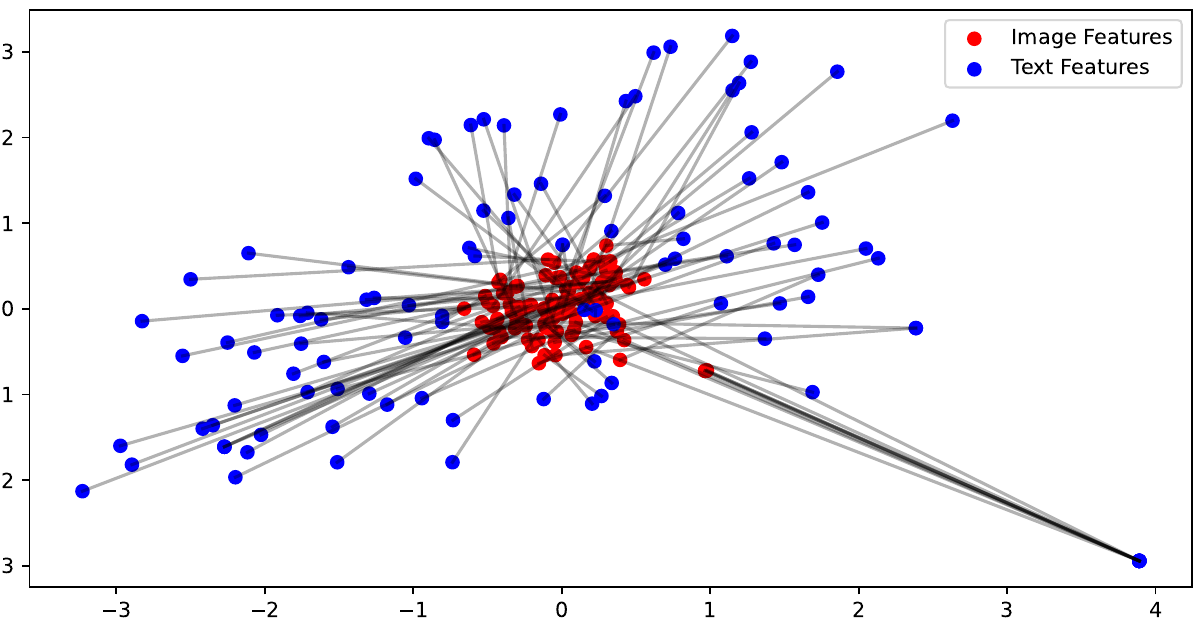}\label{CISEN(ViT)}}
	\caption{UMAP visualization of generated embeddings from models. Paired inputs are fed into the pre-trained models and the embeddings are visualized in 2D using UMAP (lines indicate pairs).}
	\vspace{-1em}
	\label{modality gap}
\end{figure}

\begin{figure*}[htbp]
	\centering
	\captionsetup{justification=justified}
	\includegraphics[width=1\textwidth]{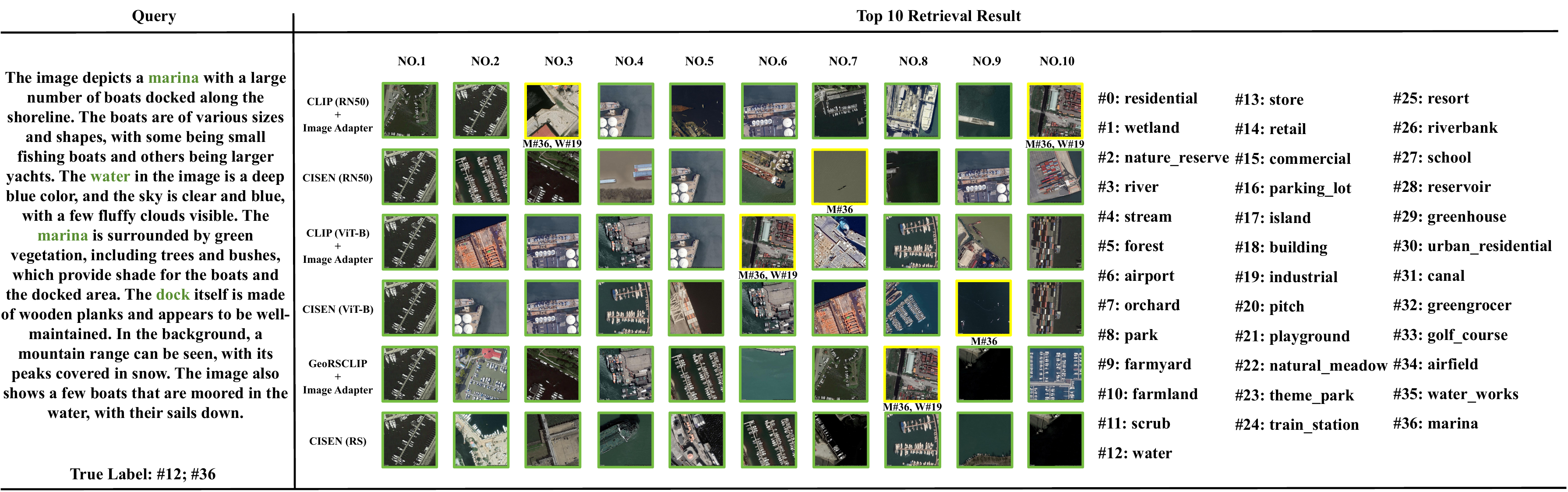}
	\caption{The T2I retrieval results on CLIP, GeoRSCLIP and CISEN. The retrieved images with red box are incorrect, with yellow box are inaccurate and with green box are correct. At the bottom are some third-level labels.}
	\label{ti results} 
\end{figure*}

\begin{figure*}[htbp]
	\centering
	\captionsetup{justification=justified} 
	\includegraphics[width=1\textwidth]{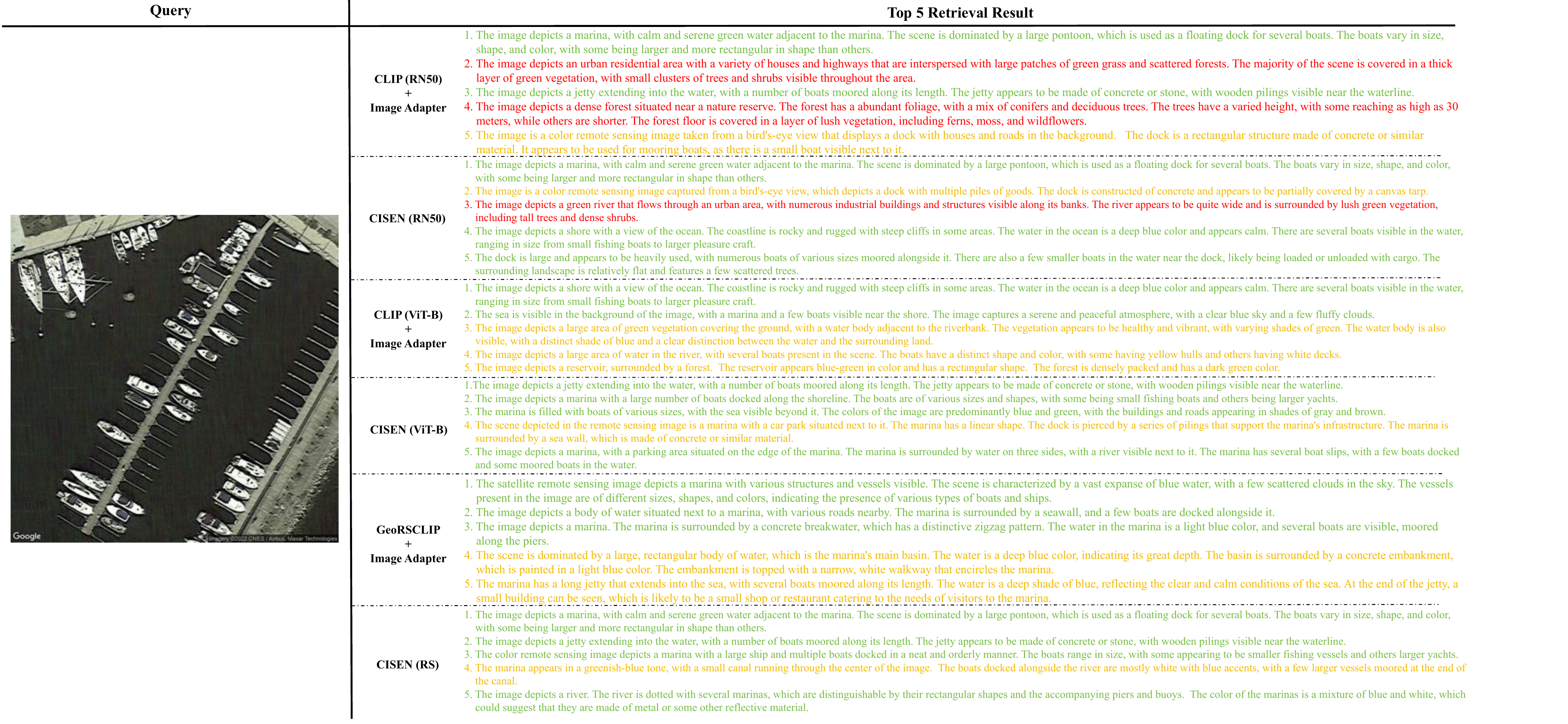}
	\caption{The I2T retrieval results on CLIP, GeoRSCLIP and CISEN. The retrieved texts in red are incorrect, in yellow are inaccurate and in green are correct.}
	\vspace{-1em}
	\label{it results}
\end{figure*}

\subsubsection{Quantitative evaluation on LuojiaHOG}
We quantify the ITR retrieval  performance by comparing current SOTA vision-language models with our CISEN in terms of MAP, WMAP, NDCG and ACG scores. Tab.~\ref{retrieval 21} and Tab.~\ref{retrieval 131} shows the quantitative results from second-level and third-level labels, respectively. By and large, GeoRSCLIP is more suited for remote sensing image retrieval tasks owing to its pretraining on remote sensing datasets. It has demonstrated notable performance in both I2T and T2I retrieval tasks. CISEN (RS), utilizing GeoRSCLIP as its backbone, achieved superior performance across all tasks. In the I2T retrieval task, the results of CLIP with ViT-B as its backbone (CLIP-ViT) exhibit an average increase of 1.9\%$\sim$2\% MAP, 2.8\%$\sim$3\% WMAP, 1\%$\sim$1.4\% NDCG and 3\%$\sim$3.4\% ACG at second-level, 0.4\%$\sim$1.2\% MAP, 1.2\%$\sim$2.3\% WMAP, 0.5\%$\sim$1\% NDCG and 0.4\%$\sim$1.8\% ACG at third-level compared to using ResNet-50 (CLIP-RN50) as the backbone. Conversely, the difference in results between the two backbones in text-image retrieval tasks is negligible. However, this phenomenon changes significantly with the introduction of CISEN. CISEN (ViT), leveraging ViT-B as the backbone, yields substantial enhancements in both I2T and T2I retrieval tasks compared with CISEN (RN50). For example, CISEN (ViT) brings increments of 5\% WMAP@5 and 3.4\% WMAP@5 on T2I and I2T retrieval task at third-level compared to CISEN (RN50). In contrast, Filip, another dual encoder model integrating fine-grained token-wise contrastive learning based on CLIP, does not yield optimal results on datasets like LuojiaHOG which is characterized by longer text lengths and more complex scenes. As its improvement, Wukong's performance matches that of the dual encoder combined with fusion module models, like Blip and Albef. It is noteworthy that ALIGN outperforms CISEN (RN50) and slightly trails behind CISEN (ViT-B).\\

\begin{figure*}[htb]
	\centering
	\subfigure[Image To Text Retrieval]{\includegraphics[width=0.64\textwidth]{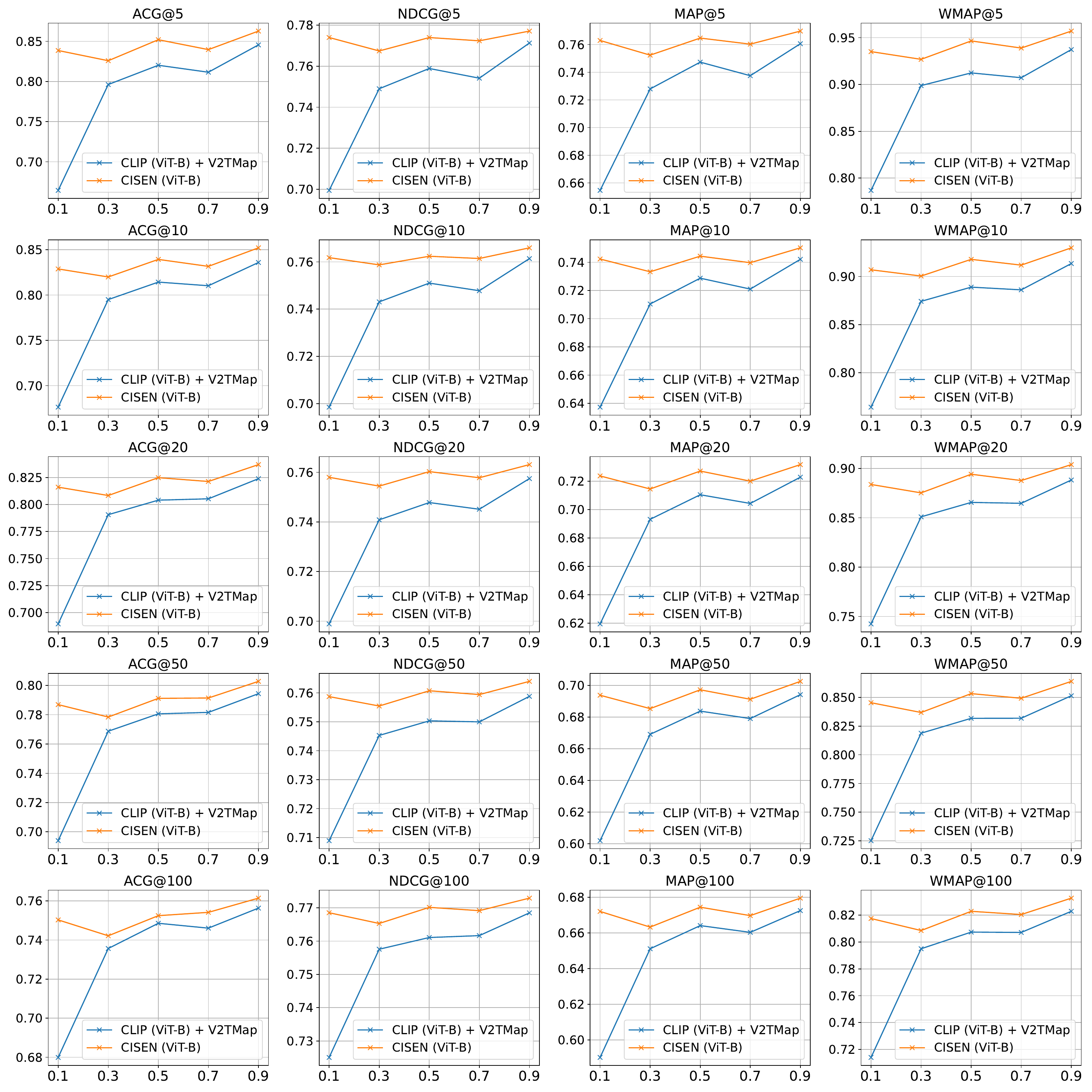}\label{vit_21_i2t}}
	\subfigure[Text To Image Retrieval]{\includegraphics[width=0.64\textwidth]{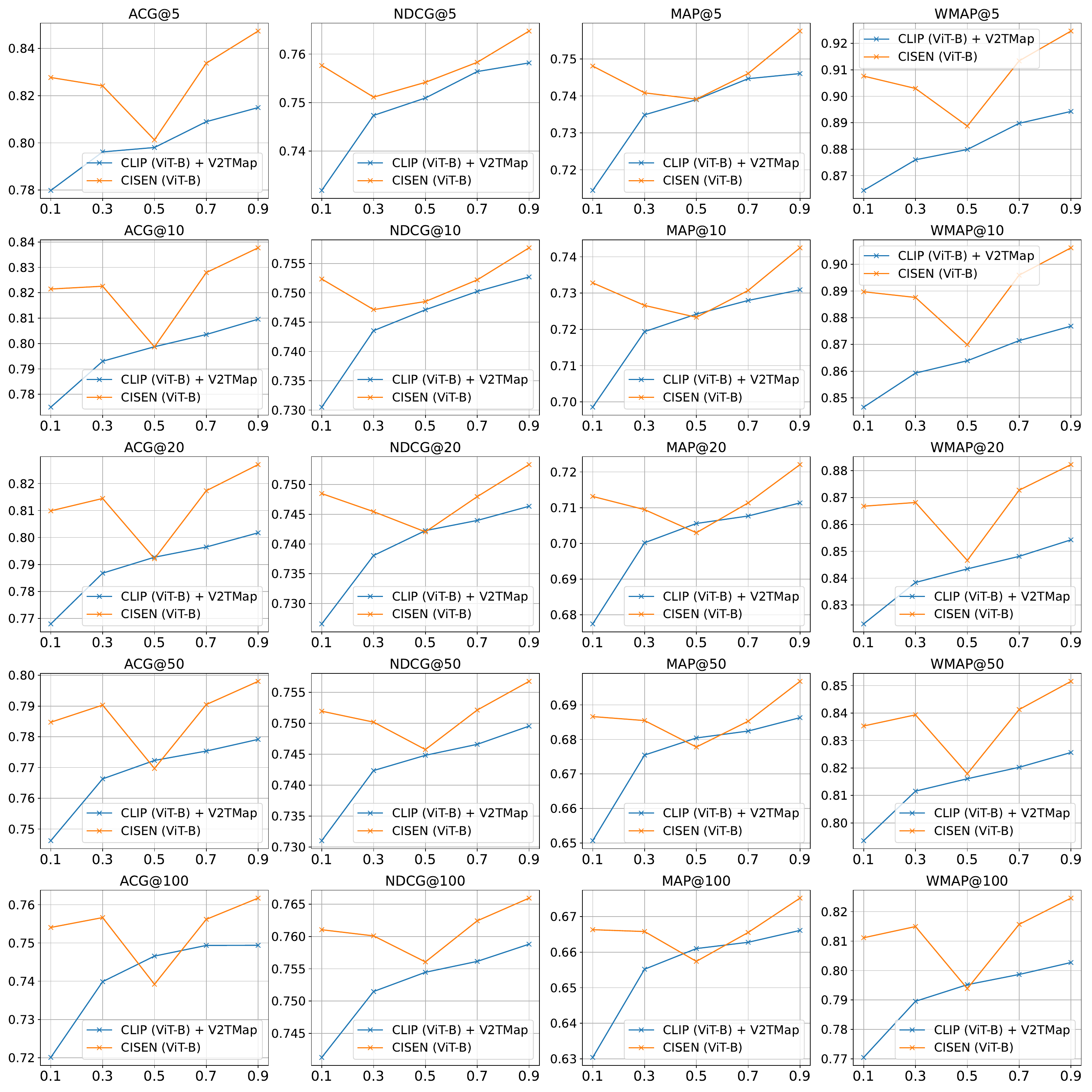}\label{vit_21_t2i}}
	\caption{Influence of different residual ratio on the second-level ITR performance based on CISEN(ViT).}
	\label{vit_21_itr} 
\end{figure*}


\begin{figure*}[htb]
	\centering
	\subfigure[Image To Text Retrieval]{\includegraphics[width=0.64\textwidth]{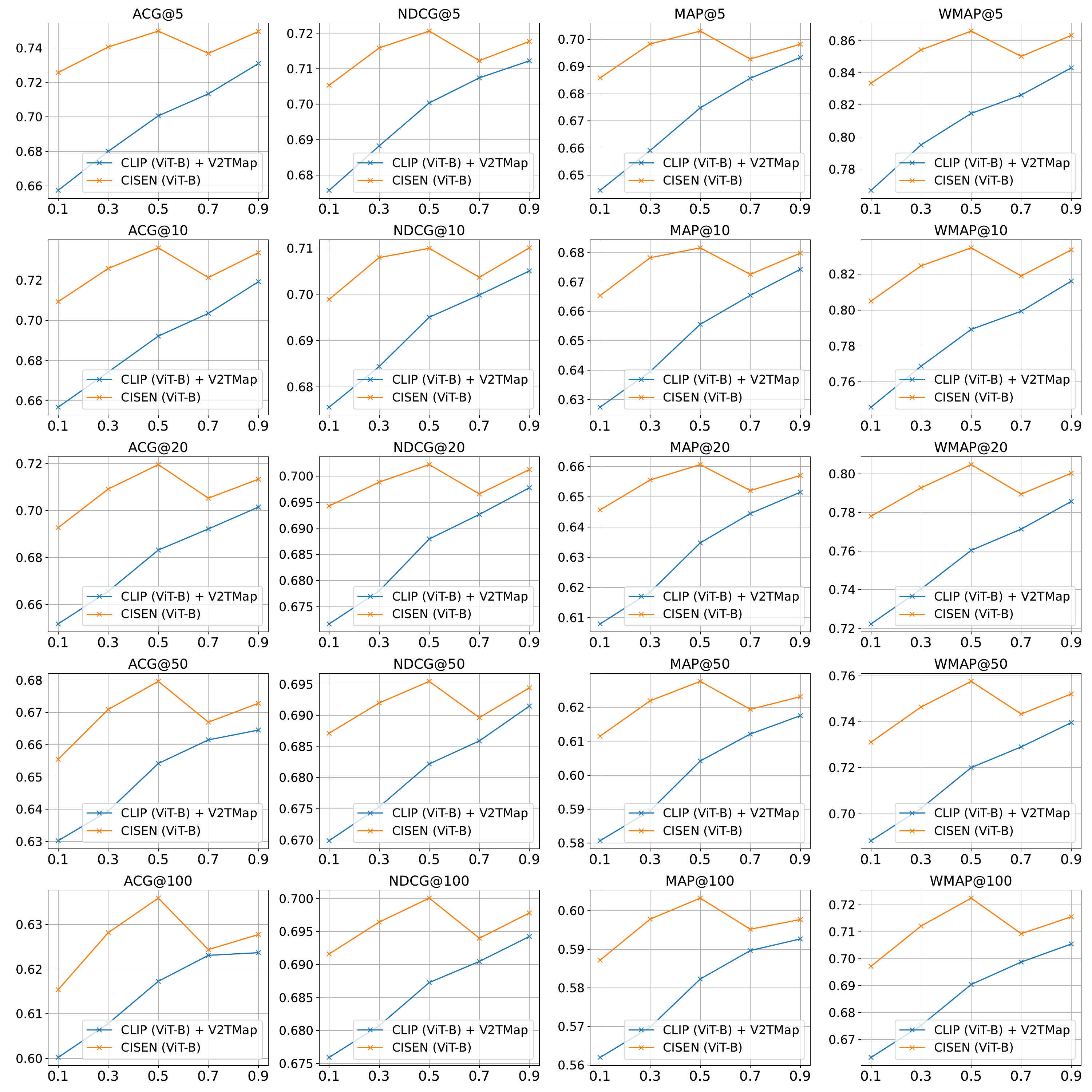}\label{vit_131_i2t}}
	\subfigure[Text To Image Retrieval]{\includegraphics[width=0.64\textwidth]{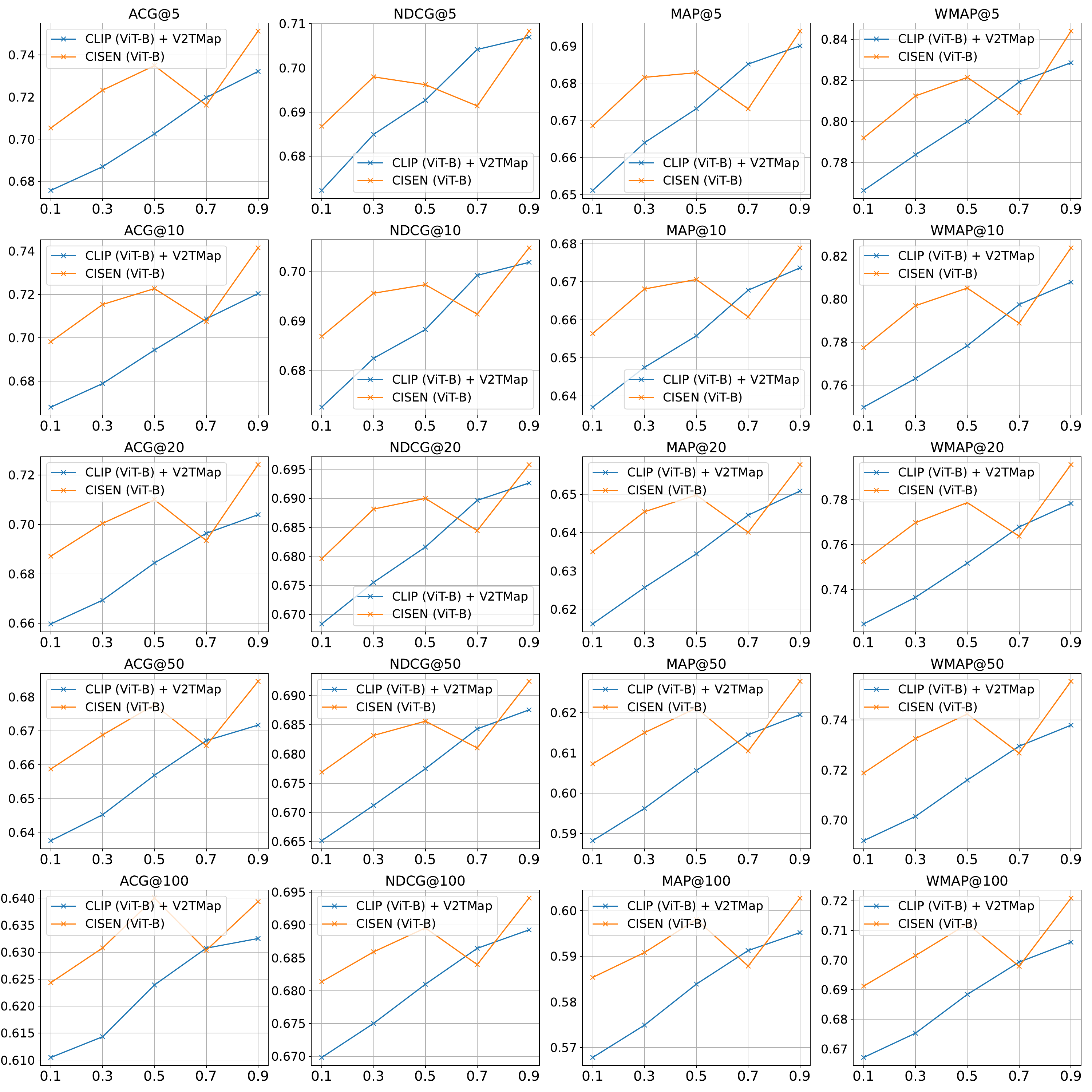}\label{vit_131_t2i}}
	\caption{Influence of different residual ratio on the third-level ITR performance based on CISEN (ViT).}
	\label{vit_131_itr}
\end{figure*}

\begin{table*}[h!]
	\renewcommand{\arraystretch}{1.5}
	\captionsetup{justification=justified}
	\caption{Quantative performance comparison of all models in terms of MAP@n, Weighted MAP@n, NDCG@n and ACG@n (n=5, 10, 20 ,50 , 100) on LuojiaHOG second-level labels. The best is marked in bold.}
	\centering
	\resizebox{1.8\columnwidth}{!}{
		\begin{tabular}{clclclcccccccccc}
			\hline
			\multicolumn{2}{c}{\multirow{2}{*}{\textbf{Methods}}} & \multicolumn{2}{c}{\multirow{2}{*}{\textbf{Image Encoder}}} & \multicolumn{2}{c}{\multirow{2}{*}{\textbf{Text Encoder}}} & \multicolumn{5}{c}{Image To Text} & \multicolumn{5}{c}{Text To Image} \\
			\multicolumn{2}{c}{} & \multicolumn{2}{c}{} & \multicolumn{2}{c}{} & @5 & @10 & @20 & @50 & @100 & @5 & @10 & @20 & @50 & @100 \\ \hline
			\multicolumn{6}{c}{} & \multicolumn{10}{c}{MAP} \\
			\multicolumn{2}{c}{\textbf{Albef}} & \multicolumn{2}{c}{ViT-B/16} & \multicolumn{2}{c}{BERT} & 0.7312 & 0.7087 & 0.6881 & 0.6591 & 0.6377 & 0.6257 & 0.6059 & 0.5732 & 0.5387 & 0.5194 \\
			\multicolumn{2}{c}{\textbf{Align}} & \multicolumn{2}{c}{EfficientNet-B7} & \multicolumn{2}{c}{BERT} & 0.7516 & 0.7304 & 0.7140 & 0.6852 & 0.6640 & 0.7339 & 0.7196 & 0.7007 & 0.6752 & 0.6555 \\
			\multicolumn{2}{c}{\multirow{2}{*}{\textbf{CLIP}}} & \multicolumn{2}{c}{RN50} & \multicolumn{2}{c}{\multirow{2}{*}{Transformer}} & \multicolumn{1}{l}{0.7411} & \multicolumn{1}{l}{0.7228} & \multicolumn{1}{l}{0.7036} & \multicolumn{1}{l}{0.6740} & \multicolumn{1}{l}{0.6519} & \multicolumn{1}{l}{0.7468} & \multicolumn{1}{l}{0.7287} & \multicolumn{1}{l}{0.7046} & \multicolumn{1}{l}{0.6725} & \multicolumn{1}{l}{0.6486} \\
			\multicolumn{2}{c}{} & \multicolumn{2}{c}{ViT-B/32} & \multicolumn{2}{c}{} & \multicolumn{1}{l}{0.7606} & \multicolumn{1}{l}{0.7420} & \multicolumn{1}{l}{0.7229} & \multicolumn{1}{l}{0.6941} & \multicolumn{1}{l}{0.6725} & \multicolumn{1}{l}{0.7460} & \multicolumn{1}{l}{0.7309} & \multicolumn{1}{l}{0.7113} & \multicolumn{1}{l}{0.6863} & \multicolumn{1}{l}{0.6661} \\
			\multicolumn{2}{c}{\textbf{Blip}} & \multicolumn{2}{c}{ViT-B/32} & \multicolumn{2}{c}{BERT} & 0.7294 & 0.7066 & 0.6780 & 0.6441 & 0.6204 & 0.6959 & 0.6783 & 0.6553 & 0.6239 & 0.5997 \\
			\multicolumn{2}{c}{\textbf{Filip}} & \multicolumn{2}{c}{ViT-B/32} & \multicolumn{2}{c}{Transformer} & 0.6564 & 0.6294 & 0.6053 & 0.5763 & 0.5603 & 0.6387 & 0.6193 & 0.5945 & 0.5669 & 0.5532 \\
			\multicolumn{2}{c}{\textbf{WuKong}} & \multicolumn{2}{c}{ViT-B/32} & \multicolumn{2}{c}{Transformer} & 0.7295 & 0.7112 & 0.6953 & 0.6659 & 0.6427 & 0.6738 & 0.6585 & 0.6403 & 0.6177 & 0.6031 \\
			\multicolumn{2}{c}{\textbf{GeoRSCLIP}} & \multicolumn{2}{c}{ViT-B/32} & \multicolumn{2}{c}{Transformer} & 0.7667 & 0.7506 & 0.7318 & 0.7041 & 0.6826 & 0.7616 & 0.7461 & 0.7263 & 0.6990 & 0.6781 \\
			\multicolumn{2}{c}{\multirow{2}{*}{\textbf{CISEN}}} & \multicolumn{2}{c}{RN50} & \multicolumn{2}{c}{\multirow{2}{*}{Transformer}} & \multicolumn{1}{l}{0.7433} & \multicolumn{1}{l}{0.7234} & \multicolumn{1}{l}{0.7049} & \multicolumn{1}{l}{0.6752} & \multicolumn{1}{l}{0.6533} & \multicolumn{1}{l}{0.7306} & \multicolumn{1}{l}{0.7133} & \multicolumn{1}{l}{0.6930} & \multicolumn{1}{l}{0.6658} & \multicolumn{1}{l}{0.6457} \\
			\multicolumn{2}{c}{} & \multicolumn{2}{c}{ViT-B/32} & \multicolumn{2}{c}{} & \multicolumn{1}{l}{0.7698} & \multicolumn{1}{l}{0.7502} & \multicolumn{1}{l}{0.7318} & \multicolumn{1}{l}{0.7025} & \multicolumn{1}{l}{0.6795} & \multicolumn{1}{l}{0.7575} & \multicolumn{1}{l}{0.7425} & \multicolumn{1}{l}{0.7221} & \multicolumn{1}{l}{0.6968} & \multicolumn{1}{l}{0.6752} \\
			\multicolumn{2}{c}{\textbf{CISEN (RS)}} & \multicolumn{2}{c}{ViT-B/32} & \multicolumn{2}{c}{Transformer} & \textbf{0.7748} & \textbf{0.7566} & \textbf{0.7387} & \textbf{0.7091} & \textbf{0.6861} & \textbf{0.7661} & \textbf{0.7516} & \textbf{0.7311} & \textbf{0.7030} & \textbf{0.6806} \\ \hline
			\multicolumn{6}{c}{} & \multicolumn{10}{c}{Weighted MAP} \\
			\multicolumn{2}{c}{\textbf{Albef}} & \multicolumn{2}{c}{ViT-B/16} & \multicolumn{2}{c}{BERT} & 0.9118 & 0.8795 & 0.8506 & 0.8103 & 0.7798 & 0.7303 & 0.7112 & 0.6781 & 0.6436 & 0.6239 \\
			\multicolumn{2}{c}{\textbf{Align}} & \multicolumn{2}{c}{EfficientNet-B7} & \multicolumn{2}{c}{BERT} & 0.9421 & 0.9106 & 0.8871 & 0.8470 & 0.8173 & 0.8720 & 0.8561 & 0.8347 & 0.8069 & 0.7853 \\
			\multicolumn{2}{c}{\multirow{2}{*}{\textbf{CLIP}}} & \multicolumn{2}{c}{RN50} & \multicolumn{2}{c}{\multirow{2}{*}{Transformer}} & \multicolumn{1}{l}{0.9089} & \multicolumn{1}{l}{0.8838} & \multicolumn{1}{l}{0.8590} & \multicolumn{1}{l}{0.8210} & \multicolumn{1}{l}{0.7928} & \multicolumn{1}{l}{0.9092} & \multicolumn{1}{l}{0.8870} & \multicolumn{1}{l}{0.8571} & \multicolumn{1}{l}{0.8178} & \multicolumn{1}{l}{0.7879} \\
			\multicolumn{2}{c}{} & \multicolumn{2}{c}{ViT-B/32} & \multicolumn{2}{c}{} & \multicolumn{1}{l}{0.9373} & \multicolumn{1}{l}{0.9135} & \multicolumn{1}{l}{0.8882} & \multicolumn{1}{l}{0.8514} & \multicolumn{1}{l}{0.8229} & \multicolumn{1}{l}{0.8943} & \multicolumn{1}{l}{0.8768} & \multicolumn{1}{l}{0.8543} & \multicolumn{1}{l}{0.8256} & \multicolumn{1}{l}{0.8027} \\
			\multicolumn{2}{c}{\textbf{Blip}} & \multicolumn{2}{c}{ViT-B/32} & \multicolumn{2}{c}{BERT} & 0.9191 & 0.8875 & 0.8489 & 0.8026 & 0.7701 & 0.8473 & 0.8229 & 0.7931 & 0.7543 & 0.7258 \\
			\multicolumn{2}{c}{\textbf{Filip}} & \multicolumn{2}{c}{ViT-B/32} & \multicolumn{2}{c}{Transformer} & 0.7910 & 0.7604 & 0.7341 & 0.7025 & 0.6848 & 0.7597 & 0.7383 & 0.7119 & 0.6834 & 0.6702 \\
			\multicolumn{2}{c}{\textbf{WuKong}} & \multicolumn{2}{c}{ViT-B/32} & \multicolumn{2}{c}{Transformer} & 0.8688 & 0.8483 & 0.8313 & 0.7980 & 0.7706 & 0.7987 & 0.7836 & 0.7643 & 0.7402 & 0.7251 \\
			\multicolumn{2}{c}{\textbf{GeoRSCLIP}} & \multicolumn{2}{c}{ViT-B/32} & \multicolumn{2}{c}{Transformer} & 0.9486 & 0.9272 & 0.9018 & 0.8649 & 0.8359 & 0.9262 & 0.9069 & 0.8823 & 0.8494 & 0.8242 \\
			\multicolumn{2}{c}{\multirow{2}{*}{\textbf{CISEN}}} & \multicolumn{2}{c}{RN50} & \multicolumn{2}{c}{\multirow{2}{*}{Transformer}} & \multicolumn{1}{l}{0.9136} & \multicolumn{1}{l}{0.8871} & \multicolumn{1}{l}{0.8619} & \multicolumn{1}{l}{0.8249} & \multicolumn{1}{l}{0.7967} & \multicolumn{1}{l}{0.8792} & \multicolumn{1}{l}{0.8573} & \multicolumn{1}{l}{0.8320} & \multicolumn{1}{l}{0.7998} & \multicolumn{1}{l}{0.7764} \\
			\multicolumn{2}{c}{} & \multicolumn{2}{c}{ViT-B/32} & \multicolumn{2}{c}{} & \multicolumn{1}{l}{0.9570} & \multicolumn{1}{l}{0.9298} & \multicolumn{1}{l}{0.9038} & \multicolumn{1}{l}{0.8639} & \multicolumn{1}{l}{0.8327} & \multicolumn{1}{l}{0.9246} & \multicolumn{1}{l}{0.9062} & \multicolumn{1}{l}{0.8822} & \multicolumn{1}{l}{0.8515} & \multicolumn{1}{l}{0.8246} \\
			\multicolumn{2}{c}{\textbf{CISEN (RS)}} & \multicolumn{2}{c}{ViT-B/32} & \multicolumn{2}{c}{Transformer} & \multicolumn{1}{l}{\textbf{0.9678}} & \multicolumn{1}{l}{\textbf{0.9407}} & \multicolumn{1}{l}{\textbf{0.9144}} & \multicolumn{1}{l}{\textbf{0.8725}} & \multicolumn{1}{l}{\textbf{0.8409}} & \multicolumn{1}{l}{\textbf{0.9407}} & \multicolumn{1}{l}{\textbf{0.9191}} & \multicolumn{1}{l}{\textbf{0.8925}} & \multicolumn{1}{l}{\textbf{0.8563}} & \multicolumn{1}{l}{\textbf{0.8274}} \\ \hline
			\multicolumn{6}{c}{} & \multicolumn{10}{c}{NDCG} \\
			\multicolumn{2}{c}{\textbf{Albef}} & \multicolumn{2}{c}{ViT-B/16} & \multicolumn{2}{c}{BERT} & 0.7491 & 0.7418 & 0.7397 & 0.7437 & 0.7536 & 0.6711 & 0.6763 & 0.6721 & 0.6764 & 0.6903 \\
			\multicolumn{2}{c}{\textbf{Align}} & \multicolumn{2}{c}{EfficientNet-B7} & \multicolumn{2}{c}{BERT} & 0.7617 & 0.7528 & 0.7518 & 0.7542 & 0.7640 & 0.7445 & 0.7407 & 0.7375 & 0.7407 & 0.7510 \\
			\multicolumn{2}{c}{\multirow{2}{*}{\textbf{CLIP}}} & \multicolumn{2}{c}{RN50} & \multicolumn{2}{c}{\multirow{2}{*}{Transformer}} & \multicolumn{1}{l}{0.7571} & \multicolumn{1}{l}{0.7494} & \multicolumn{1}{l}{0.7466} & \multicolumn{1}{l}{0.7484} & \multicolumn{1}{l}{0.7583} & 0.7548 & 0.7491 & 0.7439 & 0.7463 & 0.7556 \\
			\multicolumn{2}{c}{} & \multicolumn{2}{c}{ViT-B/32} & \multicolumn{2}{c}{} & \multicolumn{1}{l}{0.7713} & \multicolumn{1}{l}{0.7614} & \multicolumn{1}{l}{0.7575} & \multicolumn{1}{l}{0.7588} & \multicolumn{1}{l}{0.7685} & \multicolumn{1}{l}{0.7582} & \multicolumn{1}{l}{0.7527} & \multicolumn{1}{l}{0.7463} & \multicolumn{1}{l}{0.7496} & \multicolumn{1}{l}{0.7588} \\
			\multicolumn{2}{c}{\textbf{Blip}} & \multicolumn{2}{c}{ViT-B/32} & \multicolumn{2}{c}{BERT} & 0.7471 & 0.7401 & 0.7328 & 0.7348 & 0.7448 & 0.7162 & 0.7134 & 0.7103 & 0.7149 & 0.7264 \\
			\multicolumn{2}{c}{\textbf{Filip}} & \multicolumn{2}{c}{ViT-B/32} & \multicolumn{2}{c}{Transformer} & 0.6914 & 0.6877 & 0.6895 & 0.6955 & 0.7105 & 0.6792 & 0.6819 & 0.6812 & 0.6891 & 0.7041 \\
			\multicolumn{2}{c}{\textbf{WuKong}} & \multicolumn{2}{c}{ViT-B/32} & \multicolumn{2}{c}{Transformer} & 0.7452 & 0.7392 & 0.7395 & 0.7416 & 0.7511 & 0.7005 & 0.7006 & 0.7006 & 0.7078 & 0.7211 \\
			\multicolumn{2}{c}{\textbf{GeoRSCLIP}} & \multicolumn{2}{c}{ViT-B/32} & \multicolumn{2}{c}{Transformer} & 0.7739 & 0.7653 & 0.7612 & 0.7633 & 0.7735 & 0.7683 & 0.7619 & 0.7555 & 0.7577 & 0.7666 \\
			\multicolumn{2}{c}{\multirow{2}{*}{\textbf{CISEN}}} & \multicolumn{2}{c}{RN50} & \multicolumn{2}{c}{\multirow{2}{*}{Transformer}} & \multicolumn{1}{l}{0.7588} & \multicolumn{1}{l}{0.7504} & \multicolumn{1}{l}{0.7469} & \multicolumn{1}{l}{0.7487} & \multicolumn{1}{l}{0.7587} & \multicolumn{1}{l}{0.7480} & \multicolumn{1}{l}{0.7448} & \multicolumn{1}{l}{0.7404} & \multicolumn{1}{l}{0.7429} & \multicolumn{1}{l}{0.7524} \\
			\multicolumn{2}{c}{} & \multicolumn{2}{c}{ViT-B/32} & \multicolumn{2}{c}{} & \multicolumn{1}{l}{0.7771} & \multicolumn{1}{l}{0.7659} & \multicolumn{1}{l}{0.7631} & \multicolumn{1}{l}{0.7640} & \multicolumn{1}{l}{0.7729} & \multicolumn{1}{l}{0.7647} & \multicolumn{1}{l}{0.7577} & \multicolumn{1}{l}{0.7534} & \multicolumn{1}{l}{0.7568} & \multicolumn{1}{l}{0.7659} \\
			\multicolumn{2}{c}{\textbf{CISEN (RS)}} & \multicolumn{2}{c}{ViT-B/32} & \multicolumn{2}{c}{Transformer} & \multicolumn{1}{l}{\textbf{0.7822}} & \multicolumn{1}{l}{\textbf{0.7722}} & \multicolumn{1}{l}{\textbf{0.7684}} & \multicolumn{1}{l}{\textbf{0.7682}} & \multicolumn{1}{l}{\textbf{0.7773}} & \multicolumn{1}{l}{\textbf{0.7719}} & \multicolumn{1}{l}{\textbf{0.7659}} & \multicolumn{1}{l}{\textbf{0.7601}} & \multicolumn{1}{l}{\textbf{0.7608}} & \multicolumn{1}{l}{\textbf{0.7686}} \\ \hline
			\multicolumn{6}{c}{} & \multicolumn{10}{c}{ACG} \\
			\multicolumn{2}{c}{\textbf{Albef}} & \multicolumn{2}{c}{ViT-B/16} & \multicolumn{2}{c}{BERT} & 0.8021 & 0.7902 & 0.7793 & 0.7496 & 0.7191 & 0.6219 & 0.6098 & 0.6038 & 0.5980 & 0.5869 \\
			\multicolumn{2}{c}{\textbf{Align}} & \multicolumn{2}{c}{EfficientNet-B7} & \multicolumn{2}{c}{BERT} & 0.8375 & 0.8272 & 0.8178 & 0.7880 & 0.7552 & 0.7970 & 0.7913 & 0.7799 & 0.7610 & 0.7352 \\
			\multicolumn{2}{c}{\multirow{2}{*}{\textbf{CLIP}}} & \multicolumn{2}{c}{RN50} & \multicolumn{2}{c}{\multirow{2}{*}{Transformer}} & \multicolumn{1}{l}{0.8146} & \multicolumn{1}{l}{0.8048} & \multicolumn{1}{l}{0.7915} & \multicolumn{1}{l}{0.7630} & \multicolumn{1}{l}{0.7266} & \multicolumn{1}{l}{0.8189} & \multicolumn{1}{l}{0.8085} & \multicolumn{1}{l}{0.7905} & \multicolumn{1}{l}{0.7598} & \multicolumn{1}{l}{0.7267} \\
			\multicolumn{2}{c}{} & \multicolumn{2}{c}{ViT-B/32} & \multicolumn{2}{c}{} & \multicolumn{1}{l}{0.8455} & \multicolumn{1}{l}{0.8358} & \multicolumn{1}{l}{0.8239} & \multicolumn{1}{l}{0.7944} & \multicolumn{1}{l}{0.7563} & \multicolumn{1}{l}{0.8149} & \multicolumn{1}{l}{0.8096} & \multicolumn{1}{l}{0.8017} & \multicolumn{1}{l}{0.7792} & \multicolumn{1}{l}{0.7494} \\
			\multicolumn{2}{c}{\textbf{Blip}} & \multicolumn{2}{c}{ViT-B/32} & \multicolumn{2}{c}{BERT} & 0.8055 & 0.7877 & 0.7691 & 0.7391 & 0.7070 & 0.7450 & 0.7341 & 0.7201 & 0.6936 & 0.6664 \\
			\multicolumn{2}{c}{\textbf{Filip}} & \multicolumn{2}{c}{ViT-B/32} & \multicolumn{2}{c}{Transformer} & 0.6674 & 0.6658 & 0.6624 & 0.6583 & 0.6489 & 0.6462 & 0.6482 & 0.6471 & 0.6467 & 0.6421 \\
			\multicolumn{2}{c}{\textbf{WuKong}} & \multicolumn{2}{c}{ViT-B/32} & \multicolumn{2}{c}{Transformer} & 0.7756 & 0.7721 & 0.7686 & 0.7429 & 0.7079 & 0.7085 & 0.7125 & 0.7126 & 0.7033 & 0.6899 \\
			\multicolumn{2}{c}{\textbf{GeoRSCLIP}} & \multicolumn{2}{c}{ViT-B/32} & \multicolumn{2}{c}{Transformer} & 0.8611 & 0.8523 & 0.8385 & 0.8071 & 0.7664 & \multicolumn{1}{l}{0.8475} & \multicolumn{1}{l}{0.8380} & \multicolumn{1}{l}{0.8247} & \multicolumn{1}{l}{0.7972} & \multicolumn{1}{l}{0.7655} \\
			\multicolumn{2}{c}{\multirow{2}{*}{\textbf{CISEN}}} & \multicolumn{2}{c}{RN50} & \multicolumn{2}{c}{\multirow{2}{*}{Transformer}} & \multicolumn{1}{l}{0.8162} & \multicolumn{1}{l}{0.8061} & \multicolumn{1}{l}{0.7953} & \multicolumn{1}{l}{0.7672} & \multicolumn{1}{l}{0.7306} & \multicolumn{1}{l}{0.7896} & \multicolumn{1}{l}{0.7817} & \multicolumn{1}{l}{0.7724} & \multicolumn{1}{l}{0.7497} & \multicolumn{1}{l}{0.7228} \\
			\multicolumn{2}{c}{} & \multicolumn{2}{c}{ViT-B/32} & \multicolumn{2}{c}{} & \multicolumn{1}{l}{0.8627} & \multicolumn{1}{l}{0.8520} & \multicolumn{1}{l}{0.8369} & \multicolumn{1}{l}{0.8028} & \multicolumn{1}{l}{0.7614} & \multicolumn{1}{l}{0.8474} & \multicolumn{1}{l}{0.8377} & \multicolumn{1}{l}{0.8270} & \multicolumn{1}{l}{0.7980} & \multicolumn{1}{l}{0.7617} \\
			\multicolumn{2}{c}{\textbf{CISEN (RS)}} & \multicolumn{2}{c}{ViT-B/32} & \multicolumn{2}{c}{Transformer} & \multicolumn{1}{l}{\textbf{0.8740}} & \multicolumn{1}{l}{\textbf{0.8609}} & \multicolumn{1}{l}{\textbf{0.8453}} & \multicolumn{1}{l}{\textbf{0.8090}} & \multicolumn{1}{l}{\textbf{0.7669}} & \multicolumn{1}{l}{\textbf{0.8621}} & \multicolumn{1}{l}{\textbf{0.8466}} & \multicolumn{1}{l}{\textbf{0.8312}} & \multicolumn{1}{l}{\textbf{0.7989}} & \multicolumn{1}{l}{\textbf{0.7633}} \\ \hline
		\end{tabular}
		\label{retrieval 21}
	}
	\vspace{-1.2em}
\end{table*}

\subsubsection{Ablation Study on CISEN}
We conduct various ablation studies to understand the effectiveness of our method from different aspects, where CLIP with ViT-B is used as the backbone network.


\textbf{Effect of Different Components }  We study the effect of different components in our method, including dual encoder backbone, V2TMap and HFE. To evaluate the importance of these modules for ITR retrieval task, we implement the ablation study on LuojiaHOG with third-level labels by using the features extracted from these three modules. Tab.~\ref{ablation study}, we can see that: (1) The introduction of V2TMap and HFE brings the retrieval performance gain. When incorporated with RN50 as backbone, V2TMap can significantly improve at an average incremental of 3.8\% MAP, 5.1\% WMAP, 2.7\% NDCG and 5.2\% ACG on I2T retreival task, 4\% MAP, 8.1\% WMAP, 3.8\% NDCG and 6\% ACG on T2I retrieval task. When it comes to ViT as backbone, the retrieval results obtained by incorporating each of the two modules are quite similar. (2) Except utilizing CLIP (RN50) on T2I retrieval task, the combination of V2TMap and HFE gives further performance boost, which shows that our method is effective in learning and enhancing visual features via both V2TMap and HFE. (3) Although GeoRSCLIP-based model outperforming others across all metrics, the inclusion of the two modules narrows the performance gap. For instance, in terms of MAP@5 on I2T retrieval task, when RN50 is used in a zero-shot setting, it achieves only 0.4588, while GeoRSCLIP achieves 0.6597, resulting in a difference of 0.2009. However, after adding the two modules, this difference reduces to 0.0315, which indicating that our approach enables smaller models to achieve significant improvements, approaching the performance of larger models.\\

\textbf{Effect of Residual Ratio in V2TMap } In Eq.~\ref{residual ratio}, $\alpha$ balance the raw knowledge from pretrained backbone and new knowledge from V2TMap. To study the impacts of residual ratio $\alpha$ in V2TMap, we select CLIP (ViT-B) as backbone and conduct experiments with the residual ratio ranging from 0.1 to 0.9. Fig.~\ref{vit_21_itr} and Fig.~\ref{vit_131_itr} respectively illustrate the results at second-level and third-level of LuojiaHOG. The training performance of V2TMap (depicted by the blue lines) shows a gradual improvement as the ratio value increases, reaching its peak at 0.9. Afterwards, introducing HFE (depicted by the orange lines) during training yields varied outcomes. In I2T retrieval, the performance remains relatively stable across different ratio values. At level 2, fluctuations in MAP@5 and NDCG@5 are within 1\%, while those in WMAP@5 and ACG@5 are within 2.5\%. At level 3, fluctuations in ACG@5, NDCG@5, and MAP@5 are within 2\%, and within 4\% for WMAP@5. The best performance is achieved when the residual ratio is equal to 0.9. Overall, the better the feature representation obtained through V2TMap, the better the enhanced visual feature obtained from HFE in the end.\\

\begin{table*}[h!]
	\renewcommand{\arraystretch}{1.5}
	\captionsetup{justification=justified}
	\caption{Quantative performance comparison of all models in terms of MAP@n, Weighted MAP@n, NDCG@n and ACG@n (n=5, 10, 20 ,50 , 100) on LuojiaHOG third-level labels. The best is marked in bold.}
	\centering
	\resizebox{1.6\columnwidth}{!}{
		\begin{tabular}{clclclcccccccccc}
			\hline
			\multicolumn{2}{c}{\multirow{2}{*}{\textbf{Methods}}} & \multicolumn{2}{c}{\multirow{2}{*}{\textbf{Image Encoder}}} & \multicolumn{2}{c}{\multirow{2}{*}{\textbf{Text Encoder}}} & \multicolumn{5}{c}{Image To Text} & \multicolumn{5}{c}{Text To Image} \\
			\multicolumn{2}{c}{} & \multicolumn{2}{c}{} & \multicolumn{2}{c}{} & @5 & @10 & @20 & @50 & @100 & @5 & @10 & @20 & @50 & @100 \\ \hline
			\multicolumn{6}{c}{} & \multicolumn{10}{c}{MAP} \\
			\multicolumn{2}{c}{\textbf{Albef}} & \multicolumn{2}{c}{ViT-B/16} & \multicolumn{2}{c}{BERT} & 0.6620 & 0.6442 & 0.6237 & 0.5903 & 0.5657 & 0.5481 & 0.5342 & 0.5057 & 0.4719 & 0.4527 \\
			\multicolumn{2}{c}{\textbf{Align}} & \multicolumn{2}{c}{EfficientNet-B7} & \multicolumn{2}{c}{BERT} & 0.6855 & 0.6656 & 0.6473 & 0.6123 & 0.5873 & 0.6625 & 0.6476 & 0.6258 & 0.5975 & 0.5758 \\
			\multicolumn{2}{c}{\multirow{2}{*}{\textbf{CLIP}}} & \multicolumn{2}{c}{RN50} & \multicolumn{2}{c}{\multirow{2}{*}{Transformer}} & \multicolumn{1}{l}{0.6817} & \multicolumn{1}{l}{0.6642} & \multicolumn{1}{l}{0.6432} & \multicolumn{1}{l}{0.6105} & \multicolumn{1}{l}{0.5878} & \multicolumn{1}{l}{0.6873} & \multicolumn{1}{l}{0.6688} & \multicolumn{1}{l}{0.6447} & \multicolumn{1}{l}{0.6125} & \multicolumn{1}{l}{0.5877} \\
			\multicolumn{2}{c}{} & \multicolumn{2}{c}{ViT-B/32} & \multicolumn{2}{c}{} & \multicolumn{1}{l}{0.6934} & \multicolumn{1}{l}{0.6743} & \multicolumn{1}{l}{0.6515} & \multicolumn{1}{l}{0.6175} & \multicolumn{1}{l}{0.5927} & \multicolumn{1}{l}{0.6900} & \multicolumn{1}{l}{0.6737} & \multicolumn{1}{l}{0.6509} & \multicolumn{1}{l}{0.6195} & \multicolumn{1}{l}{0.5952} \\
			\multicolumn{2}{c}{\textbf{Blip}} & \multicolumn{2}{c}{ViT-B/32} & \multicolumn{2}{c}{BERT} & 0.6697 & 0.6478 & 0.6179 & 0.5799 & 0.5542 & 0.6221 & 0.6053 & 0.5801 & 0.5478 & 0.5248 \\
			\multicolumn{2}{c}{\textbf{Filip}} & \multicolumn{2}{c}{ViT-B/32} & \multicolumn{2}{c}{Transformer} & 0.5771 & 0.5550 & 0.5309 & 0.5011 & 0.4842 & 0.5672 & 0.5512 & 0.5290 & 0.5004 & 0.4857 \\
			\multicolumn{2}{c}{\textbf{WuKong}} & \multicolumn{2}{c}{ViT-B/32} & \multicolumn{2}{c}{Transformer} & 0.6531 & 0.6386 & 0.6236 & 0.5921 & 0.5673 & 0.5986 & 0.5846 & 0.5638 & 0.5386 & 0.5220 \\
			\multicolumn{2}{c}{\textbf{GeoRSCLIP}} & \multicolumn{2}{c}{ViT-B/32} & \multicolumn{2}{c}{Transformer} & 0.6942 & 0.6755 & 0.6547 & 0.6222 & 0.5983 & 0.6989 & 0.6823 & 0.6584 & 0.6263 & 0.6017 \\
			\multicolumn{2}{c}{\multirow{2}{*}{\textbf{CISEN}}} & \multicolumn{2}{c}{RN50} & \multicolumn{2}{c}{\multirow{2}{*}{Transformer}} & \multicolumn{1}{l}{0.6854} & \multicolumn{1}{l}{0.6695} & \multicolumn{1}{l}{0.6498} & \multicolumn{1}{l}{0.6167} & \multicolumn{1}{l}{0.5937} & \multicolumn{1}{l}{0.6690} & \multicolumn{1}{l}{0.6572} & \multicolumn{1}{l}{0.6378} & \multicolumn{1}{l}{0.6097} & \multicolumn{1}{l}{0.5869} \\
			\multicolumn{2}{c}{} & \multicolumn{2}{c}{ViT-B/32} & \multicolumn{2}{c}{} & \multicolumn{1}{l}{0.6983} & \multicolumn{1}{l}{0.6798} & \multicolumn{1}{l}{0.6571} & \multicolumn{1}{l}{0.6231} & \multicolumn{1}{l}{0.5977} & \multicolumn{1}{l}{0.6940} & \multicolumn{1}{l}{0.6789} & \multicolumn{1}{l}{0.6578} & \multicolumn{1}{l}{0.6278} & \multicolumn{1}{l}{0.6027} \\
			\multicolumn{2}{c}{\textbf{CISEN (RS)}} & \multicolumn{2}{c}{ViT-B/32} & \multicolumn{2}{c}{Transformer} & \textbf{0.7112} & \textbf{0.6906} & \textbf{0.6684} & \textbf{0.6337} & \textbf{0.6083} & \textbf{0.7037} & \textbf{0.6881} & \textbf{0.6661} & \textbf{0.6349} & \textbf{0.6093} \\ \hline
			\multicolumn{6}{c}{} & \multicolumn{10}{c}{Weighted MAP} \\
			\multicolumn{2}{c}{\textbf{Albef}} & \multicolumn{2}{c}{ViT-B/16} & \multicolumn{2}{c}{BERT} & 0.7959 & 0.7703 & 0.7423 & 0.6982 & 0.6651 & 0.6194 & 0.6065 & 0.5777 & 0.5427 & 0.5222 \\
			\multicolumn{2}{c}{\textbf{Align}} & \multicolumn{2}{c}{EfficientNet-B7} & \multicolumn{2}{c}{BERT} & 0.8388 & 0.8099 & 0.7831 & 0.7350 & 0.6998 & 0.7794 & 0.7617 & 0.7358 & 0.7019 & 0.6751 \\
			\multicolumn{2}{c}{\multirow{2}{*}{\textbf{CLIP}}} & \multicolumn{2}{c}{RN50} & \multicolumn{2}{c}{\multirow{2}{*}{Transformer}} & \multicolumn{1}{l}{0.8194} & \multicolumn{1}{l}{0.7943} & \multicolumn{1}{l}{0.7665} & \multicolumn{1}{l}{0.7242} & \multicolumn{1}{l}{0.6939} & \multicolumn{1}{l}{0.8206} & \multicolumn{1}{l}{0.7965} & \multicolumn{1}{l}{0.7658} & \multicolumn{1}{l}{0.7263} & \multicolumn{1}{l}{0.6947} \\
			\multicolumn{2}{c}{} & \multicolumn{2}{c}{ViT-B/32} & \multicolumn{2}{c}{} & \multicolumn{1}{l}{0.8431} & \multicolumn{1}{l}{0.8161} & \multicolumn{1}{l}{0.7857} & \multicolumn{1}{l}{0.7396} & \multicolumn{1}{l}{0.7054} & \multicolumn{1}{l}{0.8286} & \multicolumn{1}{l}{0.8078} & \multicolumn{1}{l}{0.7783} & \multicolumn{1}{l}{0.7380} & \multicolumn{1}{l}{0.7060} \\
			\multicolumn{2}{c}{\textbf{Blip}} & \multicolumn{2}{c}{ViT-B/32} & \multicolumn{2}{c}{BERT} & 0.8428 & 0.8081 & 0.7640 & 0.7082 & 0.6700 & 0.7583 & 0.7340 & 0.6994 & 0.6555 & 0.6245 \\
			\multicolumn{2}{c}{\textbf{Filip}} & \multicolumn{2}{c}{ViT-B/32} & \multicolumn{2}{c}{Transformer} & 0.6630 & 0.6400 & 0.6144 & 0.5822 & 0.5640 & 0.6544 & 0.6361 & 0.6115 & 0.5808 & 0.5654 \\
			\multicolumn{2}{c}{\textbf{WuKong}} & \multicolumn{2}{c}{ViT-B/32} & \multicolumn{2}{c}{Transformer} & 0.7618 & 0.7449 & 0.7269 & 0.6896 & 0.6592 & 0.6896 & 0.6749 & 0.6522 & 0.6251 & 0.6070 \\
			\multicolumn{2}{c}{\textbf{GeoRSCLIP}} & \multicolumn{2}{c}{ViT-B/32} & \multicolumn{2}{c}{Transformer} & 0.8592 & 0.8304 & 0.7993 & 0.7530 & 0.7180 & 0.8490 & 0.8258 & 0.7929 & 0.7499 & 0.7163 \\
			\multicolumn{2}{c}{\multirow{2}{*}{\textbf{CISEN}}} & \multicolumn{2}{c}{RN50} & \multicolumn{2}{c}{\multirow{2}{*}{Transformer}} & \multicolumn{1}{l}{0.8295} & \multicolumn{1}{l}{0.8057} & \multicolumn{1}{l}{0.7785} & \multicolumn{1}{l}{0.7352} & \multicolumn{1}{l}{0.7040} & \multicolumn{1}{l}{0.7936} & \multicolumn{1}{l}{0.7802} & \multicolumn{1}{l}{0.7572} & \multicolumn{1}{l}{0.7230} & \multicolumn{1}{l}{0.6940} \\
			\multicolumn{2}{c}{} & \multicolumn{2}{c}{ViT-B/32} & \multicolumn{2}{c}{} & \multicolumn{1}{l}{0.8633} & \multicolumn{1}{l}{0.8336} & \multicolumn{1}{l}{0.8003} & \multicolumn{1}{l}{0.7521} & \multicolumn{1}{l}{0.7155} & \multicolumn{1}{l}{0.8440} & \multicolumn{1}{l}{0.8238} & \multicolumn{1}{l}{0.7956} & \multicolumn{1}{l}{0.7555} & \multicolumn{1}{l}{0.7209} \\
			\multicolumn{2}{c}{\textbf{CISEN (RS)}} & \multicolumn{2}{c}{ViT-B/32} & \multicolumn{2}{c}{Transformer} & \textbf{0.8847} & \textbf{0.8523} & \textbf{0.8196} & \textbf{0.7689} & \textbf{0.7313} & \textbf{0.8728} & \textbf{0.8499} & \textbf{0.8183} & \textbf{0.7727} & \textbf{0.7352} \\ \hline
			\multicolumn{6}{c}{} & \multicolumn{10}{c}{NDCG} \\
			\multicolumn{2}{c}{\textbf{Albef}} & \multicolumn{2}{c}{ViT-B/16} & \multicolumn{2}{c}{BERT} & 0.6874 & 0.6854 & 0.6820 & 0.6781 & 0.6820 & 0.5938 & 0.6064 & 0.6052 & 0.6051 & 0.6128 \\
			\multicolumn{2}{c}{\textbf{Align}} & \multicolumn{2}{c}{EfficientNet-B7} & \multicolumn{2}{c}{BERT} & 0.7042 & 0.6977 & 0.6938 & 0.6873 & 0.6903 & 0.6819 & 0.6799 & 0.6729 & 0.6701 & 0.6758 \\
			\multicolumn{2}{c}{\multirow{2}{*}{\textbf{CLIP}}} & \multicolumn{2}{c}{RN50} & \multicolumn{2}{c}{\multirow{2}{*}{Transformer}} & \multicolumn{1}{l}{0.7019} & \multicolumn{1}{l}{0.6979} & \multicolumn{1}{l}{0.6909} & \multicolumn{1}{l}{0.6844} & \multicolumn{1}{l}{0.6885} & \multicolumn{1}{l}{0.7069} & \multicolumn{1}{l}{0.7014} & \multicolumn{1}{l}{0.6923} & \multicolumn{1}{l}{0.6859} & \multicolumn{1}{l}{0.6862} \\
			\multicolumn{2}{c}{} & \multicolumn{2}{c}{ViT-B/32} & \multicolumn{2}{c}{} & \multicolumn{1}{l}{0.7122} & \multicolumn{1}{l}{0.7051} & \multicolumn{1}{l}{0.6978} & \multicolumn{1}{l}{0.6915} & \multicolumn{1}{l}{0.6943} & \multicolumn{1}{l}{0.7069} & \multicolumn{1}{l}{0.7018} & \multicolumn{1}{l}{0.6927} & \multicolumn{1}{l}{0.6875} & \multicolumn{1}{l}{0.6893} \\
			\multicolumn{2}{c}{\textbf{Blip}} & \multicolumn{2}{c}{ViT-B/32} & \multicolumn{2}{c}{BERT} & 0.6911 & 0.6850 & 0.6746 & 0.6675 & 0.6718 & 0.6506 & 0.6514 & 0.6449 & 0.6428 & 0.6500 \\
			\multicolumn{2}{c}{\textbf{Filip}} & \multicolumn{2}{c}{ViT-B/32} & \multicolumn{2}{c}{Transformer} & 0.6180 & 0.6225 & 0.6252 & 0.6261 & 0.6341 & 0.6103 & 0.6174 & 0.6186 & 0.6207 & 0.6294 \\
			\multicolumn{2}{c}{\textbf{WuKong}} & \multicolumn{2}{c}{ViT-B/32} & \multicolumn{2}{c}{Transformer} & 0.6773 & 0.6778 & 0.6792 & 0.6760 & 0.6796 & 0.6334 & 0.6383 & 0.6372 & 0.6399 & 0.6466 \\
			\multicolumn{2}{c}{\textbf{GeoRSCLIP}} & \multicolumn{2}{c}{ViT-B/32} & \multicolumn{2}{c}{Transformer} & 0.7155 & 0.7073 & 0.6986 & 0.6929 & 0.6972 & 0.7118 & 0.7073 & 0.6968 & 0.6907 & 0.6933 \\
			\multicolumn{2}{c}{\multirow{2}{*}{\textbf{CISEN}}} & \multicolumn{2}{c}{RN50} & \multicolumn{2}{c}{\multirow{2}{*}{Transformer}} & \multicolumn{1}{l}{0.7024} & \multicolumn{1}{l}{0.6981} & \multicolumn{1}{l}{0.6926} & \multicolumn{1}{l}{0.6866} & \multicolumn{1}{l}{0.6907} & \multicolumn{1}{l}{0.6858} & \multicolumn{1}{l}{0.6850} & \multicolumn{1}{l}{0.6795} & \multicolumn{1}{l}{0.6790} & \multicolumn{1}{l}{0.6818} \\
			\multicolumn{2}{c}{} & \multicolumn{2}{c}{ViT-B/32} & \multicolumn{2}{c}{} & \multicolumn{1}{l}{0.7177} & \multicolumn{1}{l}{0.7100} & \multicolumn{1}{l}{0.7013} & \multicolumn{1}{l}{0.6944} & \multicolumn{1}{l}{0.6978} & \multicolumn{1}{l}{0.7083} & \multicolumn{1}{l}{0.7047} & \multicolumn{1}{l}{0.6958} & \multicolumn{1}{l}{0.6924} & \multicolumn{1}{l}{0.6941} \\
			\multicolumn{2}{c}{\textbf{CISEN (RS)}} & \multicolumn{2}{c}{ViT-B/32} & \multicolumn{2}{c}{Transformer} & \textbf{0.7274} & \textbf{0.7157} & \textbf{0.7073} & \textbf{0.6999} & \textbf{0.7029} & \textbf{0.7157} & \textbf{0.7100} & \textbf{0.7025} & \textbf{0.6969} & \textbf{0.6992} \\ \hline
			\multicolumn{6}{c}{} & \multicolumn{10}{c}{ACG} \\
			\multicolumn{2}{c}{\textbf{Albef}} & \multicolumn{2}{c}{ViT-B/16} & \multicolumn{2}{c}{BERT} & 0.6828 & 0.6697 & 0.6590 & 0.6257 & 0.5948 & 0.5115 & 0.5041 & 0.5010 & 0.4932 & 0.4835 \\
			\multicolumn{2}{c}{\textbf{Align}} & \multicolumn{2}{c}{EfficientNet-B7} & \multicolumn{2}{c}{BERT} & 0.7221 & 0.7084 & 0.6969 & 0.6588 & 0.6234 & 0.6894 & 0.6808 & 0.6678 & 0.6431 & 0.6127 \\
			\multicolumn{2}{c}{\multirow{2}{*}{\textbf{CLIP}}} & \multicolumn{2}{c}{RN50} & \multicolumn{2}{c}{\multirow{2}{*}{Transformer}} & \multicolumn{1}{l}{0.7123} & \multicolumn{1}{l}{0.7005} & \multicolumn{1}{l}{0.6858} & \multicolumn{1}{l}{0.6554} & \multicolumn{1}{l}{0.6189} & \multicolumn{1}{l}{0.7174} & \multicolumn{1}{l}{0.7074} & \multicolumn{1}{l}{0.6953} & \multicolumn{1}{l}{0.6616} & \multicolumn{1}{l}{0.6256} \\
			\multicolumn{2}{c}{} & \multicolumn{2}{c}{ViT-B/32} & \multicolumn{2}{c}{} & \multicolumn{1}{l}{0.7309} & \multicolumn{1}{l}{0.7192} & \multicolumn{1}{l}{0.7015} & \multicolumn{1}{l}{0.6645} & \multicolumn{1}{l}{0.6237} & \multicolumn{1}{l}{0.7321} & \multicolumn{1}{l}{0.7204} & \multicolumn{1}{l}{0.7039} & \multicolumn{1}{l}{0.6716} & \multicolumn{1}{l}{0.6325} \\
			\multicolumn{2}{c}{\textbf{Blip}} & \multicolumn{2}{c}{ViT-B/32} & \multicolumn{2}{c}{BERT} & 0.7106 & 0.6887 & 0.6648 & 0.6291 & 0.5949 & 0.6410 & 0.6259 & 0.6125 & 0.5849 & 0.5581 \\
			\multicolumn{2}{c}{\textbf{Filip}} & \multicolumn{2}{c}{ViT-B/32} & \multicolumn{2}{c}{Transformer} & 0.5366 & 0.5331 & 0.5352 & 0.5333 & 0.5264 & 0.5381 & 0.5403 & 0.5397 & 0.5383 & 0.5322 \\
			\multicolumn{2}{c}{\textbf{WuKong}} & \multicolumn{2}{c}{ViT-B/32} & \multicolumn{2}{c}{Transformer} & 0.6647 & 0.6585 & 0.6525 & 0.6227 & 0.5872 & 0.5895 & 0.5902 & 0.5893 & 0.5806 & 0.5655 \\
			\multicolumn{2}{c}{\textbf{GeoRSCLIP}} & \multicolumn{2}{c}{ViT-B/32} & \multicolumn{2}{c}{Transformer} & 0.7451 & 0.7323 & 0.7138 & 0.6756 & 0.6324 & 0.7502 & 0.7335 & 0.7147 & 0.6783 & 0.6378 \\
			\multicolumn{2}{c}{\multirow{2}{*}{\textbf{CISEN}}} & \multicolumn{2}{c}{RN50} & \multicolumn{2}{c}{\multirow{2}{*}{Transformer}} & \multicolumn{1}{l}{0.7257} & \multicolumn{1}{l}{0.7115} & \multicolumn{1}{l}{0.6971} & \multicolumn{1}{l}{0.6644} & \multicolumn{1}{l}{0.6269} & \multicolumn{1}{l}{0.7099} & \multicolumn{1}{l}{0.7049} & \multicolumn{1}{l}{0.6931} & \multicolumn{1}{l}{0.6613} & \multicolumn{1}{l}{0.6268} \\
			\multicolumn{2}{c}{} & \multicolumn{2}{c}{ViT-B/32} & \multicolumn{2}{c}{} & \multicolumn{1}{l}{0.7495} & \multicolumn{1}{l}{0.7336} & \multicolumn{1}{l}{0.7134} & \multicolumn{1}{l}{0.6728} & \multicolumn{1}{l}{0.6278} & \multicolumn{1}{l}{0.7513} & \multicolumn{1}{l}{0.7415} & \multicolumn{1}{l}{0.7242} & \multicolumn{1}{l}{0.6846} & \multicolumn{1}{l}{0.6393} \\
			\multicolumn{2}{c}{\textbf{CISEN (RS)}} & \multicolumn{2}{c}{ViT-B/32} & \multicolumn{2}{c}{Transformer} & \textbf{0.7691} & \textbf{0.7508} & \textbf{0.7292} & \textbf{0.6862} & \textbf{0.6399} & \textbf{0.7783} & \textbf{0.7612} & \textbf{0.7374} & \textbf{0.6955} & \textbf{0.6491} \\ \hline
		\end{tabular}
	}
	\vspace{-0.9em}
	\label{retrieval 131}
\end{table*}

\begin{table*}[h!]
	\renewcommand{\arraystretch}{1.5}
	\captionsetup{justification=justified}
	\caption{Ablation study on LuojiaHOG where Model represents backbone feature extractor, V2TMap represents visual-to-text mapping and HFE represents Hierarchie feature enhancement. The best is in bold and the second best is underlined.}
	\centering
	\resizebox{1.8\columnwidth}{!}{
		\begin{tabular}{clclclcccccccccccc}
			\hline
			\multicolumn{2}{c}{\multirow{2}{*}{Backbone}} & \multicolumn{2}{c}{\multirow{2}{*}{V2TMap}} & \multicolumn{2}{c}{\multirow{2}{*}{HFE}} & \multicolumn{12}{c}{Image To Text} \\
			\multicolumn{2}{c}{} & \multicolumn{2}{c}{} & \multicolumn{2}{c}{} & MAP@5 & MAP@20 & MAP@100 & WMAP@5 & WMAP@20 & WMAP@100 & NDCG@5 & NDCG@20 & NDCG@100 & ACG@5 & ACG@20 & ACG@100 \\ \hline
			\multicolumn{2}{c}{\multirow{4}{*}{CLIP (RN50)}} & \multicolumn{2}{c}{\XSolid} & \multicolumn{2}{c}{\XSolid} & 0.4588 & 0.4202 & 0.3690 & 0.5293 & 0.4841 & 0.4257 & 0.5091 & 0.5440 & 0.5675 & 0.4001 & 0.3948 & 0.3916 \\
			\multicolumn{2}{c}{} & \multicolumn{2}{c}{\Checkmark} & \multicolumn{2}{c}{\XSolid} & {\ul 0.7411} & {\ul 0.7036} & {\ul 0.6519} & {\ul 0.9089} & {\ul 0.8590} & {\ul 0.7928} & {\ul 0.7571} & {\ul 0.7466} & {\ul 0.7583} & {\ul 0.8146} & {\ul 0.7915} & {\ul 0.7266} \\
			\multicolumn{2}{c}{} & \multicolumn{2}{c}{\XSolid} & \multicolumn{2}{c}{\Checkmark} & 0.7031 & 0.6624 & 0.6150 & 0.8583 & 0.8043 & 0.7444 & 0.7261 & 0.7207 & 0.7358 & 0.7536 & 0.7346 & 0.6907 \\
			\multicolumn{2}{c}{} & \multicolumn{2}{c}{\Checkmark} & \multicolumn{2}{c}{\Checkmark} & \textbf{0.7433} & \textbf{0.7049} & \textbf{0.6533} & \textbf{0.9136} & \textbf{0.8619} & \textbf{0.7967} & \textbf{0.7588} & \textbf{0.7469} & \textbf{0.7587} & \textbf{0.8162} & \textbf{0.7953} & \textbf{0.7306} \\ \cline{7-18}
			\multicolumn{2}{c}{\multirow{4}{*}{CLIP (ViT)}} & \multicolumn{2}{c}{\XSolid} & \multicolumn{2}{c}{\XSolid} & 0.6546 & 0.6194 & 0.5903 & 0.7869 & 0.7425 & 0.7140 & 0.6994 & 0.6989 & 0.7250 & 0.6644 & 0.6896 & 0.6799 \\
			\multicolumn{2}{c}{} & \multicolumn{2}{c}{\Checkmark} & \multicolumn{2}{c}{\XSolid} & 0.7606 & 0.7229 & {\ul 0.6725} & {\ul 0.9373} & {\ul 0.8882} & {\ul 0.8229} & 0.7713 & 0.7575 & {\ul 0.7685} & {\ul 0.8455} & {\ul 0.8239} & {\ul 0.7563} \\
			\multicolumn{2}{c}{} & \multicolumn{2}{c}{\XSolid} & \multicolumn{2}{c}{\Checkmark} & {\ul 0.7630} & {\ul 0.7238} & 0.6721 & 0.9351 & 0.8837 & 0.8175 & {\ul 0.7740} & {\ul 0.7580} & {\ul 0.7685} & 0.8385 & 0.8160 & 0.7503 \\
			\multicolumn{2}{c}{} & \multicolumn{2}{c}{\Checkmark} & \multicolumn{2}{c}{\Checkmark} & \textbf{0.7698} & \textbf{0.7318} & \textbf{0.6795} & \textbf{0.9570} & \textbf{0.9038} & \textbf{0.8327} & \textbf{0.7771} & \textbf{0.7631} & \textbf{0.7729} & \textbf{0.8627} & \textbf{0.8369} & \textbf{0.7614} \\ \cline{7-18}
			\multicolumn{2}{c}{\multirow{4}{*}{GeoRSCLIP (ViT)}} & \multicolumn{2}{c}{\XSolid} & \multicolumn{2}{c}{\XSolid} & 0.6597 & 0.6432 & 0.6116 & 0.8051 & 0.7832 & 0.7444 & 0.6979 & 0.7092 & 0.7351 & 0.7117 & 0.7337 & 0.6996 \\
			\multicolumn{2}{c}{} & \multicolumn{2}{c}{\Checkmark} & \multicolumn{2}{c}{\XSolid} & 0.7667 & 0.7318 & {\ul 0.6826} & 0.9486 & 0.9018 & {\ul 0.8359} & 0.7739 & 0.7612 & 0.7735 & 0.8611 & {\ul 0.8385} & {\ul 0.7664} \\
			\multicolumn{2}{c}{} & \multicolumn{2}{c}{\XSolid} & \multicolumn{2}{c}{\Checkmark} & {\ul 0.7740} & {\ul 0.7342} & 0.6817 & {\ul 0.9589} & {\ul 0.9050} & 0.8334 & {\ul 0.7797} & {\ul 0.7638} & {\ul 0.7745} & {\ul 0.8636} & 0.8356 & 0.7634 \\
			\multicolumn{2}{c}{} & \multicolumn{2}{c}{\Checkmark} & \multicolumn{2}{c}{\Checkmark} & \textbf{0.7748} & \textbf{0.7387} & \textbf{0.6861} & \textbf{0.9678} & \textbf{0.9144} & \textbf{0.8409} & \textbf{0.7822} & \textbf{0.7684} & \textbf{0.7773} & \textbf{0.8740} & \textbf{0.8453} & \textbf{0.7669} \\ \hline
			\multicolumn{2}{c}{} & \multicolumn{2}{c}{} & \multicolumn{2}{c}{} & \multicolumn{12}{c}{Text To Image} \\
			\multicolumn{2}{c}{} & \multicolumn{2}{c}{} & \multicolumn{2}{c}{} & MAP@5 & MAP@20 & MAP@100 & WMAP@5 & WMAP@20 & WMAP@100 & NDCG@5 & NDCG@20 & NDCG@100 & ACG@5 & ACG@20 & ACG@100 \\ \hline
			\multicolumn{2}{c}{\multirow{4}{*}{CLIP (RN50)}} & \multicolumn{2}{c}{\XSolid} & \multicolumn{2}{c}{\XSolid} & 0.4594 & 0.4341 & 0.3922 & 0.5140 & 0.4909 & 0.4499 & 0.5183 & 0.5585 & 0.5839 & 0.4005 & 0.4221 & 0.4163 \\
			\multicolumn{2}{c}{} & \multicolumn{2}{c}{\Checkmark} & \multicolumn{2}{c}{\XSolid} & \textbf{0.7468} & \textbf{0.7046} & \textbf{0.6486} & \textbf{0.9092} & \textbf{0.8571} & \textbf{0.7879} & \textbf{0.7612} & \textbf{0.7505} & \textbf{0.7567} & \textbf{0.8189} & \textbf{0.7905} & \textbf{0.7267} \\
			\multicolumn{2}{c}{} & \multicolumn{2}{c}{\XSolid} & \multicolumn{2}{c}{\Checkmark} & 0.6885 & 0.6635 & 0.6280 & 0.7961 & 0.7732 & 0.7415 & 0.7077 & 0.7132 & 0.7326 & 0.7234 & 0.7268 & 0.7038 \\
			\multicolumn{2}{c}{} & \multicolumn{2}{c}{\Checkmark} & \multicolumn{2}{c}{\Checkmark} & {\ul 0.7306} & {\ul 0.6930} & {\ul 0.6457} & {\ul 0.8792} & {\ul 0.8320} & {\ul 0.7764} & {\ul 0.7480} & {\ul 0.7404} & {\ul 0.7524} & {\ul 0.7896} & {\ul 0.7724} & {\ul 0.7228} \\ \cline{7-18}
			\multicolumn{2}{c}{\multirow{4}{*}{CLIP (ViT)}} & \multicolumn{2}{c}{\XSolid} & \multicolumn{2}{c}{\XSolid} & 0.7144 & 0.6774 & 0.6304 & 0.8644 & 0.8230 & 0.7705 & 0.7319 & 0.7266 & 0.7413 & 0.7798 & 0.7680 & 0.7201 \\
			\multicolumn{2}{c}{} & \multicolumn{2}{c}{\Checkmark} & \multicolumn{2}{c}{\XSolid} & 0.7460 & 0.7113 & 0.6661 & 0.8943 & 0.8543 & 0.8027 & {\ul 0.7582} & 0.7463 & 0.7588 & 0.8149 & 0.8017 & 0.7494 \\
			\multicolumn{2}{c}{} & \multicolumn{2}{c}{\XSolid} & \multicolumn{2}{c}{\Checkmark} & {\ul 0.7481} & {\ul 0.7131} & {\ul 0.6663} & {\ul 0.9076} & {\ul 0.8668} & {\ul 0.8111} & 0.7576 & {\ul 0.7485} & {\ul 0.7611} & {\ul 0.8277} & {\ul 0.8099} & {\ul 0.7540} \\
			\multicolumn{2}{c}{} & \multicolumn{2}{c}{\Checkmark} & \multicolumn{2}{c}{\Checkmark} & \textbf{0.7575} & \textbf{0.7221} & \textbf{0.6752} & \textbf{0.9246} & \textbf{0.8822} & \textbf{0.8246} & \textbf{0.7647} & \textbf{0.7534} & \textbf{0.7659} & \textbf{0.8474} & \textbf{0.8270} & \textbf{0.7617} \\ \cline{7-18}
			\multicolumn{2}{c}{\multirow{4}{*}{GeoRSCLIP (ViT)}} & \multicolumn{2}{c}{\XSolid} & \multicolumn{2}{c}{\XSolid} & 0.7335 & 0.6977 & 0.6537 & 0.9011 & 0.8589 & 0.8066 & 0.7454 & 0.7383 & 0.7539 & 0.8158 & 0.8029 & 0.7515 \\
			\multicolumn{2}{c}{} & \multicolumn{2}{c}{\Checkmark} & \multicolumn{2}{c}{\XSolid} & {\ul 0.7616} & {\ul 0.7263} & 0.6781 & {\ul 0.9262} & 0.8823 & 0.8242 & {\ul 0.7683} & 0.7555 & 0.7666 & {\ul 0.8475} & 0.8247 & {\ul 0.7655} \\
			\multicolumn{2}{c}{} & \multicolumn{2}{c}{\XSolid} & \multicolumn{2}{c}{\Checkmark} & 0.7558 & 0.7234 & {\ul 0.6789} & 0.9230 & {\ul 0.8828} & \textbf{0.8296} & 0.7649 & {\ul 0.7558} & {\ul 0.7684} & 0.8424 & {\ul 0.8286} & \textbf{0.7700} \\
			\multicolumn{2}{c}{} & \multicolumn{2}{c}{\Checkmark} & \multicolumn{2}{c}{\Checkmark} & \textbf{0.7661} & \textbf{0.7311} & \textbf{0.6806} & \textbf{0.9407} & \textbf{0.8925} & {\ul 0.8274} & \textbf{0.7719} & \textbf{0.7601} & \textbf{0.7686} & \textbf{0.8621} & \textbf{0.8312} & 0.7633 \\ \hline
		\end{tabular}
		\label{ablation study}
	}
\vspace{-1.2em}
\end{table*}

\begin{figure*}[htbp]
	\centering
	\captionsetup{justification=justified}
	\vspace{-2cm}
	\includegraphics[width=0.8\textwidth]{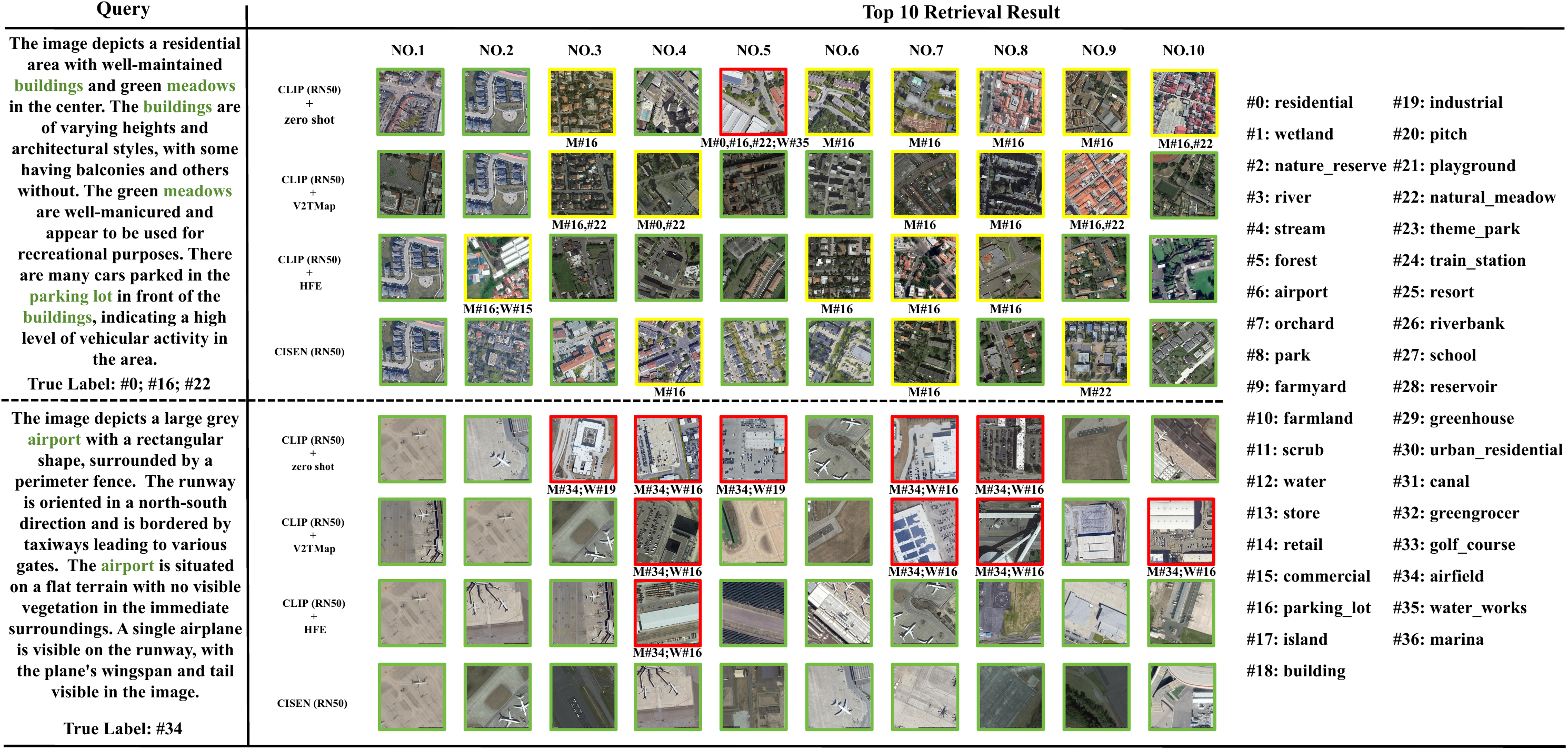}
	\caption{The T2I retrieval results of top 10 within LuojiaHOG, leveraging the integration of V2TMap and HFE with CLIP (RN50). The results with red box are incorrect, and with yellow box are inaccurate. At the bottom are some labels of third-level.}
	\label{RN50 ti retrieval results}
\end{figure*}

\begin{figure*}[htbp]
	\centering
	\captionsetup{justification=justified}
	\vspace{-2cm}
	\includegraphics[width=0.8\textwidth]{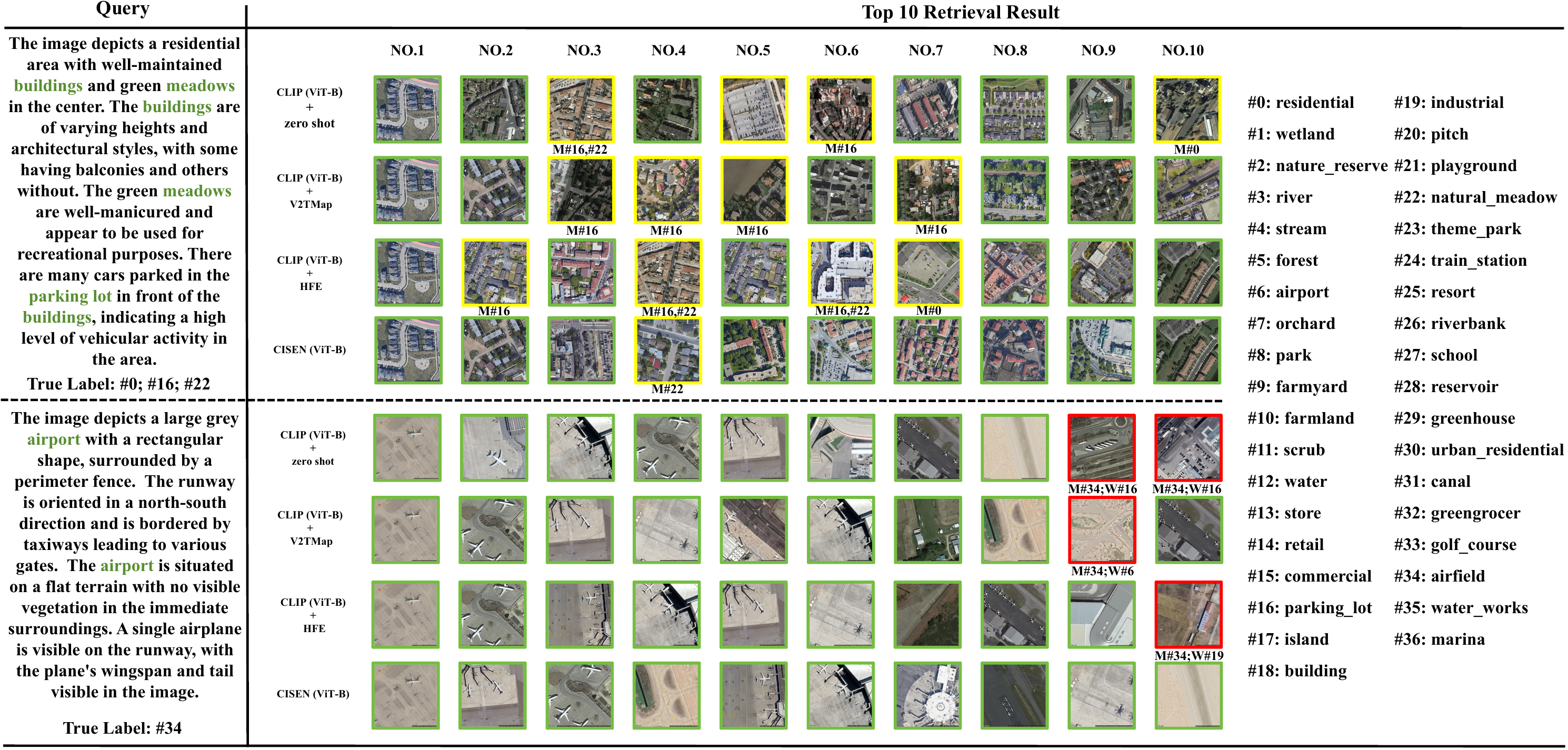}
	\caption{The T2I retrieval results of top 10 within LuojiaHOG, leveraging the integration of V2TMap and HFE with CLIP (ViT). The results with red box are incorrect, and with yellow box are inaccurate. At the bottom are some labels of third-level.}
	\label{ViT ti retrieval results}
\end{figure*}
\begin{figure*}[htbp]
	\centering
	\captionsetup{justification=justified}
	\vspace{-1cm}
	\includegraphics[width=0.8\textwidth]{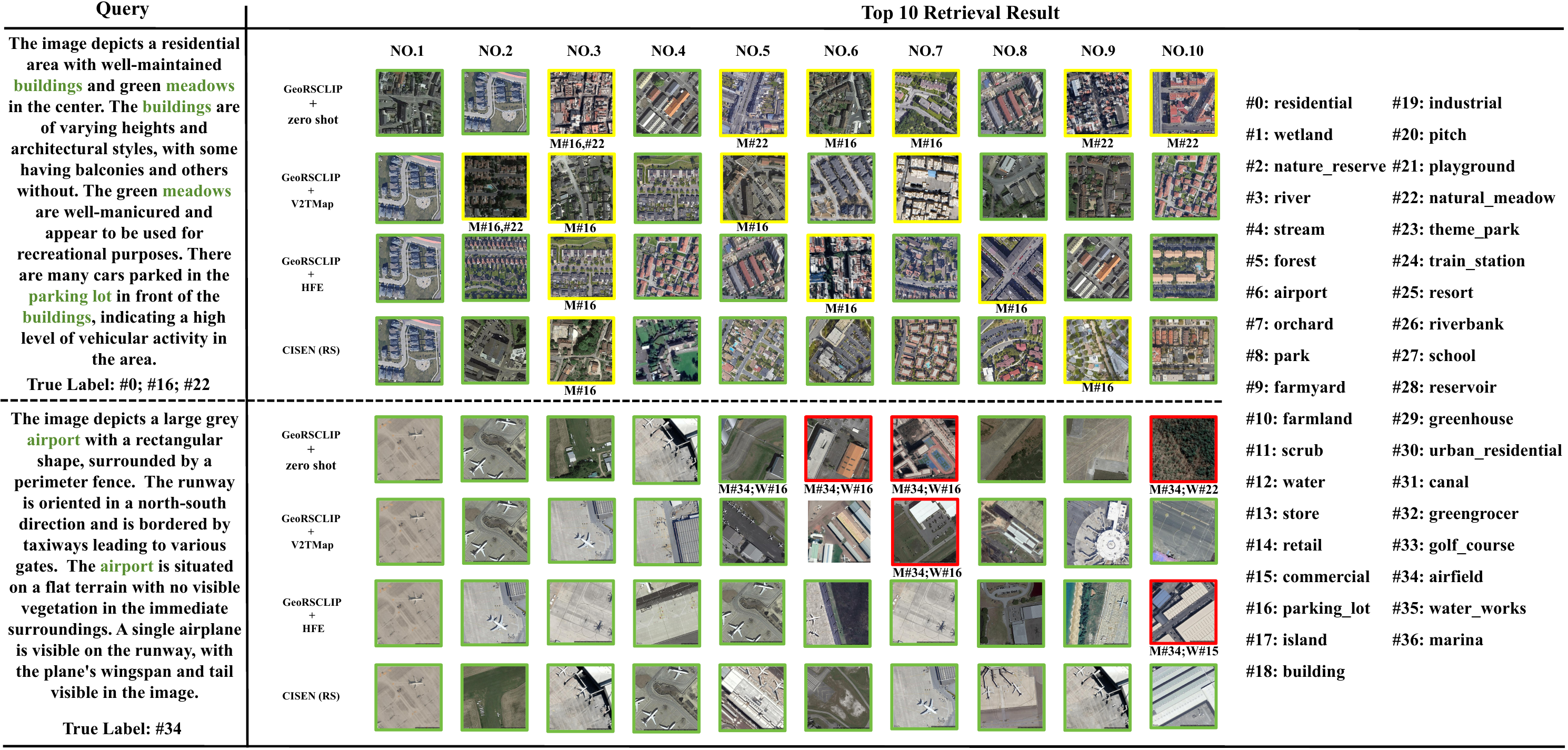}
	\caption{The T2I retrieval results of top 10 within LuojiaHOG, leveraging the integration of V2TMap and HFE with GeoRSCLIP (ViT). The results with red box are incorrect, and with yellow box are inaccurate. At the bottom are some labels of third-level.}
	\label{RS ti retrieval results}
\end{figure*}

\begin{figure*}[htbp]
	\centering
	\captionsetup{justification=justified}
	\includegraphics[width=1.\textwidth]{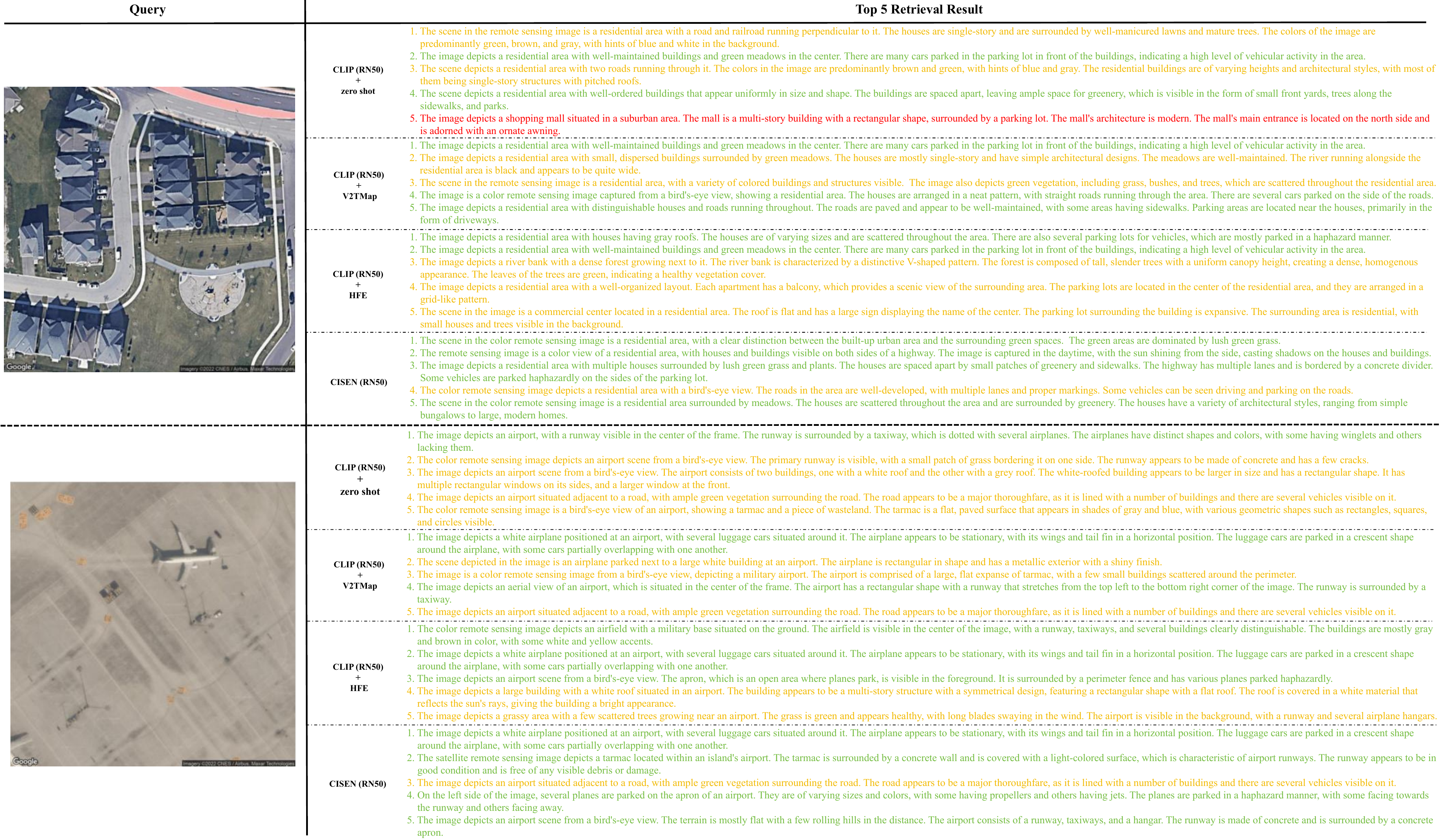}
	\caption{The I2T retrieval results of top 5 within LuojiaHOG, leveraging the integration of V2TMap and HFE with CLIP (RN50). The results in red are incorrect, and in yellow are inaccurate.}
	\label{RN50 it retrieval results}
\end{figure*}

\begin{figure*}[htbp]
	\centering
	\captionsetup{justification=justified}
	\includegraphics[width=1.\textwidth]{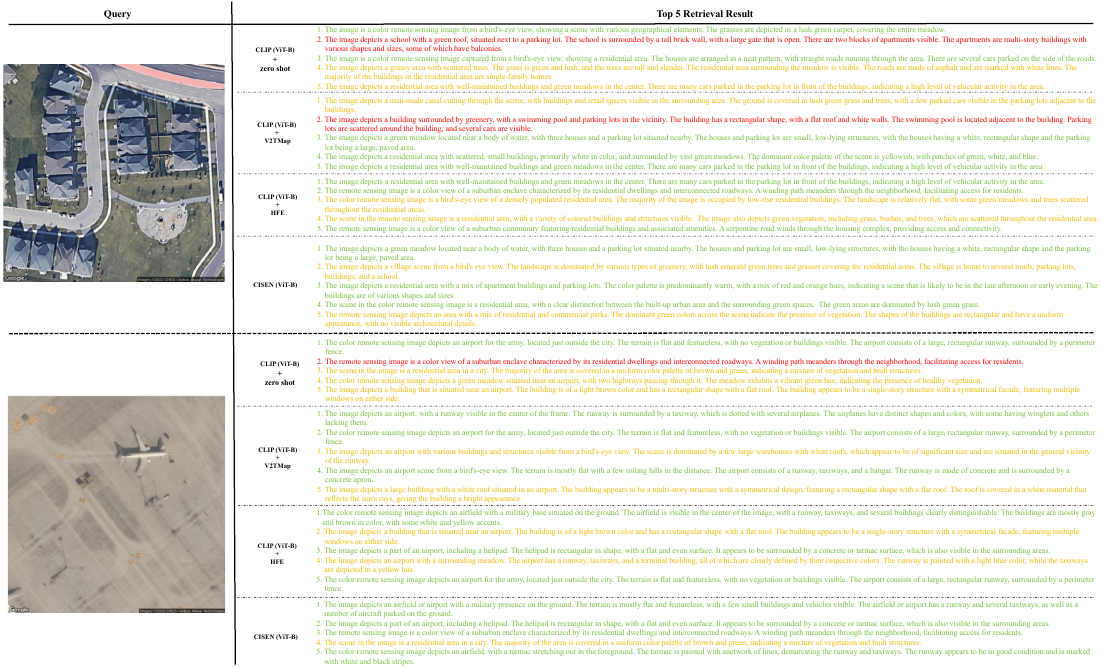}
	\caption{The I2T retrieval results of top 5 within LuojiaHOG, leveraging the integration of V2TMap and HFE with CLIP (ViT). The results in red are incorrect, and in yellow are inaccurate.}
	\label{ViT it retrieval results}
\end{figure*}
\begin{figure*}[htbp]
	\centering
	\captionsetup{justification=justified}
	\includegraphics[width=1\textwidth]{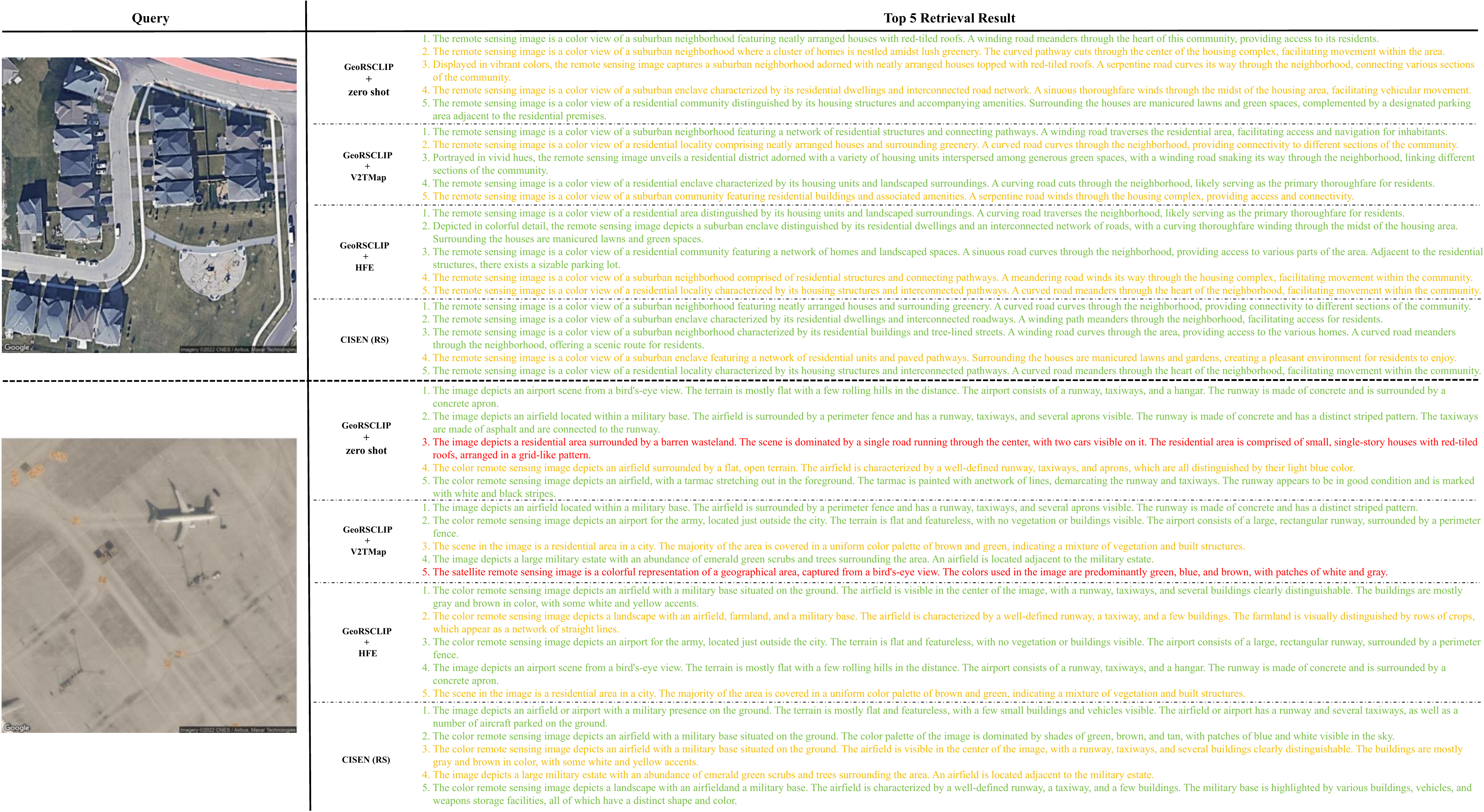}
	\caption{The I2T retrieval results of top 5 within LuojiaHOG, leveraging the integration of V2TMap and HFE with GeoRSCLIP (ViT). The results in red are incorrect, and in yellow are inaccurate.}
	\label{RS it retrieval results}
\end{figure*}

\subsubsection{Visualization Analysis}
\textbf{Feature structure }
In this section, we study the structure of the image-text features. For clarity, we select CLIP (ViT), CLIP (ViT) + V2TMap and CISEN (ViT) for comparison. First, we sample 1\% image-text pairs from LuojiaHOG and extract their features through three models. Then, their structure is displayed in the 2-D space using UMAP visualization. The visual results of the three archives can be found in Fig.~\ref{modality gap}. In CLIP (ViT), the features of images and texts are scattered in the spatial distribution. The paired image-text pairs may not necessarily be the closest in feature space, and similar image or text features are not orderly clustered together. This leads to errors in  ITR retrieval. However, features derived by our method are compact and organized. The image features (in red) are clustered internally, while the corresponding text features (in blue) are distributed externally. These results illustrate that the discrimination of enhanced visual features obtained by CISEN is high, which is beneficial to the ITR task.
\textbf{Retrieval example } To testify our method in an intuitional way, we visualize the typical ITR retrieval results. Fig \ref{it results} and Fig \ref{ti results} illustrate the retrieval performance visualization of CLIP (RN50), CLIP (ViT-B), and GeoRSCLIP fine-tuned with image adapter and corresponding CISEN. CISEN (RS) does not retrieve any incorrect images, while other models had some inaccurate retrieval results. We can also observe that the overall quality of retrieval results is largely dependent on the backbone model used. Specifically, models based on GeoRSCLIP demonstrate the best performance, followed by those based on ViT-B, and finally, those based on RN50 exhibit the poorest performance. This discrepancy is primarily attributed to differences in model architecture and training data. Notably, the GeoRSCLIP backbone, trained on remote sensing image-text data, unsurprisingly showcases superior performance in lateral comparisons under similar conditions. CISEN consistently achieves better retrieval results, primarily due to VTMap and HFE, which enable the integration of global semantic information and multi-scale image features, thereby obtaining superior feature representation. For example, features such as "boats of different sizes and types," "calm and deep water surface," and "broad and neat harbor" should be fully reflected in the retrieved text, while irrelevant words such as "residential area," "houses," and "grassland" should not be retrieved as results, as indicated by the yellow and red text representing these incorrect results. Similarly, in the task of T2I retrieval task, all models perform well in distinguishing scenes corresponding to the label "ship" but sometimes overlook the semantically related label "port". Additionally, scenes labeled with "industrial area," the neat arrangement of containers bears resemblance to containers on port docks, leading to misclassifications. When using RN50 as the backbone, the model also confuses green water surfaces with vegetation. Fig.~\ref{RN50 ti retrieval results}, Fig.~\ref{ViT ti retrieval results} and Fig.~\ref{RS ti retrieval results} are visualizations on T2I retrieval task. In general, CISEN retrieves a relatively small number of inaccurate (in yellow) and wrong results (in red), showcasting its superiority. Fig.~\ref{RN50 it retrieval results} and Fig.~\ref{ViT it retrieval results}, Fig.~\ref{RS it retrieval results} are visualizations of the retrieval performance on I2T retrieval task with three backbone model incorporating V2TMap and  HFE modules. 

	\section{Conclusions}
\label{Conclusion}
In this paper, we present LuojiaHOG with geo-awareness, comprehensive-caption and extensible-friendly, which can boost remote sensing image-caption development. Land monitoring and management heavily rely on remote sensing technology. The success of RS intelligent interpretation enables accurate identification and retrieval of interested geographic features in complex RS scenarios. With the surge of large language models and multimodal architectures, using prior knowledge as language to match and integrate with remote sensing images and further enhancing the capability of deep models in remote sensing applications is a promising research area. Existing image caption datasets often overlook geographic characteristics during sampling, and the images are mostly single-labeled, which mismatches the complexity of remote sensing images typically found in diverse scenes. Additionally, the dataset descriptions are often brief and contain a large amount of similar text, further hindering the development of multimodal models for remote sensing. To address these issues, we first explore a novel method to construct image-text datasets and create a multi-labeled image-text dataset called LuojiaHOG. Then, we propose CISEN, a method capable of enhancing features of pretrained models. Experiments conducted on LuojiaHOG dataset for RS ITR tasks, our method outperforms other state-of-the-art models across all metrics. Furthermore, we will release the LuojiaHOG dataset and demo, contributing to the advancement of research in remote sensing image-text multimodality.

In our forthcoming efforts, we aspire to expand the size of LuojiaHOG while addressing the challenges posed by the illusions inherent in large language models for automatic text generation.  Besides, more geospatial prior information will be incorporated, such as specific geographical locations, image capture seasons, climate conditions, and other relevant details. Furthermore, the dataset can be applied to a broader range of RS multi-modal downstream tasks, such as image caption, visual question answering, etc.

	\vspace{-1.45em}
	\section*{Acknowledgment}
	This work was supported by the Key Research and Development Program of Hubei Province (No. 2023BAB173), the State Key Laboratory of Geo-Information
	Engineering, NO. SKLGIE2021-M-3-1, funded by Chinese National Natural Science Foundation Projects (No. 41901265), a Major Program of the National Natural Science Foundation of China (No. 92038301), and was supported in part by the
	Special Fund of Hubei Luojia Laboratory (No. 220100028).
	\vspace{-1.2em}
%
%

	\ifCLASSOPTIONcaptionsoff
	\newpage
	\fi
	%
	%
	%
	\bibliographystyle{IEEEtran}
	\normalem
	\bibliography{LuojiaHOG.bib}
	
	\vspace{-35 pt}
\end{document}